\documentclass[acmtog,screen,nonacm]{acmart}

\makeatletter
\@namedef{ver@everyshi.sty}{}
\makeatother

\AtBeginDocument{%
  }

\usepackage{my}

\usepackage{listings}
\usepackage{xcolor}

\definecolor{codegray}{rgb}{0.5,0.5,0.5}
\definecolor{backcolour}{rgb}{0.95,0.95,0.92}

\toggletrue{arxivmode} 

\begin{document}

\title{\name: Cascaded 3D Layout Diffusion for Indoor Scene Synthesis with Implicit Relation Modeling}

\author{Yingrui Wu} 
\email{wuyingrui2023@ia.ac.cn}
\affiliation{%
  \institution{MAIS, Institute of Automation, Chinese Academy of Sciences and School of Artificial Intelligence, University of Chinese Academy of Sciences}
  \city{Beijing}
  \country{China}
}

\author{You-Kang Kong} 
\email{kykdqs@gmail.com}
\affiliation{%
  \institution{Tsinghua University}
  \city{Beijing}
  \country{China}
}

\author{Mingyang Zhao} 
\email{zhaomingyang@amss.ac.cn}
\affiliation{%
  \institution{SKLMS, Academy of Mathematics and Systems Science, Chinese Academy of Sciences and University of Chinese Academy of Sciences}
  \city{Beijing}
  \country{China}
}
\author{Weize Quan} 
\email{qweizework@gmail.com}
\authornote{Corresponding authors: Weize Quan and Yang Liu
}
\affiliation{%
  \institution{MAIS, Institute of Automation, Chinese Academy of Sciences and School of Artificial Intelligence, University of Chinese Academy of Sciences}
  \city{Beijing}
  \country{China}
}
\author{Dong-Ming Yan} 
\email{dongming.yan@nlpr.ia.ac.cn}
\affiliation{%
  \institution{MAIS, Institute of Automation, Chinese Academy of Sciences and School of Artificial Intelligence, University of Chinese Academy of Sciences}
  \city{Beijing}
  \country{China}
}
\author{Yang Liu} 
\email{yangliu@microsoft.com}

\affiliation{%
  \institution{Microsoft Research Asia}
  \city{Beijing}
  \country{China}
}

\authorsaddresses{}

\begin{abstract}
  Synthesizing realistic 3D indoor scenes remains challenging due to data scarcity and the difficulty of simultaneously enforcing global architectural constraints and local semantic consistency. Existing approaches often overlook structural boundaries or rely on fully connected relation graphs that introduce redundant generation errors. 
Inspired by human design cognition, we present \name, a cascaded diffusion framework that decomposes the joint scene generation task into four conditional sub-stages with explicit physical and semantic roles: 
(1) predicting furniture quantity and categories, 
(2) refining object sizes and feature embeddings, 
(3) modeling spatial relationships in a latent space, and 
(4) generating Oriented Bounding Boxes (OBBs). 
This decoupled architecture reduces data requirements and enables flexible integration of Large Language Models (LLMs) and Vision Language Models (VLMs) for zero-shot tasks such as image-to-scene generation. To maintain physical validity within complex floor plans, we explicitly model building elements (\eg walls, doors, and windows) as conditional constraints. Furthermore, to address the high entropy of dense relation graphs, we introduce a sparse relation graph formulation aligned with human spatial descriptions. By encoding these sparse graphs into a compact latent space using a bidirectional Variational Autoencoder (VAE), the proposed framework provides enhanced relational controllability, allowing generated layouts to better respect functional organization. Experiments demonstrate that \name achieves state-of-the-art performance in fidelity and diversity while enabling improved controllability in practical applications. 
\end{abstract}

\begin{teaserfigure}
    \centering
    \includegraphics[width=\textwidth]{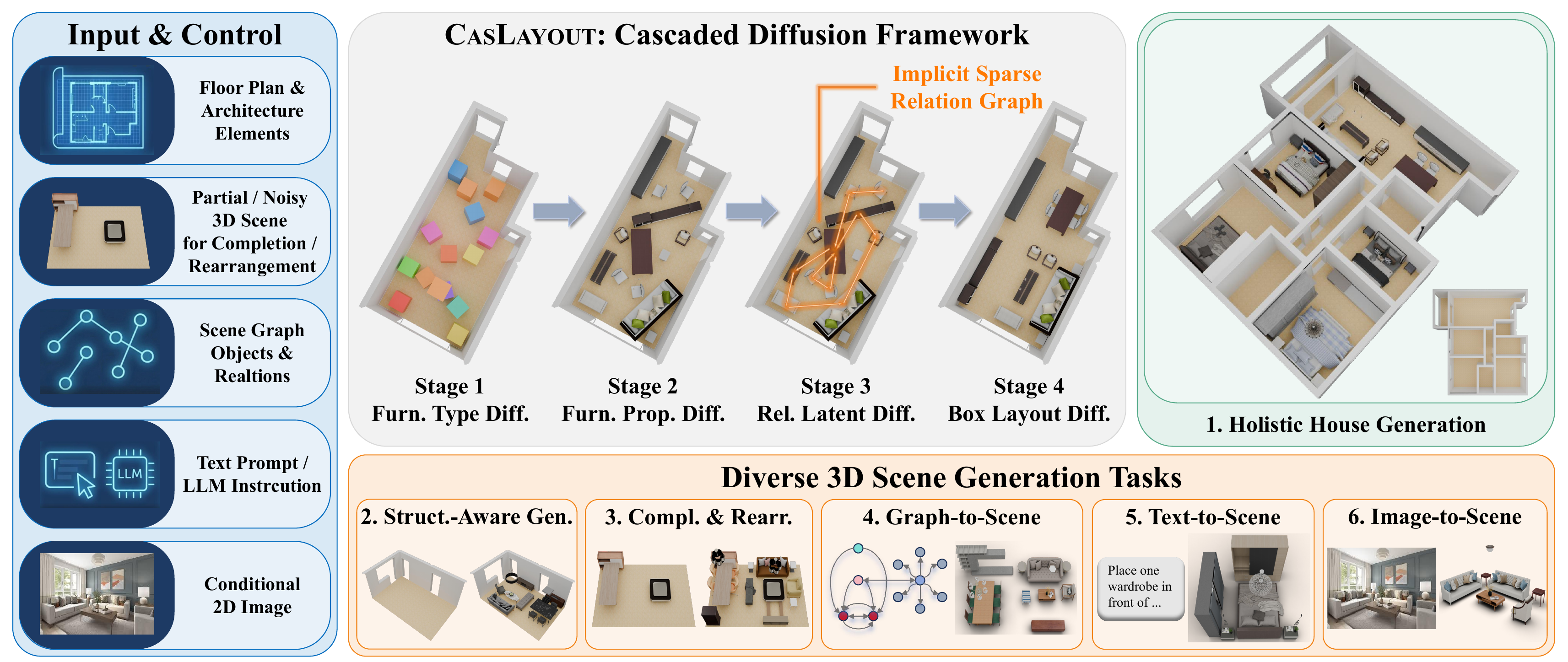}
    \caption{Aligning with designer practices, \name decomposes 3D furniture layout synthesis into four cascaded conditional diffusion processes, to deliver plausible and controllable layouts concerning room building elements, and support various layout needs.}
    \label{fig:teaser}
\end{teaserfigure}

\maketitle

\section{Introduction} \label{sec:intro}

3D indoor scene synthesis~\cite{wangcgfsurvey2023,patil2024advances} is fundamental to many applications such as virtual reality, interior design, and robotics. A key subtask is furniture layout synthesis, which involves creating and arranging furniture items based on room geometry, function, and user specifications to produce visually appealing and plausible layouts. Recent learning-based generative approaches leverage professional 3D designs to replicate realistic layouts. \looseness=-1

While recent diffusion models~\cite{tang2024diffuscene,wei2023lego,yang2024physcene,hong2024human,yi2022mime,hu2024mixed,GIULIARI2024} provide strong global context for indoor scene synthesis, they remain challenged in modeling the high-dimensional joint distribution of furniture attributes. To address this issue, several approaches incorporate scene graphs to guide the generative process~\cite{gao2023scenehgn,zhai2024commonscenes,lin2024instructscene}. However, these methods typically construct fully connected relation graphs, introducing dense interdependencies among all object pairs. During generation, such over-parameterized relational structures may result in conflicting constraints across edges, making coherent scene synthesis difficult.

Furthermore, a critical gap remains between generative capability and practical controllability. In real-world design workflows, synthesizing a plausible room is insufficient; the system must adhere to architectural constraints and respond flexibly to sparse, high-level user intents. We argue that existing limitations are largely associated with the difficulty of modeling structured dependencies in a tractable manner. First, the original joint generation space is significantly more complex than the stepwise spaces defined by conditional probabilities, making it difficult to reliably learn a complete generation manifold from limited training data. Second, prevalent relational control mechanisms rely on dense graphs, introducing high-entropy dependencies that obscure functional layout logic and misalign with human interaction patterns, which typically involve sparse relational descriptions.

To address these challenges, we present \name, a framework that prioritizes controllability to enable high-fidelity generation. Inspired by professional designer workflows, we propose a cascaded diffusion framework that decomposes the layout task into four stages: (1) \emph{furniture type}, (2) \emph{furniture property}, (3) \emph{relation latent}, and (4) \emph{box layout}. This factorization effectively reduces the high-dimensional generation problem into manageable conditional probability distributions, mitigating data scarcity while providing flexible control interfaces. Within this framework, we explicitly model \emph{building elements} as constraints to ensure architectural validity in real-world applications. Furthermore, to enable intuitive relational control, we propose \emph{implicit sparse relation modeling}. By pruning redundant information from dense graphs and encoding essential functional logic into a compact latent space via a \emph{bidirectional VAE}, we ensure that generated layouts align with both user intent and physical plausibility.

Our contributions are as follows:
\begin{enumerate}[leftmargin=*]\setlength\itemsep{0.5mm}
    \item[-] \textbf{A designer-inspired cascaded diffusion framework}  that integrates \textbf{building elements} to enforce architectural validity, decomposing the intractable high-dimensional generation space into tractable conditional distributions.
    \item[-] \textbf{A novel implicit sparse relation modeling approach}. By encoding essential functional dependencies into a compact, node-aligned latent space, this method reduces the complexity and redundancy inherent in dense relational graphs, enabling flexible layout control.
    \item[-] \textbf{The 3D-Front-Relationship dataset}, designed to capture functional relational sparsity and provide a data foundation for learning implicit relational priors.
    \item[-] \textbf{State-of-the-art performance} and \textbf{zero-shot applications} enabled by the proposed controllable generation framework, including text- and image-driven layout synthesis via integration with LLMs and VLMs.
\end{enumerate}

To facilitate future research, we release our code and dataset at \url{https://github.com/YingruiWoo/CasLayout}.
\section{Related Work} \label{sec:related}

\subsection{Indoor Scene Synthesis}
The evolution of layout synthesis has shifted from early rule-based systems~\cite{merrell2011interactive,fisher2012example} to data-driven generative models. Recent approaches largely fall into two categories: autoregressive and diffusion-based methods. Autoregressive models~\cite{wang2021sceneformer,ritchie2019fast,paschalidou2021atiss,para2023cofs,sun2024forest2seq,zhao2024roomdesigner,feng2025casagpt,wang2018deep,leimer2022layoutenhancer} generate objects sequentially, which facilitates dependency modeling but may rely on predefined ordering schemes. Diffusion-based approaches~\cite{tang2024diffuscene,wei2023lego,yang2024physcene,hong2024human,yi2022mime,hu2024mixed,GIULIARI2024,wu2025sceneflow} formulate layout generation as a denoising process and jointly model furniture attributes.

End-to-end diffusion models typically attempt to learn the joint distribution of object types, features, and oriented bounding boxes (OBBs) directly, introducing challenges associated with high dimensional generation spaces under limited training data. To alleviate this issue, several works explore partial decomposition strategies. Recent approaches~\cite{maillard2024debara,zhai2024commonscenes,naanaa20233d,zhai2024echoscene} explicitly separate OBB generation from fine-grained shape geometry. Methods such as PlanIT~\cite{wang2019planit} and GRAINS~\cite{li2019grains} adopt hierarchical structural representations, while InstructScene~\cite{lin2024instructscene} models global relational graphs independently from OBB prediction. SemLayoutDiff~\cite{sun2025semlayoutdiff} further separates top-down semantic maps, and GLTScene~\cite{li2024gltscene} introduces wall-centered local optimization to disentangle global layout dependencies.

In contrast to these partial decomposition strategies, \name implements a systematic four-stage factorization inspired by human design workflows to improve synthesis controllability. Furthermore, while some existing methods treat the room as an empty bounding volume~\cite{lin2024instructscene}, or rely on local heuristics~\cite{li2024gltscene}, \name explicitly models building elements as global conditional constraints throughout the cascade to maintain architectural consistency under complex floor plans.

\subsection{Spatial Relation Modeling}
Modeling spatial relationships is essential for capturing interior layout principles~\cite{grimley2022universal,rolph7principles,tru13principles}. Early works~\cite{merrell2011interactive,fisher2011characterizing,yeh2012synthesizing,weiss2018fast,yu2011make} relied on hand-crafted heuristics to enforce spatial constraints, which provide strong controllability but limit layout diversity.

Subsequent learning-based approaches introduced structured representations such as trees or graphs, as exemplified by GRAINS~\cite{li2019grains} and PlanIT~\cite{wang2019planit}, to capture spatial dependencies between objects. These localized structures simplify relational modeling but may require additional mechanisms to maintain global structural coherence in complex scenes.

Recent methods extend this paradigm by modeling scene layouts using dense relation graphs~\cite{xu2024set,yao2024conditional,fang2025text}. Approaches such as SceneHGN~\cite{gao2023scenehgn}, CommonScenes~\cite{zhai2024commonscenes}, and InstructScene~\cite{lin2024instructscene} employ fully connected graphs to capture broader contextual dependencies. However, modeling all-to-all interactions increases the dimensionality of the relational space and introduces additional dependencies that must be resolved during generation. Existing strategies, such as structural masking~\cite{lin2024instructscene} or assuming relational priors are externally provided~\cite{dhamo2021graph,kikuchi2021constrained,gao2023scenehgn,zhai2024commonscenes}, attempt to alleviate this issue from different perspectives.

Building upon these observations, \name adopts an implicit sparse relation modeling strategy that focuses on capturing essential functional dependencies in a compact latent space.

\subsection{LLM-Driven Scene Generation}

Recent advances in large language models (LLMs) have inspired new pipelines for scene generation. A common paradigm prompts LLMs to directly generate layout specifications (\eg JSON or coordinates)~\cite{yang2024llplace,ocal2024sceneteller,feng2023layoutgpt,wang2024chat2layout,gumin2025procedural}. However, mapping semantic outputs to continuous geometric layouts remains challenging, particularly when enforcing spatial consistency under complex floor plans.

Other hybrid approaches incorporate iterative sampling or optimization procedures~\cite{fu2025anyhome,ccelen2024design,yang2024holodeck,deng2025global,sun2025layoutvlm,aguina2024open,littlefair2025flairgpt}. These approaches typically rely on explicitly specified constraints, which may limit the enforcement of implicit spatial relationships during generation.

As an application of controllable synthesis, \name adopts an alternative integration strategy by using LLMs as semantic planners to generate furniture lists and sparse relational descriptions, while the proposed cascaded diffusion framework acts as a geometric executor that translates these high-level signals into precise 3D layouts via conditioned diffusion.

\begin{figure*}[t]
    \centering
    \includegraphics[width=0.95\linewidth]{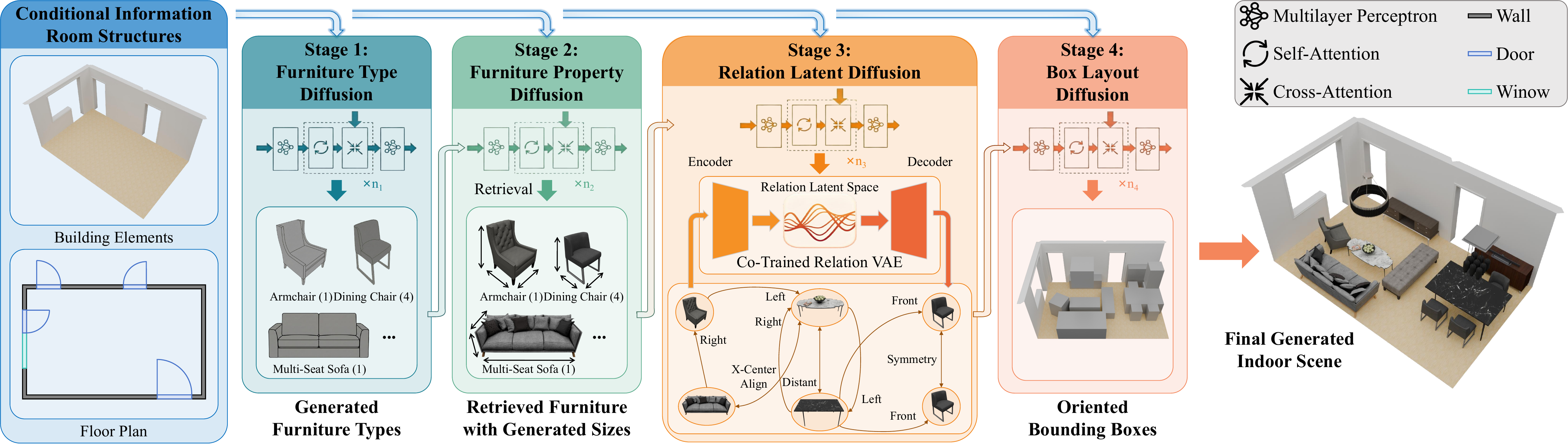}
    \caption{The pipeline of \name, which consists of four diffusion processes in a cascaded way: \emph{furniture type diffusion}, \emph{furniture property diffusion}, \emph{relation latent diffusion}, and \emph{box layout diffusion}. All diffusion networks are built upon a series of self-attention layers. The 2D floor plan (as image) serves as conditioning for each diffusion network via cross-attention layers. The output of each diffusion network is used to construct the input for the subsequent diffusion process.
    }
    \label{fig:framework}
\end{figure*}

\section{Method} \label{sec:method}

\subsection{Overview}
Given an empty room with prescribed architectural elements, such as a floor plan, walls, doors, and windows, furniture layout synthesis involves selecting and placing suitable furniture items to create a plausible arrangement. We define the key components of a furniture layout and model spatial relations in \zcref{subsec:layout,subsec:relation}. Inspired by designer workflows (as described in \zcref{sec:intro}), we decompose the synthesis process into four conditional diffusion stages:

\begin{enumerate}[leftmargin=*]
  \item \textbf{Furniture type diffusion} (\zcref{subsec:typediffusion}): Generates the number and categories of furniture items based on the given architectural elements.
  \item \textbf{Furniture property diffusion} (\zcref{subsec:sizediffusion}): Determines the detailed size and feature embedding of each furniture item. These embeddings, combined with furniture types, are used to retrieve shape geometry from a database.
  \item \textbf{Relation latent diffusion} (\zcref{subsec:relationdiffusion}): Synthesizes plausible spatial relationships among furniture items and architectural elements. Relations are represented in a compact latent space learned by a bidirectional Relation VAE, reducing the quadratic complexity of pairwise relations to linear complexity.
  \item \textbf{Box layout diffusion} (\zcref{subsec:boxdiffusion}): Generates OBBs for furniture items, conditioned on the furniture list, architectural elements, and synthesized relations.
\end{enumerate}

Each stage conditions the next, forming a cascaded pipeline illustrated in \zcref{fig:framework}. This design reduces data requirements, improves modularity, and enables fine-grained control. We refer to this framework as \textbf{\name}. Applications enabled by \name are discussed in \zcref{subsec:applications}. 
\subsection{Layout Formulation} \label{subsec:layout}

A furnished 3D room, denoted by $\mL$, consists of the following components:
\begin{enumerate}[leftmargin=*]
  \item[-] \textbf{Architectural elements}: A floor plan represented as a binary image, and a set of walls, doors, and windows, denoted as $\{\bbuilding_i\}_{i=1}^{m}$. Each vertical element is characterized by its type $c_i \in \{\texttt{wall}, \texttt{door}, \texttt{window}\}$ and its OBB $\bm{B}_i$, with the y-axis pointing toward the room interior and the z-axis upright.
  \item[-] \textbf{Furniture items}: A collection of furniture objects, denoted as $\{\bfurniture_j\}_{j=1}^{n}$. Each $\bfurniture_j$ is defined by its type $c_j$, feature embedding $\bm{feat}_j$, and OBB $\bm{B}_j$ with an upright z-axis.
\end{enumerate}

\paragraph{OBB parameterization}
Each OBB is parameterized by its size ($\bm{s} \in \mathbb{R}^3$), translation relative to the room center ($\bm{t} \in \mathbb{R}^3$), and rotation vector ($\bm{r} = (\cos \theta, \sin \theta) \in \mathbb{R}^2$), where $\theta$ is the rotation angle with respect to the canonical pose in the furniture database. Under the local coordinate system, the $+x$ and $+y$ axes of the OBB define the \emph{left} and \emph{front} directions, respectively; the opposite directions (\emph{right}, \emph{behind}) are inferred accordingly.

\begin{figure}[t]
    \centering
    \includegraphics[width=\linewidth]{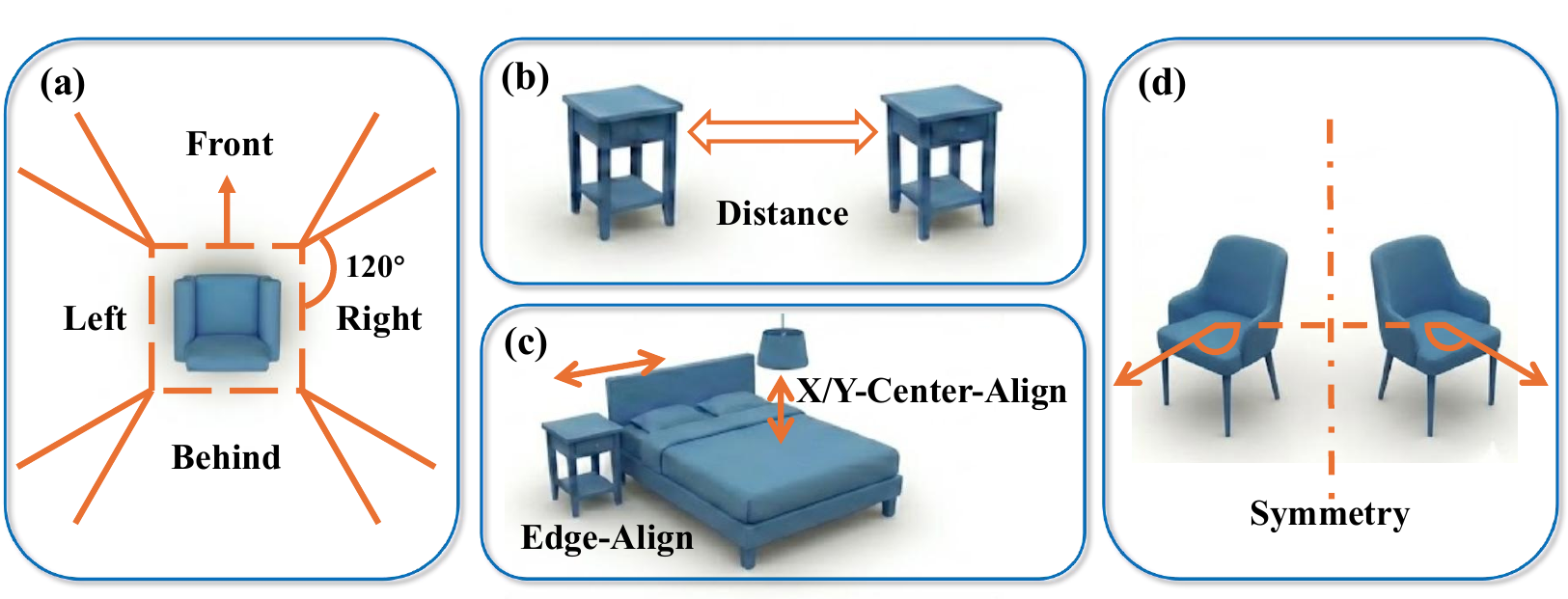}

    \caption{Spatial relationships. (a) direction determined by the local coordinate system and OBB size ratios, (b) distance defined as the minimum separation between two OBBs, (c) alignment of OBB edges and centers, and (d) symmetry between OBBs with the same type and feature embedding.}
    \label{fig:relation}
\end{figure}

\subsection{Relation Capture} \label{subsec:relation}
Professional furniture layouts adhere to common design principles, which are closely tied to spatial relationships. To capture these principles, we define a set of pairwise and relative spatial relations based on the OBBs of furniture items and architectural elements.

To ensure \emph{rotation invariance} and \emph{better reflect design principles}, all spatial relations are defined in the local coordinate system of each object. This approach captures intrinsic relationships independent of the room's global orientation, improving robustness and generalizability. As illustrated in \zcref{fig:relation}, the spatial relations are categorized into direction, distance, alignment, and symmetry between furniture items, as well as distance relations between furniture and architectural elements.

\paragraph{Relations between furniture items}

\begin{enumerate}[leftmargin=*]
  \item[-] \textbf{Direction}: The relative position of $\bm{B}_1$ to $\bm{B}_2$ can be \emph{left}, \emph{right}, \emph{front}, \emph{behind}, \emph{under}, or \emph{above}. For xy-plane directions, we project two rays at \SI{120}{\degree} angles from $\bm{B}_2$'s edges. For z-axis directions, $\bm{B}_1$ is considered \emph{under} or \emph{above} $\bm{B}_2$ if its xy-coordinates lie within $\bm{B}_2$’s xy-projection.
  \item[-] \textbf{Distance}: Defined by the minimum distance between two OBBs and categorized as \emph{attach to}, \emph{adjacent}, or \emph{distant}.
  \item[-] \textbf{2D Alignment}: Under an orthogonal view, an OBB may align with another along its edge or local axis, yielding \emph{edge-align}, \emph{x-center-align}, and \emph{y-center-align}.
  \item[-] \textbf{Symmetry}: Two furniture OBBs are \emph{symmetric} if they share the same type, feature embedding, and size, and their $+y$-axes are mirror-symmetric on the same horizontal plane.
\end{enumerate}

\paragraph{Relations between furniture and architectural elements}
\begin{enumerate}[leftmargin=*]
  \item[-] \textbf{Distance}: Similar to furniture--furniture distance, but the OBBs of architectural elements are represented as line segments. We consider the distance only to the nearest wall, which corresponds to the distance from the furniture to the floor plan boundary, but compute the distances to all doors and windows.
\end{enumerate}

\paragraph{Implicit sparse relation modeling}
While spatial relation modeling is essential, the prevalent reliance on fully connected dense graphs poses significant challenges for three key issues:
\begin{enumerate}[leftmargin=*]
  \item[-] Dense graphs impose exhaustive pairwise constraints ($N^2$ complexity), tightly coupling relations such that errors in individual predictions may propagate through these constraints, leading to logical inconsistencies and geometric conflicts.
  \item[-] Empirically, most object pairs lack intrinsic design logic, with spatial relations approximating a uniform distribution. Learning these non-functional dependencies injects high-entropy noise, \ie relation categories whose distributions approach uniformity, thereby diluting essential functional constraints.
  \item[-] Control information in practical applications is inherently sparse. Since human design relies on specifying key functional constraints rather than exhaustive pairwise relations, sparse graphs serve not merely as a computational optimization, but as a faithful mapping of human design cognition.

\end{enumerate}

To address this, we propose a \emph{sparse relation graph} extraction method. Furniture is grouped into functional zones (\eg dining, lounging), and relations within each zone are preserved for coherence. For inter-zone modeling, we select an anchor item in each zone and define relations between these anchors.
To validate this design, we measure the entropy of relational distributions for both dense and sparse graphs. Here, entropy quantifies the uncertainty of spatial relation categories across object pairs. The entropy is computed as follows~\cite{bustin2021lossy}:
\begin{equation}
  H(X) = -\sum_{i=1}^{n}\sum_{j=1}^{m} p(x_{ij}) \log_2 p(x_{ij})/n,
\end{equation}
where $n$ denotes the number of pairwise relations, representing the exhaustive set of relation pairs for dense graphs and the selected relation pairs for sparse graphs. $m$ is the total number of sub-categories for each relation, and $x_{ij}$ refers to the frequency of the $j$-th category within the $i$-th relation pair. Taking the direction relations in living rooms as an example, the average entropy decreases \emph{from 1.62 to 1.04}.
This reduction indicates an increase in information density, suggesting that we filter out near-uniform background relations while preserving low-entropy structures that capture essential design logic (see more design details in \arxiv{\zcref{subsec:rd}}{Suppl.~Sec.~1.1}).

Following graph sparsification, maintaining an explicit matrix representation ($N^2$) introduces significant redundancy. This results in an excessively sparse generation space that degrades the quality of graph synthesis. To address this, we design an implicit relation representation and leverage a bidirectional attention mechanism to encode relational edges into $N$ node embeddings, effectively compressing the generation space (see \zcref{subsec:relationdiffusion}).

\paragraph{3D-Front-Relationship}
Based on this capture strategy, all relations form a directed graph where edges indicate reference OBBs. From the widely used 3D-Front dataset~\cite{fu20213dfront}, we extract these relations to construct \emph{3D-Front-Relationship}, a dataset designed for learning and synthesizing plausible spatial relations in our method.

\subsection{Furniture Type Diffusion} \label{subsec:typediffusion}

We adopt the Denoising Diffusion Probabilistic Model (DDPM)~\cite{ho2020denoising} for all four stages in \name. The first stage predicts the number and types of furniture items, conditioned on the architectural elements of the room.

\paragraph{Network input}
A layout $\mL$ is represented as a vector set:
$
  \bm{X} = \{\bm{e}_1, \ldots, \bm{e}_{m+n}, \ldots, \bm{e}_{n_{\max}}\},
$
where $m$ and $n$ denote the number of architectural elements and furniture items, respectively, and $n_{\max}$ is a preset maximum to ensure a consistent vector length across training samples. All elements are randomly ordered, and positional encoding distinguishes elements of the same type.

Each vector $\bm{e}_i$ concatenates $\left(\bm{c}_i, \bm{s}_i, \bm{t}_i, \bm{r}_i, \bm{pe}_i\right)$,
where $\bm{c}_i$ is a one-hot type encoding and $\bm{pe}_i$ is the positional encoding. For furniture items, $\bm{s}_i$, $\bm{t}_i$, and $\bm{r}_i$ are set to zero since OBBs are not yet determined. Vectors beyond $m+n$ correspond to non-existent elements and are assigned a \emph{None} class with zeroed spatial attributes.  More details of this vector representation for four stages are provided in \arxiv{\zcref{subsec:vec}}{Suppl.~Sec.~1.3}.

\paragraph{Network design}

We implement DDPM using a Transformer backbone. Spatial attributes and category features are encoded separately, concatenated, and augmented with positional embeddings to form input tokens. The Transformer consists of five sequential modules, each combining a self-attention layer (to capture dependencies within $\bm{X}$) and a cross-attention layer (to integrate external information from the binary floor plan image $\mathcal{F}$). This architecture serves as the base model for all four diffusion stages.

During training, Gaussian noise $\boldsymbol{\epsilon}$ is added to the type vectors of furniture items:
\[
  \bm{c}_{i,t} = \sqrt{\gamma(t)}\bm{c}_i + \sqrt{1-\gamma(t)}\boldsymbol{\epsilon},
\]
where $t$ is the time step and $\gamma(t)$ is the noise scheduler. The network $\epsilon_{net}$ predicts the added noise by minimizing:
\begin{equation}
  \mathbb{E}_{\boldsymbol{\epsilon} \sim \mathcal{N}(\boldsymbol{0}, \boldsymbol{I}), \, t \sim \mathcal{U}(0, 1)} \|\boldsymbol{\epsilon}_t - \epsilon_{net}(\bm{X}_t, \mathcal{F}, t)\|_2^2,
\end{equation}
where $\bm{X}_t$ is the noisy version of $\bm{X}$ at step $t$.

To preserve architectural element information during forward propagation, we introduce an auxiliary reconstruction of their OBBs (size $\bm{s}$, translation $\bm{t}$, and rotation $\bm{r}$) at the output stage, supervised by an MSE loss:
\begin{equation}
  L_{\mathrm{rc}} = \frac{1}{m}\sum_{i=1}^m \|\bm{B}_i - \hat{\bm{B}}_i\|_2^2,
\end{equation}
where $\bm{B}_i$ and $\hat{\bm{B}}_i$ denote the ground-truth and reconstructed OBBs, respectively. This auxiliary loss is applied consistently across all diffusion stages.

\paragraph{Inference} During inference, Gaussian noise is assigned to the type vectors of all non-architectural elements. The denoising process predicts their furniture types, including potential nonexistence, thereby determining the furniture count.

\subsection{Furniture Property Diffusion} \label{subsec:sizediffusion}

In the second stage of \name, given the furniture categories and quantities predicted in the first stage, the DDPM generates the size and feature embedding for each furniture item. This step determines the precise dimensions and shape characteristics required for realistic placement within the scene.

\paragraph{Network input}
Building on the vector set from Furniture Type Diffusion, we extend each element to include feature embeddings:
$
  \left(\bm{c}_i, \bm{feat}_i, \bm{s}_i, \bm{t}_i, \bm{r}_i, \bm{pe}_i\right),
$
where $\bm{c}_i$ for furniture items is fixed from the previous stage. Gaussian noise is added to $\bm{s}_i$ (size) and $\bm{feat}_i$ (feature embedding) following the standard forward diffusion process.

\paragraph{Network design}
The type, OBB attributes, and feature embedding of each vector are encoded individually, concatenated, and combined with positional embeddings to form node tokens for the Transformer network. The architecture mirrors that used in Furniture Type Diffusion, ensuring consistency across stages.

\paragraph{Feature embedding}
We adopt a pre-trained VQ-VAE~\cite{lin2024instructscene} to extract latent feature indices for CAD models from the 3D-Future dataset~\cite{fu20213dfront}. During inference, the generated feature embeddings are used to retrieve CAD models from 3D-Future with the same class label and the closest geometry feature. The technical details are provided in \arxiv{\zcref{subsec:fi}}{Suppl.~Sec.~1.2}.

However, due to the limited size of our dataset, conditioning subsequent stages on feature embeddings leads to overfitting, degrading relation quality and placement accuracy. To mitigate this, feature embeddings are generated solely for furniture retrieval and are not used as conditioning inputs in later stages.

\subsection{Relation Latent Diffusion} \label{subsec:relationdiffusion}

Given the architectural elements and a set of furniture items with known types and sizes, our goal is to synthesize plausible spatial relations among these elements.
We model spatial relations in a compact latent space for two primary reasons: first, to mitigate the quadratic growth of pairwise relations with respect to object count; and second, to exploit the compression space created by graph sparsification, which filters out non-essential information and yields a more compressible latent structure. This latent representation is conditioned on furniture types, sizes, and architectural elements, enabling relation synthesis via diffusion.

\subsubsection{Bidirectional relation VAE}
To learn this latent space, we design a transformer-based Variational Autoencoder (VAE) that encodes relation graphs into node-aligned latent vectors, ensuring semantic consistency and facilitating diffusion-based generation.

\paragraph{Relation graph representation}
For a layout $\mL$, pairwise relations are represented as a directed graph where nodes correspond to furniture items or architectural elements. Each node is associated with a feature vector $\left(\bm{c}_i, \bm{s}_i, \bm{t}_i, \bm{r}_i, \bm{pe}_i\right)$. To prevent overfitting to absolute geometry, the VAE uses only $\bm{c}_i$ and $\bm{pe}_i$ for furniture nodes, while OBBs of architectural elements are retained as conditions. Each directed edge $\overrightarrow{v_i v_j}$ carries features $\left(\bm{pe}_i,\bm{pe}_j,\bm{R}_{ij},\bm{R}_{ij}^s\right)$, where $\bm{R}_{ij}$ encodes the relation category (\eg \emph{Distance}, \emph{Alignment}) and $\bm{R}_{ij}^s$ encodes the subcategory (\eg \emph{left}, \emph{right}), both as one-hot vectors.

\begin{figure}[t]
    \centering
    \includegraphics[width=\linewidth]{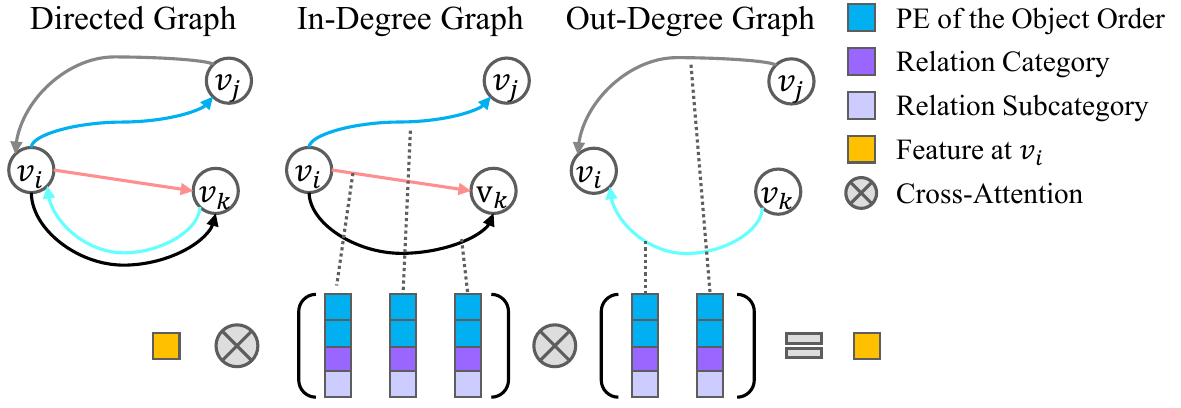}
    \caption{Illustration of the in-out cross-attention mechanism. For object $v_i$, two vector sets are separately constructed on its in-degree and out-degree graphs, respectively. The feature at $v_i$ is updated by applying cross-attention between it and the two vector sets sequentially.}
    \label{fig:inout}
\end{figure}

\paragraph{Encoder design}
For each node $v_i$, we decompose its subgraph into an in-degree graph ($\mathcal{G}_{in}$) and an out-degree graph ($\mathcal{G}_{out}$). We apply cross-attention twice: first using incoming edge features as keys/values, and then using outgoing edge features. This two-step mechanism, termed \emph{in-out cross attention}, aggregates relational context from both directions. Afterward, node features interact via self-attention. These operations form an \emph{in-out-attention block}, and multiple blocks are cascaded in the encoder. The encoder outputs the mean and variance for latent sampling. \zcref{fig:inout} illustrates this mechanism with a simple example. In practice, we find it performs better than considering all edges at once (see the ablation in \zcref{subsec:ablation}).

\paragraph{Decoder design}
The decoder consists of self-attention blocks and predicts node features. For relation prediction, features of two nodes are concatenated and passed through category-specific MLPs to predict subcategory labels, including \emph{None}. Cross-entropy loss supervises relation classification, while KL-divergence regularizes the latent space with a small weight (\num{0.001}).

\subsubsection{Relation latent diffusion}
Once trained, the VAE provides each layout with a node-aligned relation latent vector set. We then design a DDPM to synthesize these latents.

\paragraph{Network input}
Each element $\bm{e}_i$ is extended to:
$
  \left(\bm{c}_i,\bm{s}_i,\bm{t}_i,\bm{r}_i,\bm{pe}_i,\bm{rl}_i\right),
$
where $\bm{rl}_i$ is the relation latent for element $i$. $\bm{c}_i$ and $\bm{s}_i$ are conditions from previous stages. During training, Gaussian noise is injected into all $\bm{rl}_i$ values, and the network predicts the noise.

\paragraph{Network design} Node types, spatial attributes, and relation latents are encoded separately and concatenated into node tokens with positional embeddings. The Transformer architecture mirrors that used in Furniture Type Diffusion.

\subsection{Box Layout Diffusion} \label{subsec:boxdiffusion}
With furniture types, sizes, and relation latents determined, the final stage predicts the placement of furniture OBBs to produce a semantically coherent and controllable layout. We train a DDPM model for this purpose.

\paragraph{Network input}
Each element vector $\bm{e}_i$ is represented as:
\[
  \left(\bm{c}_i, \bm{s}_i, \bm{t}_i, \bm{r}_i, \bm{pe}_i, \bm{rl}_i\right),
\]
where $\bm{c}_i$, $\bm{s}_i$, and $\bm{rl}_i$ are conditional inputs from previous stages. Gaussian noise is injected into $\bm{t}_i$ and $\bm{r}_i$, and the network predicts the injected noise during denoising.

\paragraph{Network design}
The input token encodes node type and OBB attributes separately, and then concatenates them with positional embeddings. The Transformer architecture is extended with a \emph{self-cross-cross attention} mechanism:
(1) The first cross-attention layer attends to relation latents, enforcing consistency with inter-object relationships;
(2) The second cross-attention layer attends to floor plan image features, capturing global room context.
This design ensures that local placement respects both relational constraints and overall spatial structure.

\paragraph{Co-training scheme}
Since relation latents directly influence box layout diffusion, we jointly train the relation VAE and this diffusion stage. Backpropagating diffusion errors through the VAE improves latent compatibility with layout synthesis. We balance the VAE and diffusion losses with a weight ratio of $1:1$. After joint training, the learned latent space guides the training of relation latent diffusion.


\subsection{Supported Applications} \label{subsec:applications}
\name supports a variety of applications, and additional experimental results are shown in \zcref{sec:results}. The implementation details are available in \arxiv{\zcref{subsec:dapp}}{Suppl.~Sec.~.5}.
\begin{figure}[t]
    \centering
    \includegraphics[width=\columnwidth]{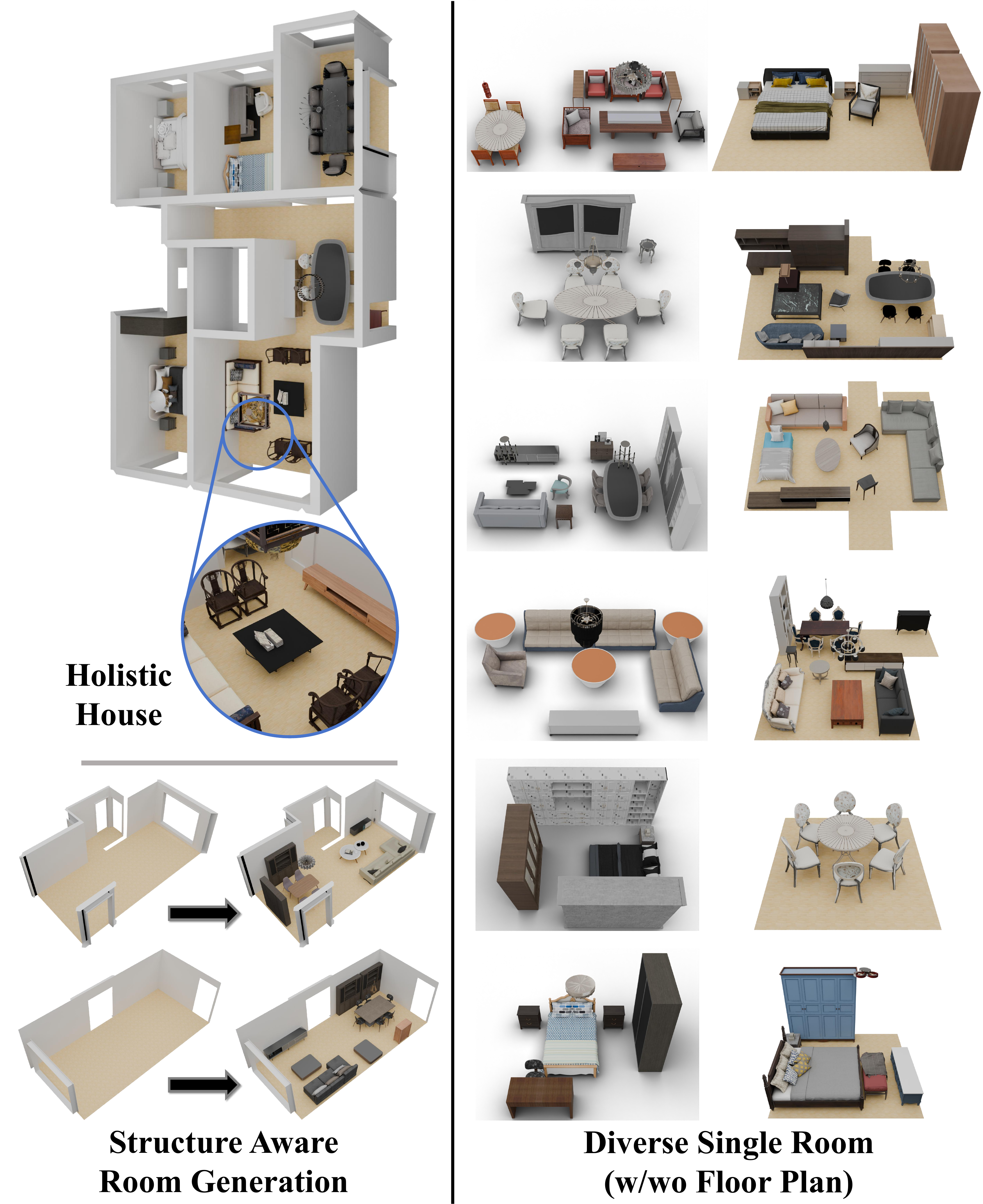}
    \caption{A gallery of synthesized scenes demonstrating generation quality and diversity.}
    \label{fig:gallery}
\end{figure}

\paragraph{Layout generation and editing}
The cascaded diffusion pipeline enables users to modify building elements, furniture types, OBB sizes, and spatial relations at any stage by adjusting network inputs, while preserving attributes intended to be retained during denoising. This flexibility supports iterative design workflows and interactive editing, which are highly desirable in practice. \zcref{fig:gallery} showcases diverse layouts generated by \name under various conditions.

\paragraph{Furniture rearrangement} \name can adapt existing layouts to better align with design principles. By providing architectural elements, furniture types, and OBB sizes as initial conditions for relation latent diffusion, new spatial relations are synthesized. Box layout diffusion then determines updated positions and orientations for furniture items. Optionally, users can omit certain OBB sizes and allow the model to suggest new dimensions during generation.

\paragraph{Layout completion}
To complete partial layouts, we introduce a masking strategy during training: randomly selected furniture items are masked in the furniture type diffusion stage but retain their original attributes as conditional inputs. At inference, users can input an incomplete layout, and the model predicts additional furniture types. These, along with existing items, are passed through subsequent stages to generate relation latents and OBBs for the new furniture.

\paragraph{LLM-guided scene generation}
Thanks to the modular design, our method can integrate Large Language Models (LLMs) by replacing the furniture type and relation diffusion stages, both naturally expressible in language, with LLM outputs. This approach leverages LLMs for semantic reasoning while avoiding their limitations in precise numerical prediction. The diffusion models then enforce physical plausibility and controllability, enabling scalable layout generation from language-based inputs. The detailed prompts are provided in \arxiv{Appendix \zref{sec:prompts}}{Suppl.~Sec.~2}.

\paragraph{User-specified sparse relation control}
Unlike prior methods that rely on fully connected relation graphs, \name employs sparse relation graphs that focus on key relationships, reducing complexity and aligning with human intuition. This sparsity simplifies manual specification of relations, allowing users to exert targeted control over scene generation without overwhelming design constraints.

\paragraph{Text-driven scene generation}
Leveraging the flexibility of our cascaded architecture, \name employs LLMs to extract furniture entities and spatial relations from the input text to construct a sparse scene graph. After completing missing furniture elements, the model generates an interior scene that aligns with the textual description.

\paragraph{Image-driven scene generation}
Analogous to text-driven synthesis, \name employs VLMs to extract furniture entities and relational information from images, enabling image-driven scene generation. Since our spatial relations are defined within local coordinate systems, the VLM can construct accurate scene graphs without requiring global reference coordinates. Furthermore, unlike the text-driven pipeline, \name refrains from performing auto-completion on the extracted information to ensure strict consistency between the generated scene and the control image.

\section{Experimental Results} \label{sec:results}

\subsection{Experimental Setup} 

\paragraph{Dataset and data augmentation} Following DiffuScene~\cite{tang2024diffuscene} and InstructScene~\cite{lin2024instructscene}, we use the synthetic indoor scene dataset 3D-FRONT~\cite{fu20213dfront}, maintaining the same training and testing splits across three categories: \num{4041} bedrooms, \num{900} dining rooms, and \num{813} living rooms. From this dataset, we extract spatial relations (\zcref{subsec:relation}) and architectural elements to construct our relation dataset.

To support layout completion, we introduce a masking strategy during training: randomly selected furniture items are masked while retaining their original attributes as conditions, simulating user-specified partial scenes. In addition, to enable generation without explicit floor-plan control, we augment the data by setting all structural elements to a \emph{None} type and assigning a value of \num{1} to every pixel in the binary floor plan mask. This augmentation enables our model to handle scene generation, completion, and rearrangement under both controlled and uncontrolled settings (see details in \arxiv{\zcref{subsec:da}}{Suppl.~Sec.~1.4}). Unlike prior methods that either address a single task without floor plan control~\cite{lin2024instructscene} or require separate models for different tasks~\cite{tang2024diffuscene}, our approach offers improved flexibility and generalization.

For fair benchmarking, we also train our model using the same data settings as competing methods, excluding these augmentations. In the comparative evaluations, we denote the version trained with augmentations as \name-G and the one without as \name. Unless explicitly stated otherwise, ``Ours'' or \name refers to the augmented version by default. Additional augmentation details are provided in the supplemental material.

\paragraph{Network training}
All diffusion networks are trained on a single NVIDIA A100 GPU with a batch size of 256 for 2000 epochs (256 iterations per epoch). The learning rate is initialized to \num{1e-4} with the AdamW optimizer and decays by \num{0.02} per epoch. The relation latent dimension is set to \num{32}, and the feature embedding dimension is set to \num{64}.

\begin{table*}[t]
    \centering
    \caption{Comparison of our method with existing approaches on floor plan conditional generation.
        The best result is shown in \best{bold}, and the second-best is  \second{underlined}.
        \textbf{N/A} indicates that the method does not support the specific application scenario or evaluation metric. \name-G is our general model trained with data augmentation, which achieves state-of-the-art performance across various tasks and experimental settings without the need for retraining. \name is a counterpart trained under the same data setting as other methods for a fair comparison.}
    \label{tab:comp_w_floor}

    \footnotesize
    \setlength{\tabcolsep}{1.8pt}
    \renewcommand{\arraystretch}{1.35}

    \begin{tabular}{l ? ccccc ? ccccc ? ccccc}
        \toprule

        \multicolumn{1}{l}{\multirow{2.5}{*}{\normalsize \textbf{Method}}} &
        \multicolumn{5}{c}{\small \textbf{Bedroom}}                        &
        \multicolumn{5}{c}{\small \textbf{Livingroom}}                     &
        \multicolumn{5}{c}{\small \textbf{Diningroom}}                                                                                                                             \\

        \cmidrule(lr){2-6}
        \cmidrule(lr){7-11}
        \cmidrule(lr){12-16}

        \multicolumn{1}{l}{}

                                                                           & \headerNoUnit{FID}
                                                                           & \header{KID}{\times 10^{-3}}
                                                                           & \header{SCA}{\%}
                                                                           & \header{TKL}{\times 10^{-2}}
                                                                           & \multicolumn{1}{c}{\header{IoU}{\%}}

                                                                           & \headerNoUnit{FID}
                                                                           & \header{KID}{\times 10^{-3}}
                                                                           & \header{SCA}{\%}
                                                                           & \header{TKL}{\times 10^{-2}}
                                                                           & \multicolumn{1}{c}{\header{IoU}{\%}}

                                                                           & \headerNoUnit{FID}
                                                                           & \header{KID}{\times 10^{-3}}
                                                                           & \header{SCA}{\%}
                                                                           & \header{TKL}{\times 10^{-2}}
                                                                           & \header{IoU}{\%}                                                                                      \\
        \midrule

        LayoutGPT-GPT-4o~\shortcite{feng2023layoutgpt}
                                                                           & 20.42                                & 6.45          & 75.68          & 7.12          & 0.56
                                                                           & 27.88                                & 9.88          & 75.24          & 8.67          & 0.43
                                                                           & N/A                                  & N/A           & N/A            & N/A           & N/A           \\

        LayoutGPT-LLaMA3.1~\shortcite{feng2023layoutgpt}
                                                                           & 29.76                                & 10.93         & 82.31          & 18.14         & 1.46
                                                                           & 38.20                                & 14.50         & 86.22          & 23.66         & 1.55
                                                                           & N/A                                  & N/A           & N/A            & N/A           & N/A           \\

        ATISS~\shortcite{paschalidou2021atiss}
                                                                           & 18.77                                & 2.60          & 64.54          & 0.96          & 0.67
                                                                           & 20.12                                & 3.61          & 68.07          & \second{1.60} & 1.45
                                                                           & 21.07                                & 3.03          & 66.49          & 1.83          & 1.71          \\

        DiffuScene~\shortcite{tang2024diffuscene}
                                                                           & 18.48                                & 2.73          & 62.62          & 1.53          & 1.16
                                                                           & 24.62                                & 8.39          & 76.12          & 1.65          & 0.88
                                                                           & 25.78                                & 7.71          & 68.22          & 2.49          & 1.29          \\

        GLTScene~\shortcite{li2024gltscene}
                                                                           & 19.80                                & 2.93          & 65.34          & \best{0.81}   & 0.62
                                                                           & 22.01                                & 7.26          & 74.47          & 2.25          & 0.95
                                                                           & 20.94                                & 5.66          & 67.79          & 2.33          & 1.83          \\

        \midrule

        \name (Ours)
                                                                           & \second{17.53}                       & \best{1.28}   & \best{59.62}   & \second{0.95} & \second{0.55}
                                                                           & \best{18.14}                         & \best{1.95}   & \second{65.53} & \best{1.15}   & \second{0.40}
                                                                           & \best{18.61}                         & \best{1.96}   & \second{64.86} & \best{1.61}   & \best{0.43}   \\

        \name-G (Ours)
                                                                           & \best{17.17}                         & \second{1.50} & \second{60.51} & 1.02          & \best{0.38}
                                                                           & \second{18.51}                       & \second{2.38} & \best{63.88}   & \best{1.15}   & \best{0.36}
                                                                           & \second{18.66}                       & \second{2.60} & \best{62.47}   & \second{1.68} & \second{0.44} \\
        \bottomrule
    \end{tabular}
\end{table*}
\begin{table*}[t]
    \centering
    \caption{Comparison of our method with existing approaches on layout generation without floor plan. Both our general model and the model trained without floor plan data achieve superior performance compared to other methods including ATISS~\cite{paschalidou2021atiss}, DiffuScene~\cite{tang2024diffuscene}, and InstructScene~\cite{lin2024instructscene}, across various metrics.}
    \label{tab:comp_wo_floor}

    \footnotesize

    \setlength{\tabcolsep}{2.0pt}
    \renewcommand{\arraystretch}{1.35}

    \begin{tabular}{l ? ccccc ? ccccc ? ccccc}
        \toprule

        \multicolumn{1}{l}{\multirow{2.5}{*}{\normalsize \textbf{Method}}} &
        \multicolumn{5}{c}{\small \textbf{Bedroom}}                        &
        \multicolumn{5}{c}{\small \textbf{Livingroom}}                     &
        \multicolumn{5}{c}{\small \textbf{Diningroom}}                                                                                                                             \\

        \cmidrule(lr){2-6}
        \cmidrule(lr){7-11}
        \cmidrule(lr){12-16}

        \multicolumn{1}{l}{}

                                                                           & \headerNoUnit{FID}
                                                                           & \header{KID}{\times 10^{-3}}
                                                                           & \header{SCA}{\%}
                                                                           & \header{TKL}{\times 10^{-2}}
                                                                           & \multicolumn{1}{c}{\header{IoU}{\%}}

                                                                           & \headerNoUnit{FID}
                                                                           & \header{KID}{\times 10^{-3}}
                                                                           & \header{SCA}{\%}
                                                                           & \header{TKL}{\times 10^{-2}}
                                                                           & \multicolumn{1}{c}{\header{IoU}{\%}}

                                                                           & \headerNoUnit{FID}
                                                                           & \header{KID}{\times 10^{-3}}
                                                                           & \header{SCA}{\%}
                                                                           & \header{TKL}{\times 10^{-2}}
                                                                           & \header{IoU}{\%}                                                                                      \\
        \midrule

        ATISS~\shortcite{paschalidou2021atiss}
                                                                           & 20.71                                & 2.01          & 65.00          & 0.87          & 0.48
                                                                           & 25.54                                & 3.07          & 71.87          & 0.94          & 0.58
                                                                           & 24.72                                & 3.24          & 60.87          & 0.86          & 1.13          \\

        DiffuScene~\shortcite{tang2024diffuscene}
                                                                           & 19.75                                & 2.13          & 62.90          & 0.71          & 0.46
                                                                           & 23.94                                & 2.69          & 64.68          & 0.88          & 0.56
                                                                           & 24.07                                & 2.74          & 65.70          & 0.90          & 0.75          \\

        InstructScene~\shortcite{lin2024instructscene}
                                                                           & 20.03                                & \second{1.74} & \best{56.36}   & 0.36          & 0.49
                                                                           & 22.82                                & 2.03          & 61.05          & \second{0.39} & 0.53
                                                                           & 21.76                                & 1.91          & 64.30          & \best{0.52}   & 0.61          \\

        \midrule

        \name (Ours)
                                                                           & \best{18.85}                         & \best{1.35}   & \second{56.44} & \best{0.26}   & \second{0.35}
                                                                           & \best{21.67}                         & \best{1.06}   & \best{58.33}   & \best{0.24}   & \best{0.37}
                                                                           & \best{19.69}                         & \best{0.66}   & \second{62.62} & 0.60          & \best{0.45}   \\

        \name-G (Ours)
                                                                           & \second{19.27}                       & 1.99          & 59.46          & \second{0.31} & \best{0.33}
                                                                           & \second{22.05}                       & \second{1.88} & \second{60.16} & \second{0.39} & \second{0.39}
                                                                           & \second{19.85}                       & \second{0.92} & \best{61.59}   & \second{0.56} & \second{0.50} \\
        \bottomrule
    \end{tabular}
\end{table*}

\paragraph{Evaluation metrics} Following existing works, we assess generation quality using the following metrics:
\begin{enumerate}[leftmargin=*]
    \item[-] \textbf{Fréchet Inception Distance (FID)~\cite{heusel2017gans} \& Kernel Inception Distance (KID)~\cite{binkowski2018demystifying}}: Measure the distribution distance between generated and ground-truth layouts using rendered images, reflecting distributional similarity and diversity. Layouts are rendered from a top view at $256\times256$ resolution, excluding architectural elements, with furniture textured using predefined semantic colors.
    \item[-] \textbf{Type Kullback--Leibler Divergence (TKL)~\cite{paschalidou2021atiss}}: Computes the KL-divergence between generated and ground-truth furniture type distributions, evaluating type and count accuracy.
    \item[-] \textbf{Scene Classification Accuracy (SCA)~\cite{paschalidou2021atiss}}: Measures classification accuracy of generated layouts using an image classifier trained on rendered ground-truth and synthesized layouts.
    \item[-] \textbf{OBB Intersection over Union (IoU)~\cite{tang2024diffuscene}}: Quantifies spatial overlap between predicted and ground-truth OBBs.
\end{enumerate}
Following DiffuScene~\cite{tang2024diffuscene}, FID, KID, and TKL are computed using both the training and test datasets as ground truth. Additional evaluation in terms of physical plausibility and layout diversity is provided in \arxiv{\zcref{subsec:pc}}{Suppl.~Sec.~3.1} and \arxiv{\zcref{subsec:div}}{Suppl.~Sec.~3.4}, respectively.

\paragraph{Baselines} Given the varying training conditions of existing methods, we compare \name in two scenarios: with floor-plan control and without it.
\begin{enumerate}[leftmargin=*]
    \item[-] \textbf{With floor plan}:
          3D-FRONT-trained models, including ATISS \cite{paschalidou2021atiss}, GLTScene~\cite{li2024gltscene}, and DiffuScene~\cite{tang2024diffuscene} , which utilize floor plan data for training, as well as LLM-based methods such as LayoutGPT~\cite{feng2023layoutgpt}. For LayoutGPT, we substitute the original backend with GPT-4o~\cite{achiam2023gpt} and LLaMA-3.1-8B~\cite{grattafiori2024llama} for comparison against modern reasoning standards.
    \item[-] \textbf{Without floor plan}:
          ATISS~\cite{paschalidou2021atiss}, DiffuScene \cite{tang2024diffuscene}, and InstructScene~\cite{lin2024instructscene}. All methods are trained without floor plan conditioning.
\end{enumerate}
In \arxiv{\zcref{subsec:layoutvlm}}{Suppl.~Sec.~3.2}, we further compare \name with the latest VLM-based method --- LayoutVLM~\cite{sun2025layoutvlm}, both quantitatively and qualitatively.

\subsection{Evaluation} \label{subsec:quantitative}

We evaluate \name on three tasks---scene generation, layout completion, and furniture rearrangement---under two settings: with and without floor-plan control. In the floor-plan-controlled setting, 2D floor plans from the test set are provided as conditional inputs. For FID, KID, and SCA evaluations, both furniture items and the conditioned floor plans are rendered. In the without-floor-plan setting, scenes are generated without floor plan constraints, and only furniture items are rendered for metric computation.

\paragraph{Layout Synthesis}
The primary application of \name is synthesizing plausible furniture layouts for empty rooms. Quantitative comparisons under various metrics are reported in \zcref{tab:comp_w_floor} and \zcref{tab:comp_wo_floor}. Under both settings (with and without floor plan conditioning), our method consistently outperforms all baselines across FID, KID, TKL, and SCA. Notably, our unified model achieves strong performance across all tasks without retraining, unlike competing methods, highlighting the robustness of our pipeline design and data augmentation strategy.
\begin{figure*}[t]
    \centering
    \includegraphics[width=0.96\textwidth]{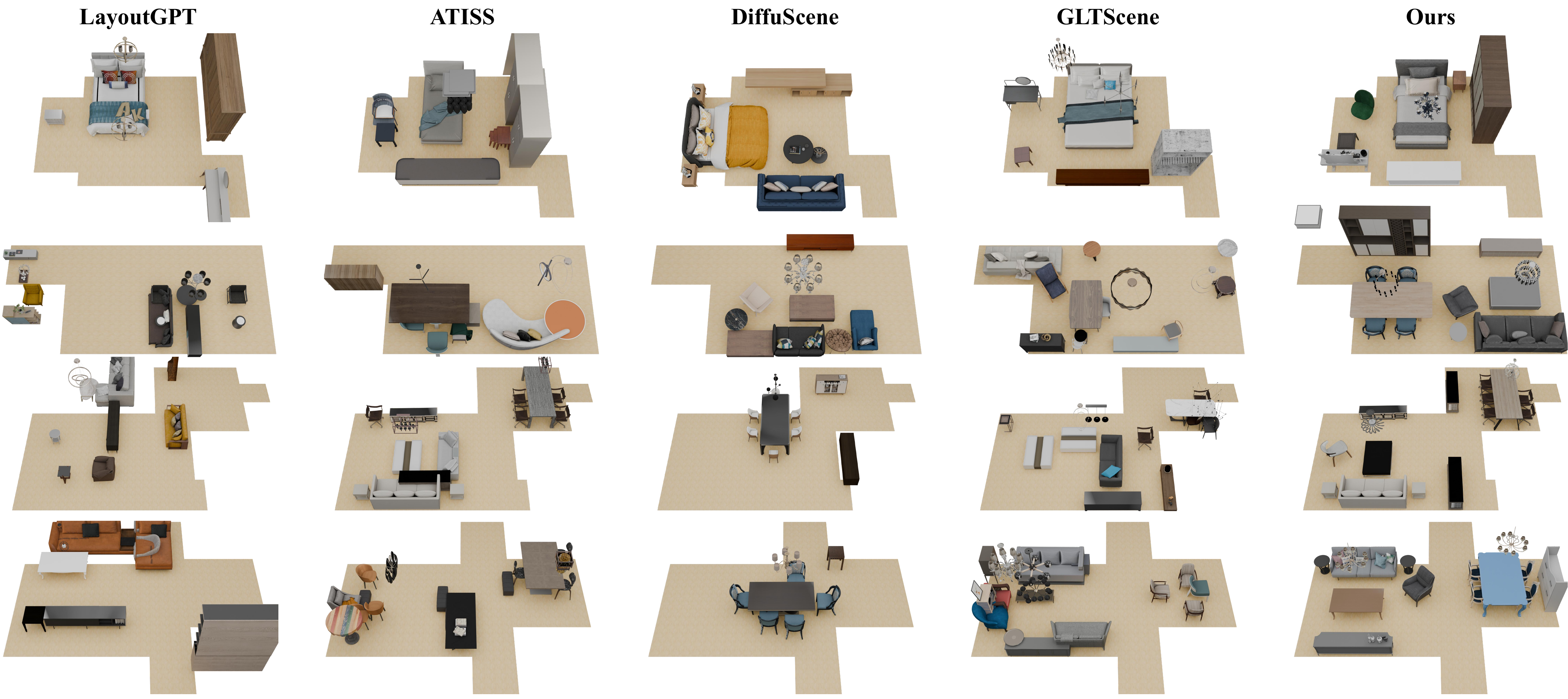}

    \caption{Comparison with LayouGPT~\cite{feng2023layoutgpt}, ATISS~\cite{paschalidou2021atiss}, DiffuScene~\cite{tang2024diffuscene}, and GLTScene~\cite{li2024gltscene} with floor plan condition. Our generated results almost completely avoid incorrect overlaps and out-of-bound placements, while making full use of the given space.}
    \label{fig:comp_w_floor}
\end{figure*}
\begin{figure*}[t]
    \centering
    \includegraphics[width=0.85\textwidth]{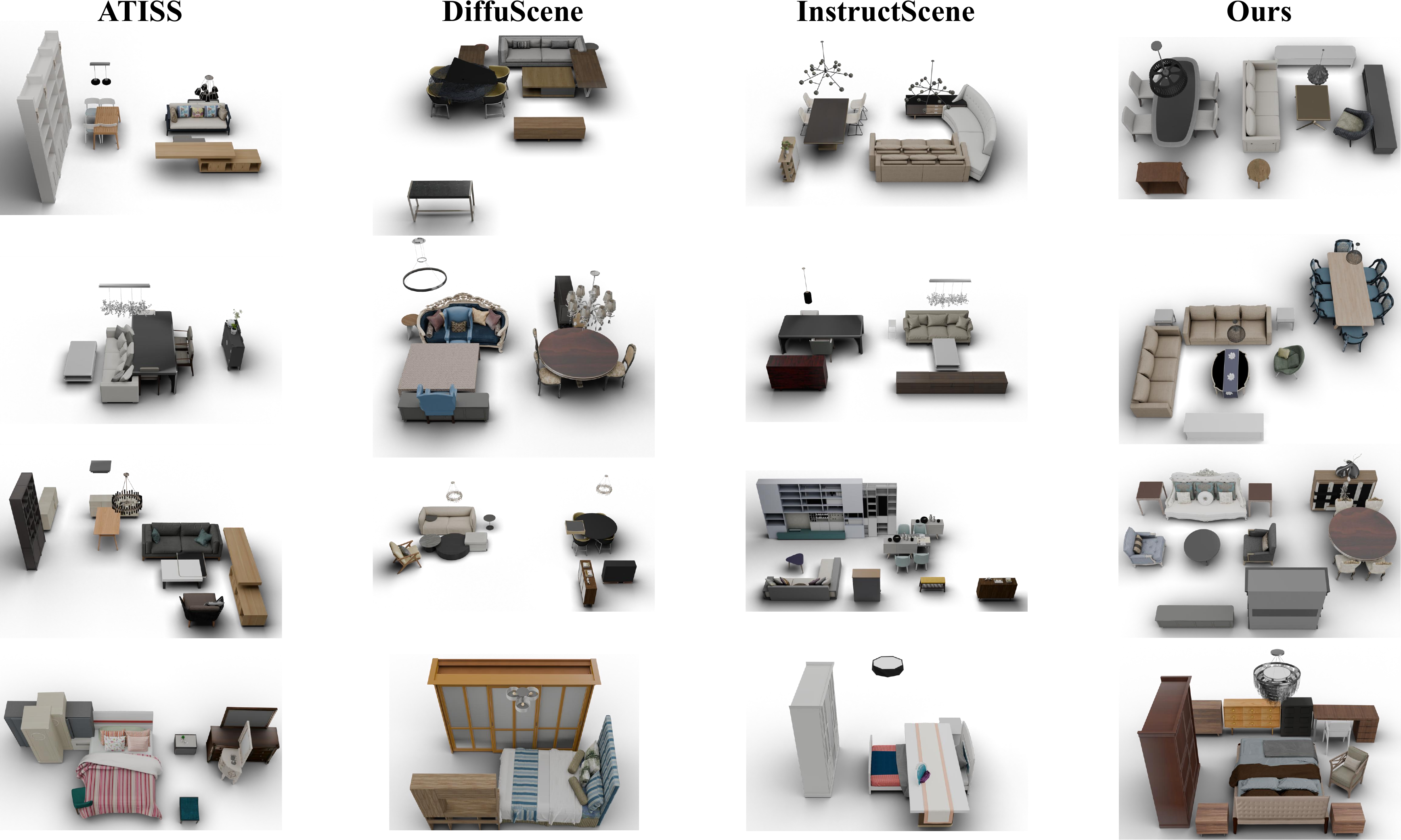}

    \caption{Qualitative comparison with SOTA methods, including ATISS~\cite{paschalidou2021atiss}, DiffuScene~\cite{tang2024diffuscene}, and InstructScene~\cite{lin2024instructscene}. Our method demonstrates better local plausibility and clearer functional zoning.}
    \label{fig:comp_wo_floor}
\end{figure*}

Qualitative results with floor plan conditioning are shown in \zcref{fig:comp_w_floor}. Learning-based methods often produce overly compact placements in complex floor plans, leading to overlaps or underutilized space. LLM-based methods struggle to handle complex room geometries, frequently violating boundary constraints and placing objects out of bounds. In contrast, \name tends to utilize available space more effectively while maintaining local coherence. Results without floor plan conditioning are shown in \zcref{fig:comp_wo_floor}. Our sparse relation modeling reduces spurious interactions while preserving meaningful relationships within and across functional zones. Consequently, even without floor plans, generated layouts tend to exhibit distinguishable functional areas (\eg dining versus living regions) with semantically consistent object placement. Additional results are provided in \arxiv{\zcref{subsec:mr}}{Suppl.~Sec.~3.6}.

We conduct a user study with 36 participants on two tasks: generation with and without floor plans. For the ``With Floor Plan'' task, each participant evaluated 15 cases (7 bedrooms and 8 living rooms) sampled from a pool of \num{2000} generated samples. Comparing \name against LayoutGPT~\cite{feng2023layoutgpt}, ATISS~\cite{paschalidou2021atiss}, DiffuScene~\cite{tang2024diffuscene}, and GLTScene~\cite{li2024gltscene} under identical floor plans, participants selected the best layout based on rationality (\eg collision avoidance) and floor plan consistency (\eg boundary respect). For the ``Without Floor Plan'' task, each participant evaluated 15 cases (5 each for bedrooms, living rooms, and dining rooms) against ATISS~\cite{paschalidou2021atiss}, DiffuScene~\cite{tang2024diffuscene}, and InstructScene~\cite{lin2024instructscene}, focusing on layout rationality. As shown in \zcref{tab:user}, \name receives substantially higher user preference than competing methods in both scenarios.

\begin{table}[t]
    \centering
    \caption{User study results across different room types under w/wo floor plan settings. The results demonstrate that our generated scenes align significantly better with user preferences.}
    \label{tab:user}

    \footnotesize
    \setlength{\tabcolsep}{1pt}
    \renewcommand{\arraystretch}{1.2}

    \begin{tabular}{l | cc}
        \toprule
        \small \textbf{Method}
                           & \small \textbf{With Floor}
                           & \small \textbf{Without Floor}                      \\
        \midrule

        LayoutGPT          & $0.74\%$                      & N/A                \\
        ATISS              & $19.07\%$                     & $9.63\%$           \\
        DiffuScene         & $6.67\%$                      & $10.00\%$          \\
        InstructScene      & N/A                           & $17.96\%$          \\
        GLTScene           & $9.44\%$                      & N/A                \\
        \midrule

        CasLayout-G (Ours) & $\mathbf{64.07\%}$            & $\mathbf{62.41\%}$ \\
        \bottomrule
    \end{tabular}
\end{table}
\begin{table}[t]
  \centering
  \caption{Relational consistency between the output results and control conditions across different room types under w/wo floor plan settings, given an object list and a corresponding sparse relationship graph as inputs.}
  \label{tab:rela_acc}

  \footnotesize
  \setlength{\tabcolsep}{3pt}
  \renewcommand{\arraystretch}{1.2}

  \begin{tabular}{ll | cc | cc | cc}
    \toprule

    \multicolumn{2}{c}{\small \textbf{Room Type}}  &
    \multicolumn{2}{c}{\small \textbf{Bedroom}}    &
    \multicolumn{2}{c}{\small \textbf{Livingroom}} &
    \multicolumn{2}{c}{\small \textbf{Diningroom}}                                                                                           \\

    \cmidrule(lr){1-2}
    \cmidrule(lr){3-4}
    \cmidrule(lr){5-6}
    \cmidrule(lr){7-8}

    \multicolumn{2}{c}{\textbf{Floor Condition}}
                                                   & \multicolumn{1}{c}{\textbf{\Checkmark}}
                                                   & \multicolumn{1}{c}{\textbf{\XSolid}}
                                                   & \multicolumn{1}{c}{\textbf{\Checkmark}}
                                                   & \multicolumn{1}{c}{\textbf{\XSolid}}
                                                   & \multicolumn{1}{c}{\textbf{\Checkmark}}
                                                   & \multicolumn{1}{c}{\textbf{\XSolid}}                                                    \\
    \midrule

    \multirow{4}{*}{\textbf{Obj. to Obj.}}
                                                   & Direction                               & 89.57 & 92.49 & 85.98 & 88.62 & 86.53 & 88.03 \\
                                                   & Distance                                & 91.81 & 93.48 & 88.87 & 90.07 & 89.66 & 90.82 \\
                                                   & Alignment                               & 93.02 & 94.35 & 90.11 & 90.84 & 91.03 & 92.19 \\
                                                   & Symmetry                                & 94.92 & 95.08 & 92.93 & 93.04 & 92.56 & 92.96 \\
    \midrule

    \textbf{Obj. to Wall}
                                                   & Distance                                & 89.89 & N/A   & 88.42 & N/A   & 88.91 & N/A   \\
    \midrule

    \textbf{Obj. to Door/Win.}
                                                   & Distance                                & 94.80 & N/A   & 91.67 & N/A   & 92.08 & N/A   \\
    \bottomrule
  \end{tabular}
\end{table}

\begin{table}[t]
    \centering
    \caption{Comparisons of rearrangement and completion tasks guided by floor plan constraints, where our method significantly outperforms baseline approaches across various metrics.}
    \label{tab:comp_app_w_floor}

    \footnotesize
    \setlength{\tabcolsep}{3pt}
    \renewcommand{\arraystretch}{1.2}

    \begin{tabular}{l ? cccc ? ccccc}
        \toprule

        \multicolumn{1}{l}{\multirow{2.5}{*}{\normalsize \textbf{Method}}} &
        \multicolumn{4}{c}{\small \textbf{Rearrangement}}                  &
        \multicolumn{5}{c}{\small \textbf{Completion}}                                                                                                                         \\

        \cmidrule(lr){2-5}
        \cmidrule(lr){6-10}

        \multicolumn{1}{l}{}

                                                                           & \textbf{FID}
                                                                           & \textbf{KID}
                                                                           & \textbf{SCA}
                                                                           & \multicolumn{1}{c}{\textbf{IoU}}                                                                  

                                                                           & \textbf{FID}
                                                                           & \textbf{KID}
                                                                           & \textbf{SCA}
                                                                           & \textbf{TKL}
                                                                           & \textbf{IoU}                                                                                      \\
        \midrule

        DiffuScene
                                                                           & 28.60                            & 7.23          & 72.87          & 1.69
                                                                           & 23.04                            & 5.85          & 76.93          & 1.56          & 0.85          \\
        \midrule

        Ours
                                                                           & \best{18.34}                     & \best{1.03}   & \second{63.17} & \second{0.87}
                                                                           & \best{16.50}                     & \best{1.10}   & \second{63.80} & \second{1.25} & \second{0.50} \\

        Our-G
                                                                           & \second{18.97}                   & \second{2.04} & \best{62.41}   & \best{0.80}
                                                                           & \second{16.71}                   & \second{1.74} & \best{63.63}   & \best{1.08}   & \best{0.46}   \\
        \bottomrule
    \end{tabular}
\end{table}
\begin{table}[t]
    \centering
    \caption{Rearrangement and completion without floor plan, where our approach achieves superior results compared to existing methods.}
    \label{tab:comp_app_wo_floor}

    \footnotesize
    \setlength{\tabcolsep}{3.0pt}
    \renewcommand{\arraystretch}{1.2}

    \begin{tabular}{l ? cccc ? ccccc}
        \toprule

        \multicolumn{1}{l}{\multirow{2.5}{*}{\normalsize \textbf{Method}}} &
        \multicolumn{4}{c}{\small \textbf{Rearrangement}}                  &
        \multicolumn{5}{c}{\small \textbf{Completion}}                                                                                                                         \\

        \cmidrule(lr){2-5}
        \cmidrule(lr){6-10}

        \multicolumn{1}{l}{}

                                                                           & \textbf{FID}
                                                                           & \textbf{KID}
                                                                           & \textbf{SCA}
                                                                           & \multicolumn{1}{c}{\textbf{IoU}}                                                                  

                                                                           & \textbf{FID}
                                                                           & \textbf{KID}
                                                                           & \textbf{SCA}
                                                                           & \textbf{TKL}
                                                                           & \textbf{IoU}                                                                                      \\
        \midrule

        DiffuScene
                                                                           & 26.35                            & 6.40          & 72.31          & 0.78
                                                                           & 33.85                            & 6.89          & 75.38          & 2.67          & 0.61          \\

        InstructScene
                                                                           & 24.54                            & 3.79          & 67.94          & \second{0.50}
                                                                           & 25.25                            & 3.74          & 68.52          & 0.52          & \second{0.37} \\

        \midrule

        Ours
                                                                           & \second{22.65}                   & \second{2.27} & \best{63.28}   & \second{0.50}
                                                                           & \second{23.16}                   & \second{2.89} & \second{66.96} & \second{0.39} & \best{0.36}   \\

        Our-G
                                                                           & \best{22.41}                     & \best{2.11}   & \second{64.57} & \best{0.47}
                                                                           & \best{21.11}                     & \best{1.78}   & \best{64.19}   & \best{0.38}   & \best{0.36}   \\
        \bottomrule
    \end{tabular}
\end{table}
\begin{figure}[t]
    \centering
    \includegraphics[width=0.9\linewidth]{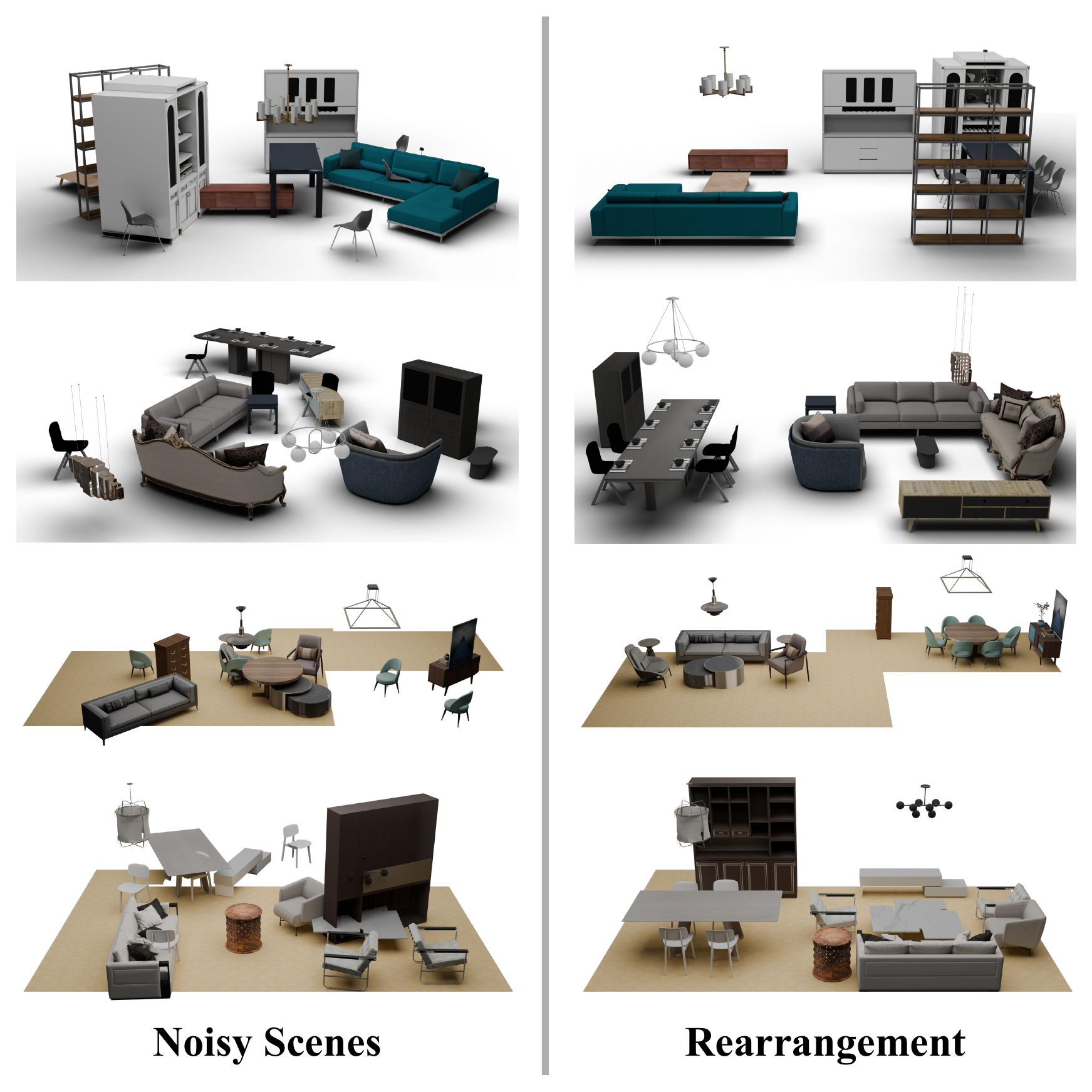}
    \caption{Qualitative comparisons of scene rearrangement are: the upper two rows correspond to completions without floor plan input, and the lower two rows correspond to completions with floor plan guidance.}
    \label{fig:app_r}
\end{figure}
\begin{figure}[t]
    \centering
    \includegraphics[width=0.85\linewidth]{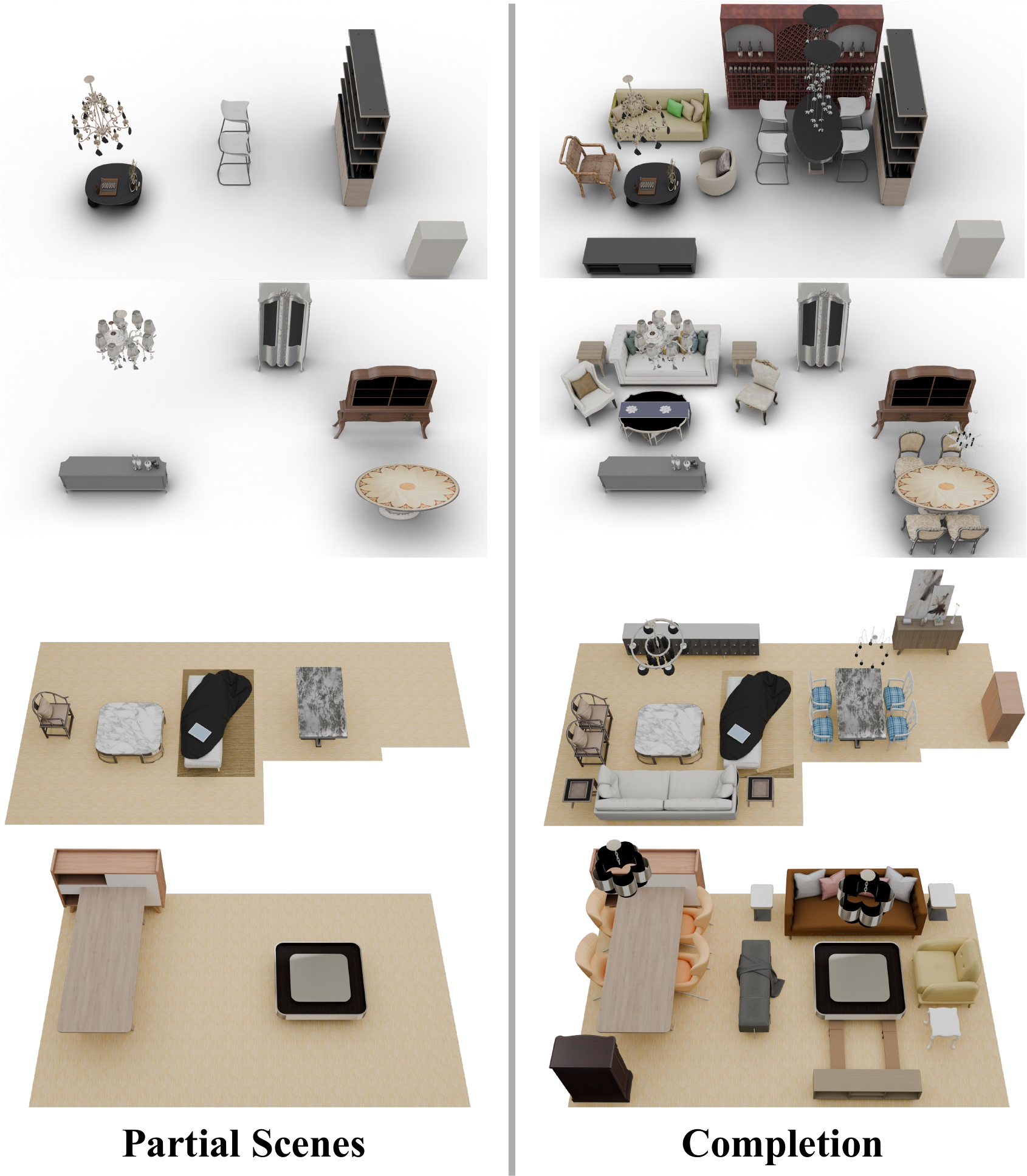}

    \caption{Qualitative comparison of completion results, where the first two rows depict completions without floor plan guidance, and the last two rows show completions guided by floor plan information.}
    \label{fig:app_c}
\end{figure}

Furthermore, we analyze satisfaction rates for different relation categories across room types, both with and without floor plan constraints. As shown in \zcref{tab:rela_acc}, our method achieves relation compliance rates of approximately $90\%$ or higher across diverse environments, indicating strong controllability.


\paragraph{Furniture rearrangement}
For rearrangement, we randomly initialize furniture placements within rooms and apply \name and competing methods to reorganize the scenes. Rearranged layouts are sourced from the test dataset and compared against publicly available implementations of this application. Leveraging our modular pipeline, we execute only the final two diffusion stages, enabling efficient rearrangement. Quantitative results in \zcref{tab:comp_app_w_floor} and \zcref{tab:comp_app_wo_floor} show that \name outperforms all baselines under both with-floor-plan and without-floor-plan settings. Qualitative examples in \zcref{fig:app_r} further demonstrate coherent and visually appealing rearrangements, even for complex object configurations.

\paragraph{Layout completion}
For layout completion, partial scenes from the test dataset serve as conditional inputs. Our model infers missing furniture items and predicts their placement. As shown in \zcref{tab:comp_app_w_floor} and \zcref{tab:comp_app_wo_floor}, \name consistently delivers strong performance in both controlled and uncontrolled settings. Qualitative results in \zcref{fig:app_c} illustrate that our method predicts plausible furniture types and produces arrangements that preserve functional coherence.

\paragraph{Inference time}
We evaluate inference acceleration for \name. As shown in \zcref{tab:time}, incorporating recent diffusion sampling techniques~\cite{zhao2023unipc,lu2025dpm} achieves quality comparable to the 1000-step baseline using only 15--25 steps. On a single NVIDIA A6000 GPU, this takes approximately 25--40 milliseconds, effectively enabling real-time performance. Notably, the latency overhead introduced by the four cascaded stages is negligible. More detailed inference-time breakdowns for different applications are provided in \arxiv{\zcref{subsec:tapp}}{Suppl.~Sec.~3.3}.

\begin{table}[t]
    \centering
    \caption{Comparison of average inference time per scene and generation quality across different methods. With the latest inference method, we can generate scenes of comparable quality in a significantly shorter time.}
    \label{tab:time}

    \footnotesize
    \setlength{\tabcolsep}{4pt}
    \renewcommand{\arraystretch}{1.2}

    \scalebox{1.1}{
        \begin{tabular}{l | c | cc}
            \toprule

            \textbf{Sampler (Steps)}
             & \header{Time}{ms}
             & \headerNoUnit{FID}
             & \header{KID}{\times 10^{-3}}                 \\
            \midrule


            DDPM~\shortcite{ho2020denoising} (1000)
             & 1562.50
             & \textbf{18.51}               & \textbf{2.38} \\

            DPM-Solver++~\shortcite{lu2025dpm} (25)
             & 40.95
             & 18.72                        & 2.56          \\

            UniPC~\shortcite{zhao2023unipc} (15)
             & \textbf{25.57}
             & 18.57                        & 2.69          \\
            \bottomrule
        \end{tabular}}
\end{table}

\begin{figure}[t]
    \centering
    \includegraphics[width=\linewidth]{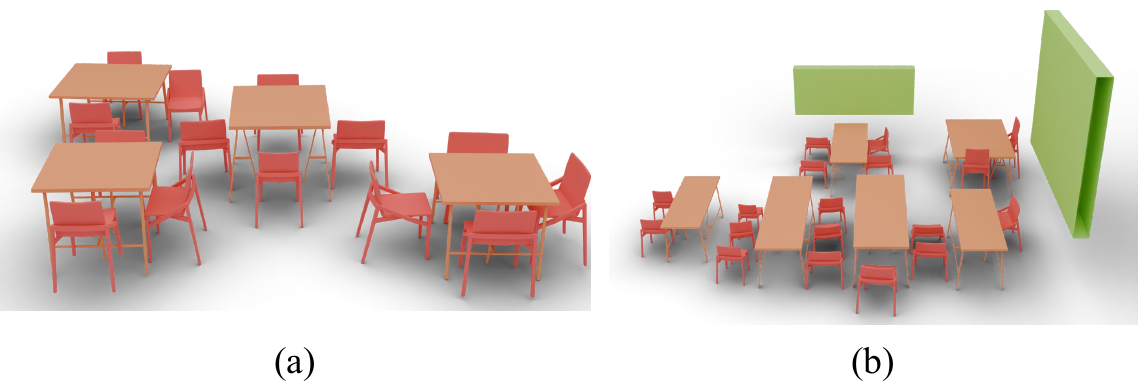}

    \caption{Qualitative results of \name trained on ScanNet layout data for a small cafeteria (a) and a classroom (b). Guided by sparse relationship graphs, \name synthesizes plausible layouts even in scenes with a large number of objects.}
    \label{fig:scannet}
\end{figure}

\paragraph{Generalization to other scene types}
To evaluate generalization beyond 3D-FRONT~\cite{fu20213dfront}, we conduct experiments on ScanNet~\cite{dai2017scannet}. Using layouts extracted from ScanNet, we annotate scene graphs with our relationship extraction method and retrain \name accordingly. As shown in \zcref{fig:scannet}, \name produces plausible layouts with delineated functional zones even in scenes containing more objects, indicating that our relationship formulation and generation pipeline generalize to more complex indoor environments given appropriate training data.

\begin{figure}[t]
    \centering
    \includegraphics[width=\linewidth]{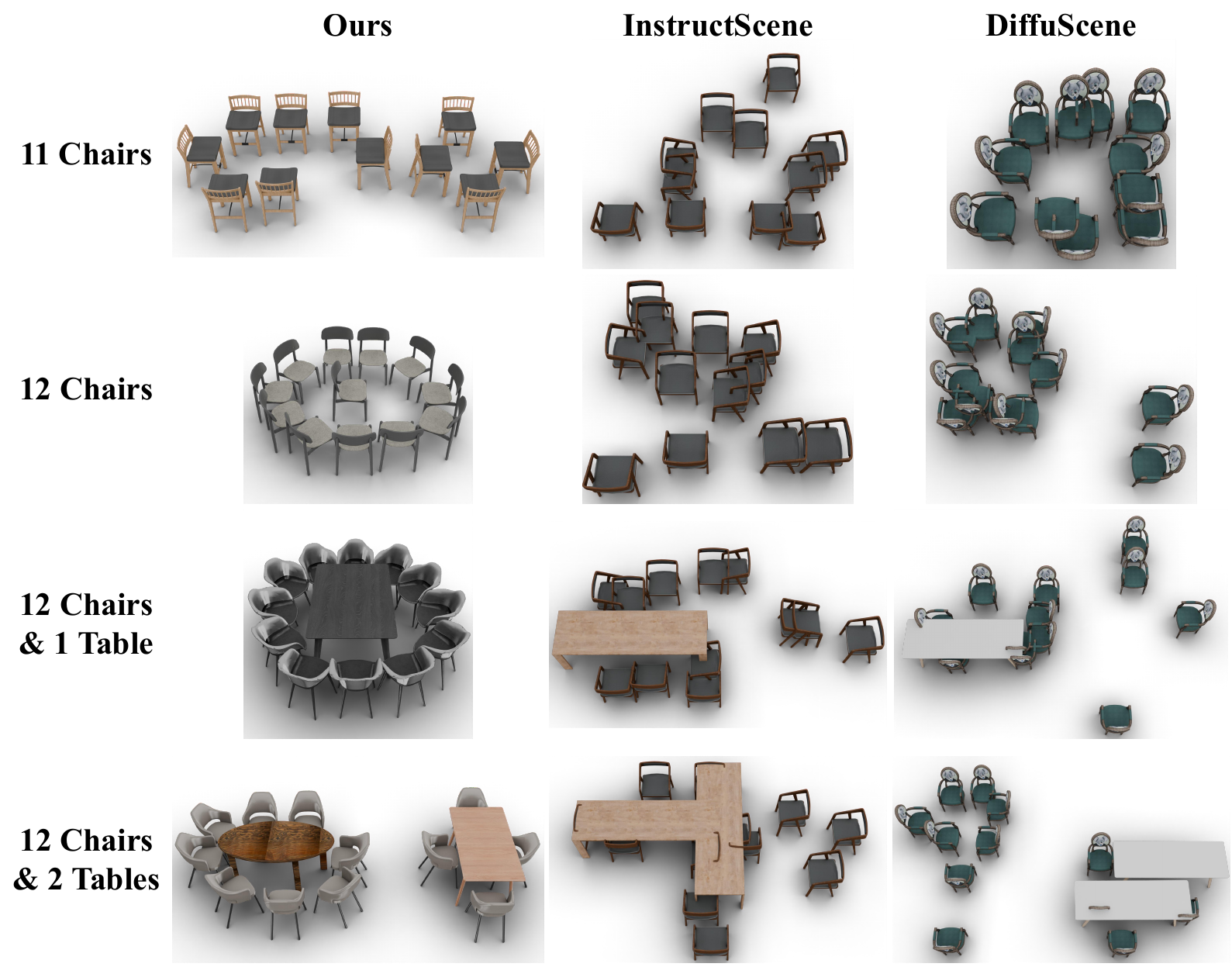}

    \caption{\name is capable of generating plausible scenes based on out-of-distribution object lists. Even when given input containing numbers of chairs and tables that exceed the dataset's maximums, \name produces reasonable and well-arranged results.}
    \label{fig:ood}
\end{figure}
\begin{figure}[t]
    \centering
    \includegraphics[width=0.98\linewidth]{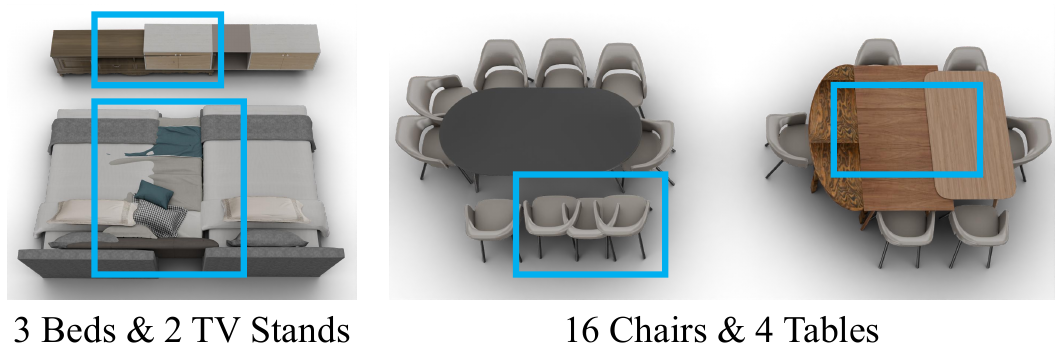}

    \caption{Failure cases where the input object list deviates significantly from the training distribution, \ie excessive furniture count. \name generates unreasonable layouts, with collisions marked by blue bounding boxes.}
    \label{fig:failure_case}
\end{figure}
\paragraph{Out of distribution object list}
We further evaluate generalization under object lists that fall outside the training distribution. For example, although the training set contains only two scenes with 10 chairs (and none with more than 10), we test inputs containing 11 and 12 chairs. Benefiting from robust relation modeling, \name generates plausible layouts, whereas InstructScene~\cite{lin2024instructscene} and DiffuScene~\cite{tang2024diffuscene} often yield disordered arrangements, as shown in \zcref{fig:ood}. When inputs deviate substantially from the training distribution (\eg excessively large furniture counts), \name may still produce unreasonable layouts with collisions, as illustrated in \zcref{fig:failure_case}.

\subsection{More Applications} \label{subsec:application}

\paragraph{Layout editing}

\begin{figure}[t]
    \centering
    \includegraphics[width=0.9\linewidth]{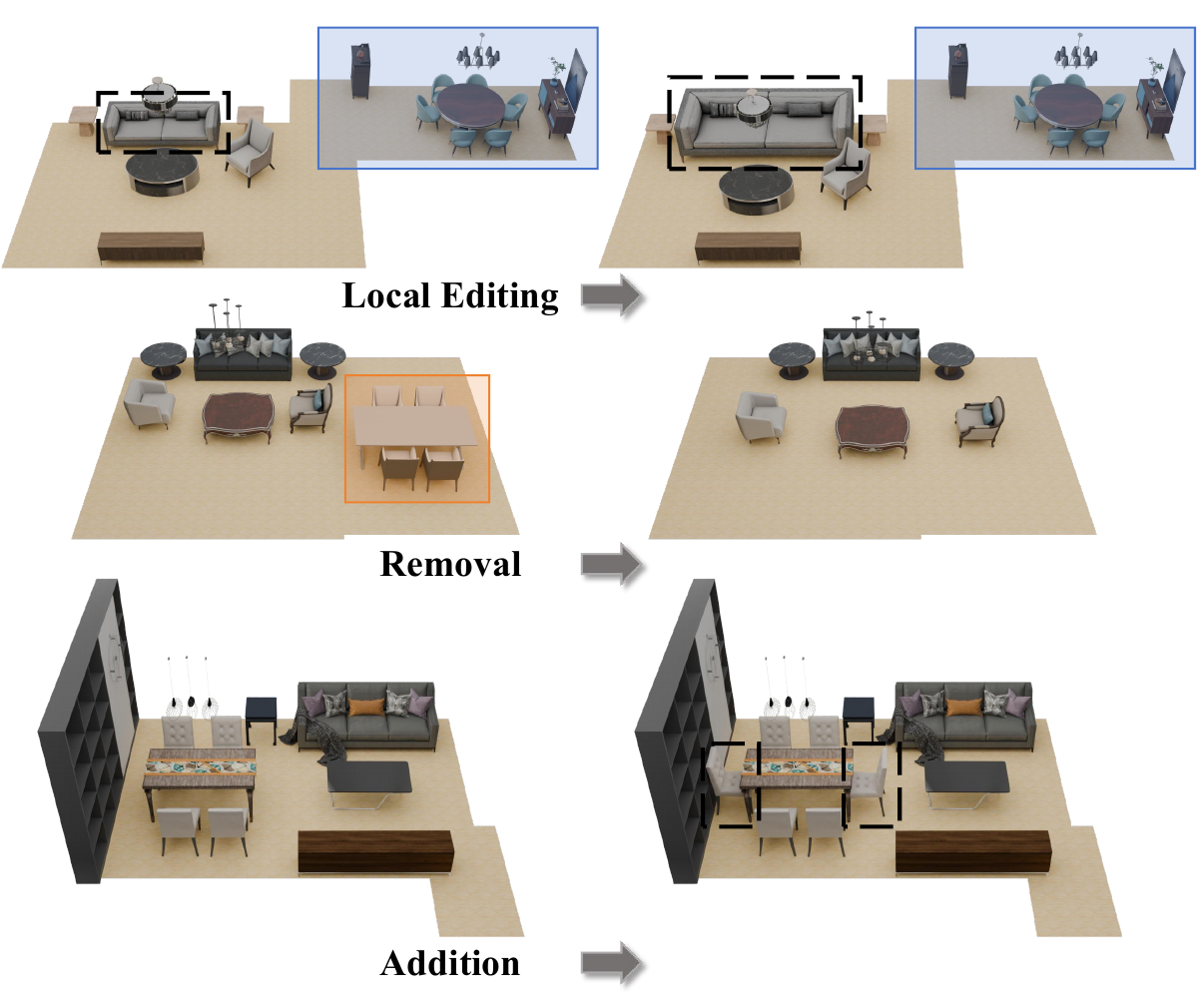}

    \caption{Qualitative examples of editing: (Top) Modifying an object attribute and adjusting its surrounding local layout accordingly, without altering the region within the blue box. (Middle) Removing certain objects (highlighted by the orange box) and adapting the layout to fit the floor plan. (Bottom) Adding new objects while preserving the existing layout.}
    \label{fig:edit}
\end{figure}

\name enables users to interact with the four generation stages for flexible scene editing. Example applications include: modifying a single object and its surrounding local layout while preserving a specified region; removing certain objects and adjusting the remaining layout to accommodate the room structure; and inserting specified objects without altering the existing layout, while maintaining physical plausibility, as shown in \zcref{fig:edit}.

\begin{figure}[t]
    \centering
    \includegraphics[width=0.85\linewidth]{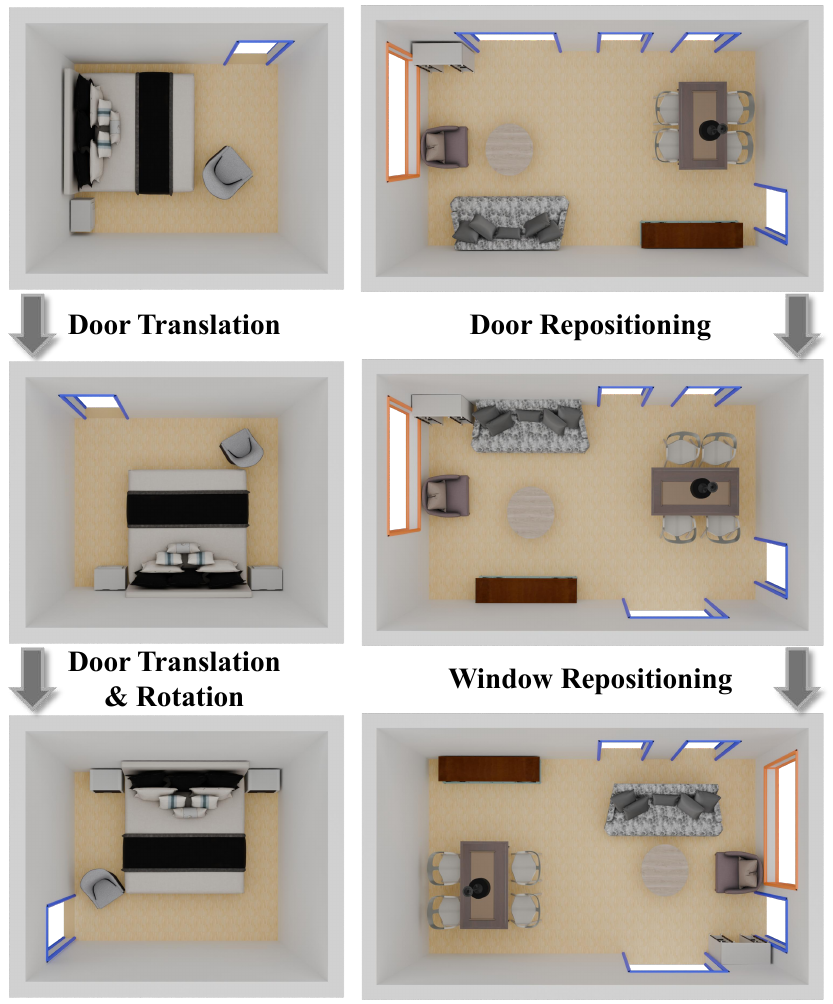}

    \caption{Layout variations in response to door or window changes under fixed control conditions. Doors are marked in blue, and windows in orange.}
    \label{fig:door}
\end{figure}

\paragraph{Architectural element control}
\name explicitly incorporates architectural elements (\eg walls, doors, and windows), allowing generated layouts to respect real-world constraints. As shown in \zcref{fig:door}, when door or window positions are modified while the surrounding floor plan remains unchanged, our system updates the scene accordingly and preserves sufficient walking space behind the modified element. This demonstrates sensitivity to local architectural changes and highlights practical applicability in real-world design workflows.

\begin{figure}[t]
    \centering
    \includegraphics[width=0.9\linewidth]{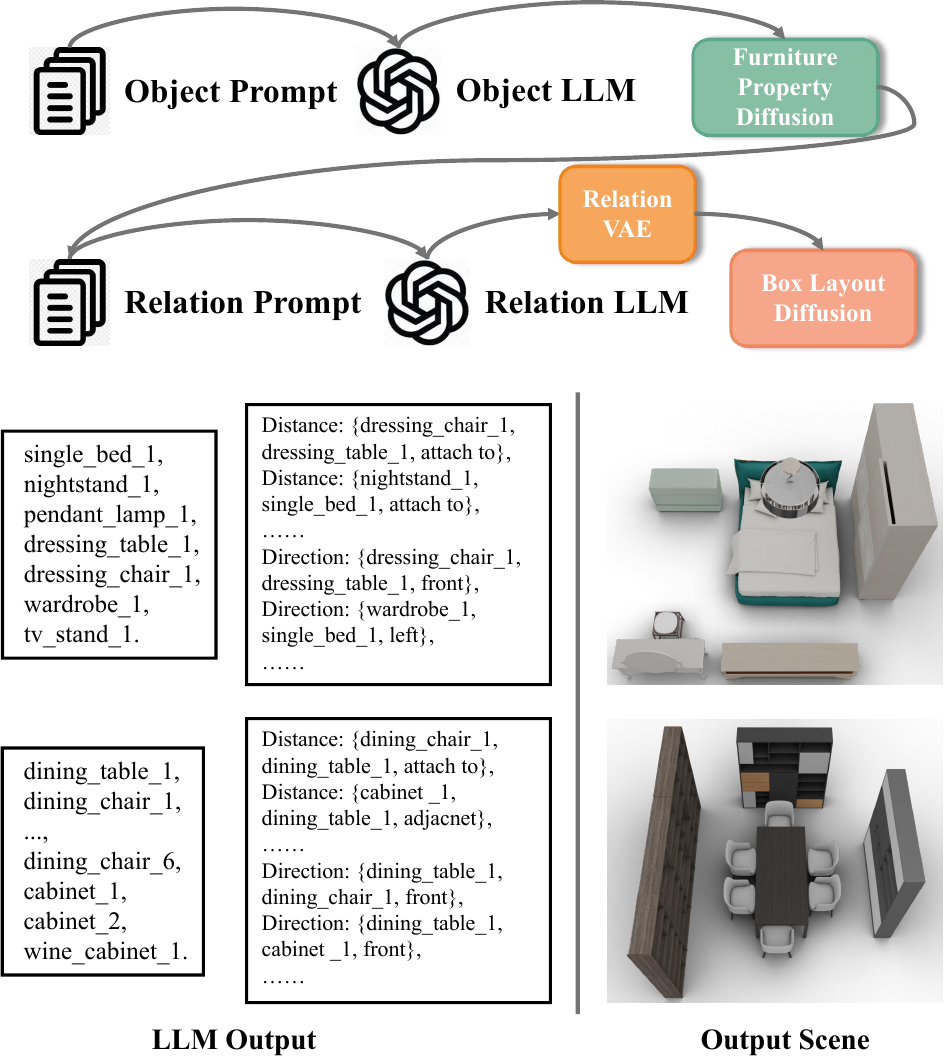}
    \caption{Integration with LLM. When LLM produces coherent and contextually appropriate outputs, \name effectively translates these into spatial arrangements that honor the specified constraints.}
    \label{fig:llm_add}
\end{figure}

\paragraph{LLM-guided scene generation}
Leveraging the flexibility of our cascaded pipeline, we replace the first and third stages---both naturally expressible in language---with outputs from LLMs~\cite{achiam2023gpt}. Under carefully designed prompts, LLMs generate coherent furniture lists and spatial relations, while \name applies learned priors to produce physically plausible layouts. Users provide prompts specifying task descriptions, room types, output formats, and in-context examples. The LLM first outputs a furniture list, which our diffusion model uses to generate sizes and geometric codes. The LLM then predicts key spatial relationships, which are encoded via the relation VAE to guide box layout diffusion. As illustrated in \zcref{fig:llm_add}, this hybrid approach combines the semantic diversity of LLM outputs with the physical realism of diffusion models, demonstrating potential for scalable 3D scene generation under language-based supervision.

\begin{figure}[t]
    \centering
    \includegraphics[width=0.95\linewidth]{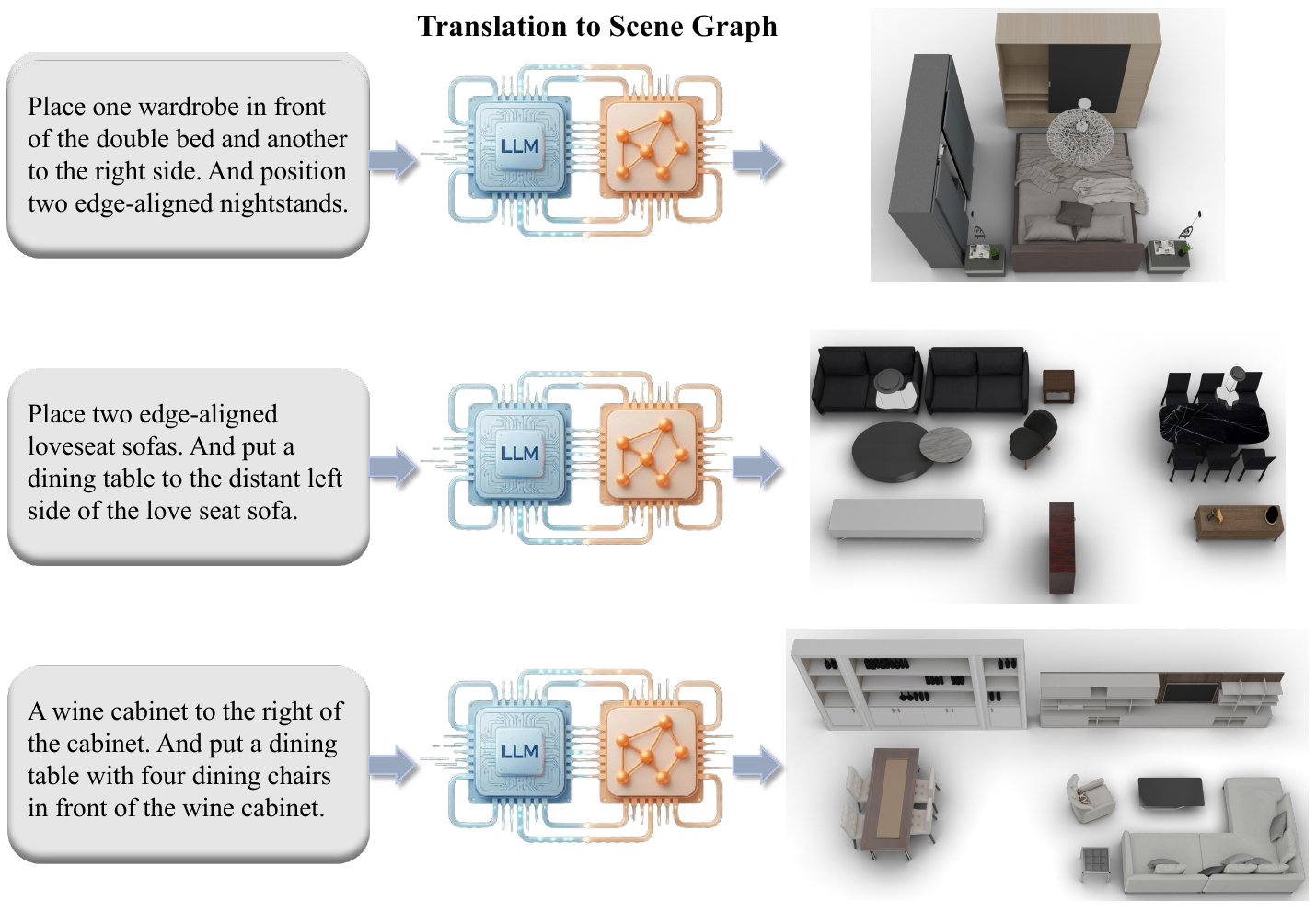}

    \caption{Text-controlled layout generation. The control text is converted into a sparse partial scene graph via an LLM. Conditioned on this translated graph, \name can generate a plausible and complete scene.}
    \label{fig:text_con}
\end{figure}
\paragraph{Text-driven scene generation}
With carefully designed prompts, \name leverages LLMs~\cite{achiam2023gpt} to translate textual input into a sparse scene graph composed of object nodes and inter-object relations. The final scene is synthesized using the sparse graph-conditioned generation mechanism, producing layouts that follow the description while reasonably completing missing elements, as shown in \zcref{fig:text_con}.

\begin{figure}[t]
    \centering
    \includegraphics[width=0.92\linewidth]{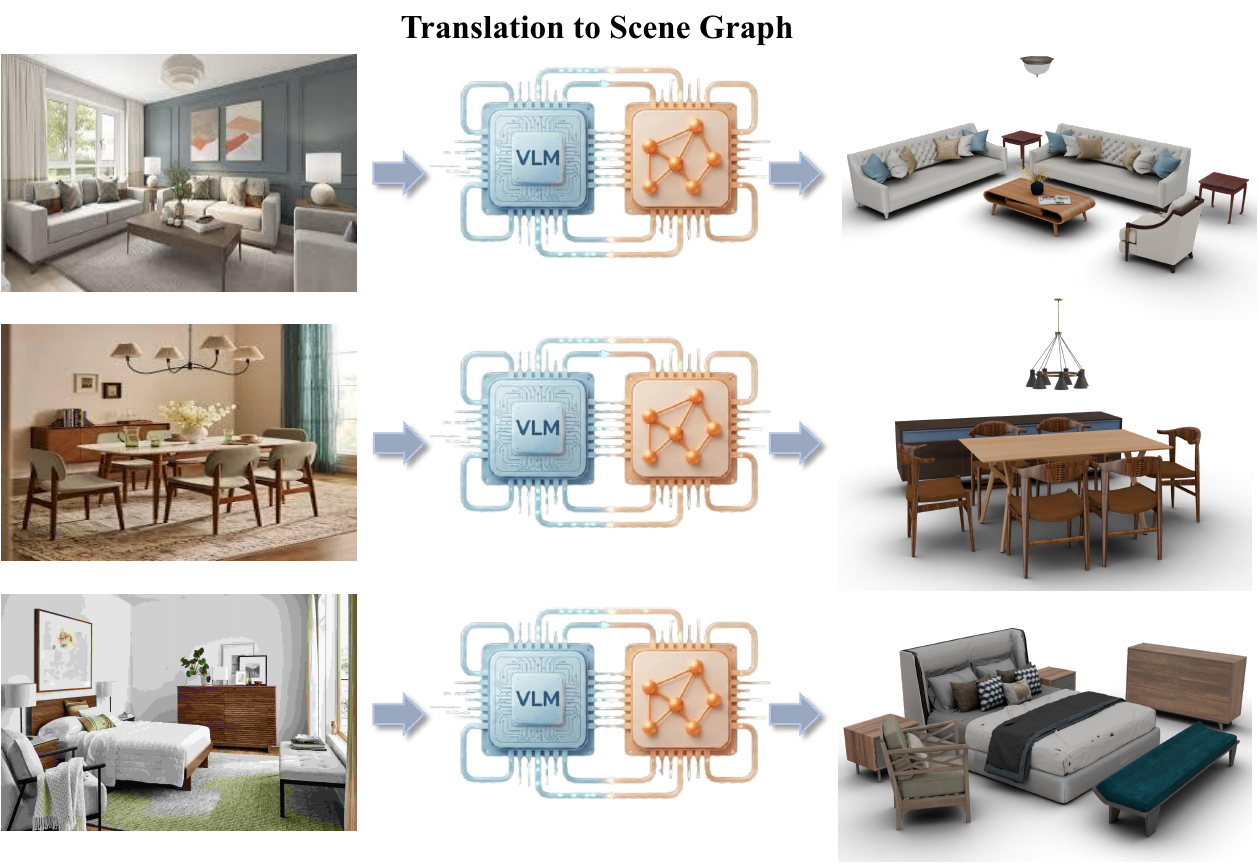}

    \caption{Image-controlled layout generation. The control image is converted into a corresponding scene graph via an VLM. Conditioned on this translated graph, \name can generate a consistent 3D scene.}
    \label{fig:img_con}
\end{figure}
\paragraph{Image-driven scene generation}
Similar to text-driven generation, \name utilizes VLMs~\cite{team2024gemini} to parse objects and relations from input images to condition box layout diffusion (\zcref{fig:img_con}). Unlike the text-driven pipeline, \name performs no extra completion to preserve fidelity to the visual input. Two design choices facilitate this: (1) local relation definitions allow VLMs to predict relations without global coordinate calibration; and (2) sparse graphs reduce complexity, mitigating potential VLM errors induced by dense relations.

\begin{figure*}[h]
    \centering
    \includegraphics[width=0.9\textwidth]{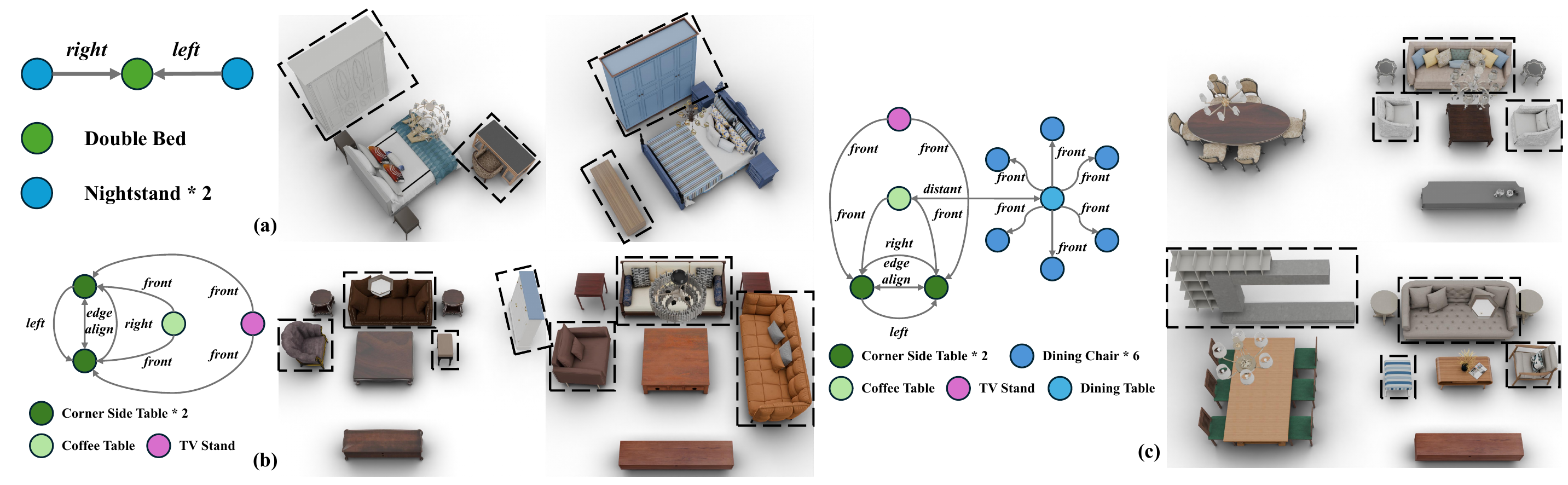}

    \caption{User-provided relation-controlled layout generation. Under varying densities of relational graphs, \name is able to produce diverse results that conform to the given controls and reasonably complete the missing parts (marked by dashed lines) in the conditional graph.}
    \label{fig:sparse_relation}
\end{figure*}
\paragraph{User-specified relation control}
\name employs sparse relation graphs for scene control, aligning with how humans describe object relationships. This design enables user-specified relation control by completing furniture lists and encoding sparse relation edges into relation latents. As shown in \zcref{fig:sparse_relation}, across user-provided relation graphs of varying sparsity levels, \name generates layouts that respect specified relationships while maintaining overall plausibility. The model can also complete missing scene elements and exhibits diversity, producing distinct yet valid arrangements under identical control signals.

\subsection{Ablation Study}
\label{subsec:ablation}
We conduct a series of ablation studies on the living room dataset, chosen for its greater layout diversity, to evaluate key components of our design.

\paragraph{Design of cascaded diffusion procedures}
We compare our four-stage cascaded diffusion pipeline against three alternative designs:
\begin{itemize}[leftmargin=*]
    \item \emph{\uppercase\expandafter{\romannumeral1}-Stage}: A single-stage diffusion model that predicts all layout properties jointly. To implement this, we modify the furniture type diffusion network to add noise to furniture OBBs and feature embeddings during prediction.
    \item \emph{\uppercase\expandafter{\romannumeral2}-Stage}: A two-stage design that first predicts furniture types, followed by OBBs and feature embeddings.
    \item \emph{\uppercase\expandafter{\romannumeral3}-Stage}: A three-stage design that sequentially predicts furniture types, sizes and feature embeddings, translations, and rotations, but omits relation latent modeling.
\end{itemize}
Quantitative results in \zcref{tab:ablation} confirm the effectiveness of our cascaded design. In particular, the comparison with \emph{\uppercase\expandafter{\romannumeral3}-Stage} highlights the importance of relation modeling in achieving high-quality 3D layouts.
\begin{table}[t]
  \centering
  \caption{Ablation studies of our proposed framework containing the cascaded generation procedure, implicit sparse relation modeling, and co-training scheme.}
  \label{tab:ablation}

  \footnotesize
  \setlength{\tabcolsep}{3pt}
  \renewcommand{\arraystretch}{1.2}
  \scalebox{1.1}{
    \begin{tabular}{ll|ccc}
      \toprule
      \multicolumn{2}{c|}{\textbf{Method}}
       & \headerNoUnit{FID}
       & \header{KID}{\times 10^{-3}}
       & \header{SCA}{\%}                                                      \\
      \midrule

      \multirow{3}{*}{\textbf{Cascade}}
       & I-Stage                      & 22.26         & 6.23  & 73.31          \\
       & II-Stage                     & 21.03         & 4.48  & 74.94          \\
       & III-Stage                    & 20.52         & 3.71  & 71.02          \\
      \cmidrule(lr){1-5}

      \textbf{Sequence}
       & Obj.-Rel.-Prop.-OBBs         & 19.04         & 2.59  & \textbf{67.13} \\
      \cmidrule(lr){1-5}

      \multirow{2}{*}{\textbf{Relation}}
       & Dense-Relation               & 20.03         & 4.01  & 69.77          \\
       & Exp-Relation                 & 19.61         & 3.54  & 70.19          \\
      \cmidrule(lr){1-5}

      \textbf{Training}
       & No Co-Training               & 18.95         & 3.07  & 69.05          \\
      \midrule

      \multicolumn{2}{c|}{\textbf{Ours}}
       & \textbf{18.51}               & \textbf{2.38} & 68.88                  \\
      \bottomrule
    \end{tabular}}
\end{table}

\begin{table}[t]
  \centering
  \caption{Ablation studies of our proposed VAE structure, which demonstrate the effectiveness of the in-out attention mechanism.}
  \label{tab:ablation_vae}

  \footnotesize
  \setlength{\tabcolsep}{3pt}
  \renewcommand{\arraystretch}{1.2}

  \scalebox{1.1}{
    \begin{tabular}{ll | c}
      \toprule
      \multicolumn{2}{c|}{\textbf{Method}}
       & $\mathrm{\textbf{Accuracy}}^{\uparrow}_{\mathrm{\mathbf{\%}}\alignphantom}$         \\
      \midrule

      \textbf{Relation Definition}
       & Dense Graph                                                                 & 91.07 \\
      \cmidrule(lr){1-3}

      \multirow{3}{*}{\textbf{VAE Network}}
       & In-Degree Only                                                              & 97.38 \\
       & Out-Degree Only                                                             & 97.95 \\
       & In-Out Mixing                                                               & 98.19 \\
      \midrule

      \multicolumn{2}{c|}{\textbf{Ours}}
       & \textbf{99.39}                                                                      \\
      \bottomrule
    \end{tabular}}
\end{table}

\paragraph{Cascade sequence}
Given that the object list must be generated prior to object properties, and relations must precede OBB layout, there exists only one viable alternative to \name's generation sequence (Object $\to$ Property $\to$ Relation $\to$ OBBs): the Object $\to$ Relation $\to$ Property $\to$ OBBs order. As shown in \zcref{tab:ablation}, since relation definitions are correlated with object sizes, defining object properties before relations yields superior generation quality.

\paragraph{Relation modeling: explicit vs. implicit}
We further examine relation representations by comparing our sparse relation graph with a fully connected version (\emph{Dense-Relation}), where every node pair is linked. As shown in \zcref{tab:ablation}, Training with dense graphs introduce complex dependencies that degrade layout quality. To assess explicit relation modeling, we replace our latent representation with an explicit relation graph following InstructScene~\cite{lin2024instructscene} (\emph{Exp-Relation}). Results show that our implicit approach achieves stronger performance by compressing redundant information and reducing quadratic complexity to a linear latent representation.

\paragraph{Co-training scheme} We ablate the co-training strategy between the relation VAE and box layout diffusion by training the VAE independently (\emph{No Co-Training}). As shown in \zcref{tab:ablation}, this reduces synthesis quality, underscoring the importance of joint training for capturing relational structure.

\paragraph{Relation definition: dense vs. sparse}
To further validate graph sparsification, we compare the reconstruction accuracy of the VAE (with identical architecture) on both fully connected and sparse graphs, as shown in \zcref{tab:ablation_vae}. Reconstruction accuracy for dense graphs is substantially lower than that for sparse graphs. Since relation reconstruction fidelity affects the quality of relation representations used for layout generation, these results suggest that dense graphs are harder to model and can degrade relation generation quality.

\paragraph{In-out attention mechanism}
We also evaluate the proposed in-out attention mechanism in the relation VAE. Instead of two-step attention over in-degree and out-degree subgraphs, we combine all edges in a single cross-attention layer and increase block depth to match the parameter count. We also compare variants using only in-degree or only out-degree attention. Reconstruction errors on the test set (\zcref{tab:ablation_vae}) show that in-out attention improves relation reconstruction quality, validating its effectiveness.

\paragraph{Building elements: doors \& windows}
To validate the effectiveness of building element modeling, we compare the overlap between generated furniture layouts and the operational spaces (opening/closing zones) of architectural elements across different room types, both with and without these constraints. As shown in \zcref{tab:door_iou}, incorporating architectural elements prevents generated layouts from obstructing the operation of doors and windows, improving physical plausibility.
\begin{table}[t]
    \centering
    \caption{Comparison of IoU with door/window operational spaces (w/wo building elements). Explicitly modeling building elements effectively prevents objects from obstructing the operation of doors and windows.}
    \label{tab:door_iou}

    \footnotesize
    \setlength{\tabcolsep}{2pt}
    \renewcommand{\arraystretch}{1.2}

    \scalebox{1.1}{
        \begin{tabular}{l | ccc}
            \toprule
            \textbf{Method}
                                      & \textbf{Bedroom}
                                      & \textbf{Livingroom}
                                      & \textbf{Diningroom}                                 \\
            \midrule

            With Building Elements    & 18.91               & 25.43         & 17.20         \\
            Without Building Elements & \textbf{9.05}       & \textbf{6.97} & \textbf{8.75} \\
            \bottomrule
        \end{tabular}}
\end{table}


\section{Conclusion and Discussion} \label{sec:conclusion}

We introduced \name, a cascaded diffusion framework for controllable 3D furniture layout synthesis that achieves consistent performance across multiple tasks under both constrained and unconstrained settings. By decomposing layout generation into sequential stages and modeling spatial relations implicitly, our approach enables flexible scene control within a unified generative pipeline.

Despite these advances, several challenges remain. The scarcity of large-scale annotated 3D scene datasets limits diversity and generalization, particularly in out-of-distribution or extreme scenarios, as illustrated in \zcref{fig:failure_case}. While our decomposition strategy mitigates data scarcity by learning task-specific priors, improving robustness under distribution shifts remains an open problem. Promising directions include leveraging synthetic data generation, multi-modal learning, and image-to-3D pipelines to expand scene diversity. Furthermore, when severe errors are introduced in earlier generation stages, subsequent stages currently lack the ability to correct them (see the discussion in \arxiv{\zcref{subsec:ep}}{Suppl.~Sec.~3.5}).

Furniture retrieval in our current pipeline relies on a fixed CAD library, which constrains stylistic diversity. Although this approach provides stable surface quality and geometric consistency compared to existing 3D generative models, future work could explore hybrid retrieval--generation strategies or integrate neural shape and texture synthesis to improve stylistic flexibility.

In addition, our relational modeling currently focuses on pairwise interactions and does not capture higher-order or group relations commonly used in professional design workflows. Extending the framework to incorporate group-level relational structures may enable richer relational reasoning and improve layout coherence.

Finally, our framework is currently designed for indoor furniture layouts. Extending it to outdoor scene generation---such as city blocks and urban planning---is a potential direction for future research, although relation definitions and generation stages will likely require adaptation to domain-specific design logic.

\bibliographystyle{ACM-Reference-Format}
\bibliography{src/ref}

@String {CAGD     = {Comput. Aided Geom. Des.} }

@String {CGF      = {Comput. Graph. Forum} }

@string {CVPR     = {IEEE/CVF Conference on Computer Vision and Pattern Recognition} }

@String {ECCV = {European Conference on Computer Vision}}

@String {ICCV = {International Conference on Computer Vision}}

@String {ICLR     = {International Conference on Learning Representations} }

@String {ICML = {International Conference on Machine Learning}}

@String {NeurIPS = {International Conference on Neural Information Processing Systems}}

@String {NIPS = {International Conference on Neural Information Processing Systems}}

@String {SA       = {ACM Trans. Graph.} }

@String {SIG      = {ACM Trans. Graph.} }

@String {threeDV     = {International Conference on 3D Vision} }

@String {TOG      = {ACM Trans. Graph.} }

@String {TPAMI     = {IEEE Trans. Pattern Anal. Mach. Intell.} }

@String {TVCG     = {IEEE Trans. Vis. Comput. Graphics} }

@article{gao2023scenehgn,
  title     = {\href{https://arxiv.org/abs/2302.10237}{SceneHGN: Hierarchical graph networks for 3D indoor scene generation with fine-grained geometry}},
  author    = {Gao, Lin and Sun, Jia-Mu and Mo, Kaichun and Lai, Yu-Kun and Guibas, Leonidas J and Yang, Jie},
  journal   = TPAMI,
  volume    = {45},
  number    = {7},
  pages     = {8902--8919},
  year      = {2023},
  publisher = {IEEE}
}

@inproceedings{xu2024set,
  title     = {\href{https://arxiv.org/abs/2405.11928}{``Set It Up!'': Functional object arrangement with compositional generative models}},
  author    = {Xu, Yiqing and Mao, Jiayuan and Du, Yilun and Loz{\'a}no-P{\'e}rez, Tomas and Kaebling, Leslie Pack and Hsu, David},
  booktitle = {Robotics: Science and Systems},
  year      = {2024}
}

@article{wang2019planit,
  author  = {Wang, Kai and Lin, Yu-An and Weissmann, Ben and Savva, Manolis and Chang, Angel X. and Ritchie, Daniel},
  title   = {\href{https://dl.acm.org/doi/pdf/10.1145/3306346.3322941}{PlanIT: Planning and instantiating indoor scenes with relation graph and spatial prior networks}},
  year    = {2019},
  volume  = {38},
  number  = {4},
  journal = TOG,
  pages   = {132:1-132:15}
}

@inproceedings{paschalidou2021atiss,
  author    = {Despoina Paschalidou and Amlan Kar and Maria Shugrina and Karsten Kreis and Andreas Geiger and Sanja Fidler},
  title     = {\href{https://research.nvidia.com/labs/toronto-ai/ATISS/}{ATISS: Autoregressive transformers for indoor scene synthesis}},
  booktitle = NeurIPS,
  pages     = {12013--12026},
  year      = {2021}
}

@inproceedings{para2023cofs,
  title     = {\href{https://arxiv.org/abs/2205.14657}{COFS: Controllable furniture layout synthesis}},
  author    = {Para, Wamiq Reyaz and Guerrero, Paul and Mitra, Niloy and Wonka, Peter},
  booktitle = {SIGGRAPH},
  pages     = {36:1--36:11},
  year      = {2023}
}

@inproceedings{sun2024forest2seq,
  title     = {\href{https://arxiv.org/abs/2407.05388}{FOREST2SEQ: Revitalizing order prior for sequential indoor scene synthesis}},
  author    = {Sun, Qi and Zhou, Hang and Zhou, Wengang and Li, Li and Li, Houqiang},
  booktitle = ECCV,
  pages     = {251--268},
  year      = {2025}
}

@inproceedings{wang2021sceneformer,
  title     = {\href{https://arxiv.org/abs/2012.09793}{SceneFormer: Indoor scene generation with transformers}},
  author    = {Wang, Xinpeng and Yeshwanth, Chandan and Nie{\ss}ner, Matthias},
  booktitle = threeDV,
  pages     = {106--115},
  year      = {2021}
}

@inproceedings{ritchie2019fast,
  title     = {\href{https://openaccess.thecvf.com/content_CVPR_2019/papers/Ritchie_Fast_and_Flexible_Indoor_Scene_Synthesis_via_Deep_Convolutional_Generative_CVPR_2019_paper.pdf}{Fast and flexible indoor scene synthesis via deep convolutional generative models}},
  author    = {Ritchie, Daniel and Wang, Kai and Lin, Yu-an},
  booktitle = CVPR,
  pages     = {6182--6190},
  year      = {2019}
}

@inproceedings{zhao2024roomdesigner,
  title     = {\href{https://arxiv.org/abs/2310.10027}{RoomDsigner: Encoding anchor-latents for style-consistent and shape-compatible indoor scene generation}},
  author    = {Zhao, Yiqun and Zhao, Zibo and Li, Jing and Dong, Sixun and Gao, Shenghua},
  booktitle = threeDV,
  pages     = {1413--1423},
  year      = {2024}
}

@inproceedings{zhai2024commonscenes,
  title   = {\href{https://sites.google.com/view/commonscenes}{CommonScenes: Generating commonsense 3D indoor scenes with scene graphs}},
  author  = {Zhai, Guangyao and {\"O}rnek, Evin P{\i}nar and Wu, Shun-Cheng and Di, Yan and Tombari, Federico and Navab, Nassir and Busam, Benjamin},
  journal = NIPS,
  pages   = {30026--30038},
  year    = {2024}
}

@inproceedings{maillard2024debara,
  title     = {\href{https://arxiv.org/abs/2409.18336}{DeBaRA: Denoising-based 3D room arrangement generation}},
  author    = {Maillard, L{\'e}opold and Sereyjol-Garros, Nicolas and Durand, Tom and Ovsjanikov, Maks},
  booktitle = NIPS,
  year      = {2024}
}

@inproceedings{tang2024diffuscene,
  title     = {\href{https://tangjiapeng.github.io/projects/DiffuScene/}{DiffuScene: Denoising diffusion models for generative indoor scene synthesis}},
  author    = {Tang, Jiapeng and Nie, Yinyu and Markhasin, Lev and Dai, Angela and Thies, Justus and Nie{\ss}ner, Matthias},
  booktitle = CVPR,
  pages     = {20507--20518},
  year      = {2024}
}

@inproceedings{zhai2024echoscene,
  title     = {\href{https://arxiv.org/abs/2405.00915}{EchoScene: Indoor scene generation via information echo over scene graph diffusion}},
  author    = {Zhai, Guangyao and {\"O}rnek, Evin P{\i}nar and Chen, Dave Zhenyu and Liao, Ruotong and Di, Yan and Navab, Nassir and Tombari, Federico and Busam, Benjamin},
  booktitle = ECCV,
  year      = {2024}
}

@misc{naanaa20233d,
  title        = {\href{https://arxiv.org/abs/2308.04468}{3D scene diffusion guidance using scene graphs}},
  author       = {Naanaa, Mohammad and Schmid, Katharina and Nie, Yinyu},
  howpublished = {arXiv:2308.04468},
  year         = {2023}
}

@misc{hu2024mixed,
  title        = {\href{https://arxiv.org/abs/2405.21066}{Mixed diffusion for 3D indoor scene synthesis}},
  author       = {Hu, Siyi and Arroyo, Diego Martin and Debats, Stephanie and Manhardt, Fabian and Carlone, Luca and Tombari, Federico},
  howpublished = {arXiv:2405.21066},
  year         = {2024}
}

@inproceedings{ccelen2024design,
  title     = {\href{https://arxiv.org/abs/2404.02838}{I-design: Personalized LLM interior designer}},
  author    = {{\c{C}}elen, Ata and Han, Guo and Schindler, Konrad and Van Gool, Luc and Armeni, Iro and Obukhov, Anton and Wang, Xi},
  booktitle = {ECCV 2024 Workshops},
  year      = {2024}
}

@inproceedings{lin2024instructscene,
  title     = {\href{https://chenguolin.github.io/projects/InstructScene/}{InstructScene: Instruction-driven 3D indoor scene synthesis with semantic graph prior}},
  author    = {Lin, Chenguo and Mu, Yadong},
  booktitle = ICLR,
  year      = {2024}
}

@misc{yang2024llplace,
  title        = {\href{https://arxiv.org/abs/2406.03866}{LLplace: The 3D indoor scene layout generation and editing via large language model}},
  author       = {Yang, Yixuan and Lu, Junru and Zhao, Zixiang and Luo, Zhen and Yu, James JQ and Sanchez, Victor and Zheng, Feng},
  howpublished = {arXiv:2406.03866},
  year         = {2024}
}

@misc{wang2024chat2layout,
  title        = {\href{https://arxiv.org/abs/2407.21333}{Chat2Layout: Interactive 3D furniture layout with a multimodal LLM}},
  author       = {Wang, Can and Zhong, Hongliang and Chai, Menglei and He, Mingming and Chen, Dongdong and Liao, Jing},
  howpublished = {arXiv:2407.21333},
  year         = {2024}
}

@inproceedings{yang2024holodeck,
  title     = {\href{yueyang1996.github.io/holodeck/}{Holodeck: Language guided generation of 3D embodied AI environments}},
  author    = {Yang, Yue and Sun, Fan-Yun and Weihs, Luca and VanderBilt, Eli and Herrasti, Alvaro and Han, Winson and Wu, Jiajun and Haber, Nick and Krishna, Ranjay and Liu, Lingjie and others},
  booktitle = CVPR,
  pages     = {16227--16237},
  year      = {2024}
}

@inproceedings{fu2025anyhome,
  title     = {\href{https://arxiv.org/abs/2312.06644}{AnyHome: Open-vocabulary generation of structured and textured 3D homes}},
  author    = {Fu, Rao and Wen, Zehao and Liu, Zichen and Sridhar, Srinath},
  booktitle = ECCV,
  pages     = {52--70},
  year      = {2025}
}

@inproceedings{ocal2024sceneteller,
  title     = {\href{https://sceneteller.github.io/}{SceneTeller: Language-to-3D Scene generation}},
  author    = {{\"O}cal, Ba{\c{s}}ak Melis and Tatarchenko, Maxim and Karaoglu, Sezer and Gevers, Theo},
  booktitle = ECCV,
  pages     = {362--378},
  year      = {2025}
}

@inproceedings{feng2023layoutgpt,
  title     = {\href{https://layoutgpt.github.io/}{LayoutGPT: Compositional Visual planning and generation with large language models}},
  author    = {Feng, Weixi and Zhu, Wanrong and Fu, Tsu-jui and Jampani, Varun and Akula, Arjun and He, Xuehai and Basu, Sugato and Wang, Xin Eric and Wang, William Yang},
  booktitle = NeurIPS,
  pages     = {18225--18250},
  year      = {2023}
}

@article{wangcgfsurvey2023,
  author  = {Wang, Y.T. and Liang, C. and Huai, N. and Chen, J. and Zhang, C.J.},
  title   = {\href{https://diglib.eg.org/server/api/core/bitstreams/8cbeb9f1-fcdf-4b2f-954f-c60fcd367c1a/content}{A survey of personalized interior design}},
  journal = CGF,
  volume  = {42},
  number  = {6},
  year    = {2023}
}

@inproceedings{patil2024advances,
  title     = {\href{https://arxiv.org/abs/2304.03188}{Advances in data-driven analysis and synthesis of 3D indoor scenes}},
  author    = {Patil, Akshay Gadi and Patil, Supriya Gadi and Li, Manyi and Fisher, Matthew and Savva, Manolis and Zhang, Hao},
  booktitle = CGF,
  volume    = {43},
  number    = {1},
  pages     = {e14927},
  year      = {2024}
}

@article{fisher2012example,
  title     = {\href{https://graphics.stanford.edu/projects/scenesynth/}{Example-based synthesis of 3D object arrangements}},
  author    = {Fisher, Matthew and Ritchie, Daniel and Savva, Manolis and Funkhouser, Thomas and Hanrahan, Pat},
  journal   = SA,
  volume    = {31},
  number    = {6},
  pages     = {135:1--135:11},
  year      = {2012},
  publisher = {ACM New York, NY, USA}
}

@article{merrell2011interactive,
  title     = {\href{http://graphics.berkeley.edu/papers/Merrell-IFL-2011-08/Merrell-IFL-2011-08.pdf}{Interactive furniture layout using interior design guidelines}},
  author    = {Merrell, Paul and Schkufza, Eric and Li, Zeyang and Agrawala, Maneesh and Koltun, Vladlen},
  journal   = SIG,
  volume    = {30},
  number    = {4},
  pages     = {87:1--87:10},
  year      = {2011},
  publisher = {ACM New York, NY, USA}
}

@inproceedings{wei2023lego,
  title     = {\href{https://openaccess.thecvf.com/content/CVPR2023/papers/Wei_LEGO-Net_Learning_Regular_Rearrangements_of_Objects_in_Rooms_CVPR_2023_paper.pdf}{LEGO-Net: Learning regular rearrangements of objects in rooms}},
  author    = {Wei, Qiuhong Anna and Ding, Sijie and Park, Jeong Joon and Sajnani, Rahul and Poulenard, Adrien and Sridhar, Srinath and Guibas, Leonidas},
  booktitle = CVPR,
  pages     = {19037--19047},
  year      = {2023}
}

@article{GIULIARI2024,
  title   = {\href{https://www.sciencedirect.com/science/article/pii/S0167865524002988}{Positional diffusion: Graph-based diffusion models for set ordering}},
  journal = {Pattern Recognition Letters},
  year    = {2024},
  volume  = {186},
  number  = {C},
  pages   = {272--278},
  author  = {Francesco Giuliari and Gianluca Scarpellini and Stefano Fiorini and Stuart James and Pietro Morerio and Yiming Wang and Alessio Del Bue}
}

@inproceedings{yang2024physcene,
  title     = {\href{https://physcene.github.io/}{PhyScene: Physically interactable 3D scene synthesis for embodied AI}},
  author    = {Yang, Yandan and Jia, Baoxiong and Zhi, Peiyuan and Huang, Siyuan},
  booktitle = CVPR,
  pages     = {16262--16272},
  year      = {2024}
}

@inproceedings{li2024gltscene,
  title     = {\href{https://diglib.eg.org/server/api/core/bitstreams/5cae258d-6bbd-435d-b1e1-97e7206d45cf/content}{GLTScene: Global-to-local transformers for indoor scene synthesis with general room boundaries}},
  author    = {Li, Yijie and Xu, Pengfei and Ren, Junquan and Shao, Zefan and Huang, Hui},
  booktitle = CGF,
  volume    = {43},
  number    = {7},
  pages     = {e15236},
  year      = {2024}
}

@inproceedings{kikuchi2021constrained,
  title     = {\href{https://arxiv.org/abs/2108.00871}{Constrained graphic layout generation via latent optimization}},
  author    = {Kikuchi, Kotaro and Simo-Serra, Edgar and Otani, Mayu and Yamaguchi, Kota},
  booktitle = {ACM Multimedia},
  pages     = {88--96},
  year      = {2021}
}

@inproceedings{dhamo2021graph,
  title     = {\href{https://openaccess.thecvf.com/content/ICCV2021/papers/Dhamo_Graph-to-3D_End-to-End_Generation_and_Manipulation_of_3D_Scenes_Using_Scene_ICCV_2021_paper.pdf}{Graph-to-3D: End-to-end generation and manipulation of 3D scenes using scene graphs}},
  author    = {Dhamo, Helisa and Manhardt, Fabian and Navab, Nassir and Tombari, Federico},
  booktitle = ICCV,
  pages     = {16352--16361},
  year      = {2021}
}

@online{rolph7principles,
  title        = {\href{https://www.dessource.com/post/a-guide-to-7-principles-of-interior-design}{A guide to 7 principles of interior design}},
  author       = {Linday Rolph},
  howpublished = {https://www.dessource.com/post/a-guide-to-7-principles-of-interior-design},
  lastaccessed = {Nov. 22, 2024},
  year         = {2024}
}

@online{tru13principles,
  title        = {\href{https://www.redesigndaily.com/home-design/design-trends/material-trends/principles-of-interior-design-44589574}{The 13 key principles of interior design explained}},
  author       = {Andrea Tru},
  howpublished = {https://www.redesigndaily.com/home-design/design-trends/material-trends/principles-of-interior-design-44589574},
  lastaccessed = {Nov. 22, 2024},
  year         = {2024}
}

@book{grimley2022universal,
  title     = {\href{https://quarto.com/books/9780760372128/universal-principles-of-interior-design}{Universal Principles of interior design: 100 ways to develop innovative ideas, enhance usability, and design effective solutions}},
  author    = {Grimley, Chris and Smith, Kelly Harris},
  year      = {2022},
  publisher = {Rockport Publishers}
}

@article{li2019grains,
  title     = {\href{https://arxiv.org/abs/1807.09193}{GRAINS: Generative recursive autoencoders for indoor scenes}},
  author    = {Li, Manyi and Patil, Akshay Gadi and Xu, Kai and Chaudhuri, Siddhartha and Khan, Owais and Shamir, Ariel and Tu, Changhe and Chen, Baoquan and Cohen-Or, Daniel and Zhang, Hao},
  journal   = TOG,
  volume    = {38},
  number    = {2},
  pages     = {12:1--12:16},
  year      = {2019},
  publisher = {ACM New York, NY, USA}
}

@article{yao2024conditional,
  title     = {\href{https://www.sciencedirect.com/science/article/abs/pii/S0097849324001067}{Conditional room layout generation based on graph neural networks}},
  author    = {Yao, Zhihan and Chen, Yuhang and Cui, Jiahao and Zhang, Shoulong and Li, Shuai and Hao, Aimin},
  journal   = {Computers \& Graphics},
  pages     = {103971},
  year      = {2024},
  publisher = {Elsevier}
}

@article{fisher2011characterizing,
  title   = {\href{https://graphics.stanford.edu/~mdfisher/papers/graphKernel.pdf}{Characterizing structural relationships in scenes using graph kernels}},
  author  = {Fisher, Matthew and Savva, Manolis and Hanrahan, Pat},
  journal = TOG,
  volume  = {30},
  number  = {4},
  pages   = {34:1--34:12},
  year    = {2011}
}

@inproceedings{ho2020denoising,
  title     = {\href{https://arxiv.org/abs/2006.11239}{Denoising diffusion probabilistic models}},
  author    = {Ho, Jonathan and Jain, Ajay and Abbeel, Pieter},
  booktitle = NIPS,
  pages     = {6840--6851},
  year      = {2020}
}

@inproceedings{fu20213dfront,
  title     = {\href{https://arxiv.org/abs/2011.09127}{3D-FRONT: 3D furnished rooms with layouts and semantics}},
  author    = {Fu, Huan and Cai, Bowen and Gao, Lin and Zhang, Ling-Xiao and Wang, Jiaming and Li, Cao and Zeng, Qixun and Sun, Chengyue and Jia, Rongfei and Zhao, Binqiang and Zhang, Hao},
  booktitle = ICCV,
  pages     = {10933--10942},
  year      = {2021}
}

@inproceedings{heusel2017gans,
  title     = {\href{https://arxiv.org/abs/1706.08500}{GANs trained by a two time-scale update rule converge to a local NASH equilibrium}},
  author    = {Heusel, Martin and Ramsauer, Hubert and Unterthiner, Thomas and Nessler, Bernhard and Hochreiter, Sepp},
  booktitle = NIPS,
  pages     = {6629--6640},
  year      = {2017}
}

@inproceedings{yi2022mime,
  title     = {\href{https://arxiv.org/abs/2212.04360}{{MIME}: Human-aware {3D} scene generation}},
  author    = {Yi, Hongwei and Huang, Chun-Hao P. and Tripathi, Shashank and Hering, Lea and 
               Thies, Justus and Black, Michael J.},
  booktitle = CVPR,
  pages     = {12965-12976},
  year      = {2023}
}

@misc{hong2024human,
  title        = {\href{https://arxiv.org/abs/2212.04360}{Human-aware 3D scene generation with spatially-constrained diffusion models}},
  author       = {Hong, Xiaolin and Yi, Hongwei and He, Fazhi and Cao, Qiong},
  howpublished = {arXiv:2406.18159},
  year         = {2024}
}

@misc{achiam2023gpt,
  title        = {\href{https://arxiv.org/pdf/2303.08774}{GPT-4 technical report}},
  author       = {Achiam, Josh and Adler, Steven and Agarwal, Sandhini and Ahmad, Lama and Akkaya, Ilge and Aleman, Florencia Leoni and Almeida, Diogo and Altenschmidt, Janko and Altman, Sam and Anadkat, Shyamal and others},
  howpublished = {arXiv:2303.08774},
  year         = {2023}
}

@inproceedings{zhao2023unipc,
  title     = {\href{https://arxiv.org/abs/2302.04867}{UniPC: A unified predictor-corrector framework for fast sampling of diffusion models}},
  author    = {Zhao, Wenliang and Bai, Lujia and Rao, Yongming and Zhou, Jie and Lu, Jiwen},
  booktitle = NIPS,
  pages     = {49842--49869},
  year      = {2023}
}

@article{lu2025dpm,
  title     = {\href{https://arxiv.org/abs/2211.01095}{Dpm-solver++: Fast solver for guided sampling of diffusion probabilistic models}},
  author    = {Lu, Cheng and Zhou, Yuhao and Bao, Fan and Chen, Jianfei and Li, Chongxuan and Zhu, Jun},
  journal   = {Machine Intelligence Research},
  pages     = {1--22},
  year      = {2025},
  publisher = {Springer}
}

@misc{sun2025semlayoutdiff,
  title        = {\href{https://arxiv.org/abs/2508.18597}{SemLayoutDiff: Semantic layout generation with diffusion model for indoor scene synthesis}},
  author       = {Sun, Xiaohao and Goel, Divyam and Chang, Angel X},
  howpublished = {arXiv:2508.18597},
  year         = {2025}
}

@inproceedings{feng2025casagpt,
  title     = {\href{https://arxiv.org/abs/2504.19478}{CasaGPT: cuboid arrangement and scene assembly for interior design}},
  author    = {Feng, Weitao and Zhou, Hang and Liao, Jing and Cheng, Li and Zhou, Wenbo},
  booktitle = CVPR,
  pages     = {29173--29182},
  year      = {2025}
}

@inproceedings{deng2025global,
  title     = {\href{https://arxiv.org/abs/2503.18476}{Global-local tree search in VLMs for 3D indoor scene generation}},
  author    = {Deng, Wei and Qi, Mengshi and Ma, Huadong},
  booktitle = CVPR,
  pages     = {8975--8984},
  year      = {2025}
}

@inproceedings{sun2025layoutvlm,
  title     = {\href{https://arxiv.org/abs/2412.02193}{LayoutVLM: Differentiable optimization of 3D layout via vision-language models}},
  author    = {Sun, Fan-Yun and Liu, Weiyu and Gu, Siyi and Lim, Dylan and Bhat, Goutam and Tombari, Federico and Li, Manling and Haber, Nick and Wu, Jiajun},
  booktitle = CVPR,
  pages     = {29469--29478},
  year      = {2025}
}

@inproceedings{gumin2025procedural,
  title     = {\href{https://arxiv.org/abs/2510.16147}{Procedural Scene programs for open-universe scene generation: LLM-free error correction via program search}},
  author    = {Gumin, Maxim and Han, Do Heon and Yoo, Seung Jean and Ganeshan, Aditya and Jones, R Kenny and Fu, Kailiang and Aguina-Kang, Rio and Morris, Stewart and Ritchie, Daniel},
  booktitle = {SIGGRAPH Asia},
  pages     = {141:1--141:11},
  year      = {2025}
}

@inproceedings{binkowski2018demystifying,
  title     = {\href{https://arxiv.org/pdf/1801.01401}{Demystifying MMD GANs}},
  author    = {Bi{\'n}kowski, Miko{\l}aj and Sutherland, Danica J and Arbel, Michael and Gretton, Arthur},
  booktitle = ICML,
  year      = {2018}
}

@misc{grattafiori2024llama,
  title        = {\href{https://arxiv.org/abs/2407.21783}{The LLaMA 3 herd of models}},
  author       = {Grattafiori, Aaron and Dubey, Abhimanyu and Jauhri, Abhinav and Pandey, Abhinav and Kadian, Abhishek and Al-Dahle, Ahmad and Letman, Aiesha and Mathur, Akhil and Schelten, Alan and Vaughan, Alex and others},
  howpublished = {arXiv:2407.21783},
  year         = {2024}
}

@misc{team2024gemini,
  title        = {\href{https://arxiv.org/abs/2403.05530}{Gemini 1.5: Unlocking multimodal understanding across millions of tokens of context}},
  author       = {Team, Gemini and Georgiev, Petko and Lei, Ving Ian and Burnell, Ryan and Bai, Libin and Gulati, Anmol and Tanzer, Garrett and Vincent, Damien and Pan, Zhufeng and Wang, Shibo and others},
  howpublished = {arXiv:2403.05530},
  year         = {2024}
}

@misc{aguina2024open,
  title        = {\href{https://arxiv.org/abs/2403.09675}{Open-universe indoor scene generation using LLM program synthesis and uncurated object databases}},
  author       = {Aguina-Kang, Rio and Gumin, Maxim and Han, Do Heon and Morris, Stewart and Yoo, Seung Jean and Ganeshan, Aditya and Jones, R Kenny and Wei, Qiuhong Anna and Fu, Kailiang and Ritchie, Daniel},
  howpublished = {arXiv:2403.09675},
  year         = {2024}
}

@article{yeh2012synthesizing,
  title     = {\href{https://graphics.stanford.edu/~lfyg/owl.pdf}{Synthesizing open worlds with constraints using locally annealed reversible jump MCMC}},
  author    = {Yeh, Yi-Ting and Yang, Lingfeng and Watson, Matthew and Goodman, Noah D and Hanrahan, Pat},
  journal   = TOG,
  volume    = {31},
  number    = {4},
  pages     = {56:1--56:11},
  year      = {2012},
  publisher = {ACM New York, NY, USA}
}

@article{weiss2018fast,
  title     = {\href{https://arxiv.org/abs/1809.10526}{Fast and scalable position-based layout synthesis}},
  author    = {Weiss, Tomer and Litteneker, Alan and Duncan, Noah and Nakada, Masaki and Jiang, Chenfanfu and Yu, Lap-Fai and Terzopoulos, Demetri},
  journal   = TVCG,
  volume    = {25},
  number    = {12},
  pages     = {3231--3243},
  year      = {2018},
  publisher = {IEEE}
}

@article{yu2011make,
  title   = {\href{https://dl.acm.org/doi/10.1145/2010324.1964981}{Make it home: automatic optimization of furniture arrangement}},
  author  = {Yu, Lap-Fai and Yeung, Sai Kit and Tang, Chi-Keung and Terzopoulos, Demetri and Chan, Tony F and Osher, Stanley J},
  journal = TOG,
  volume  = {30},
  number  = {4},
  pages   = {86:1--86:12},
  year    = {2011}
}

@article{bustin2021lossy,
  title     = {On lossy compression of directed graphs},
  author    = {Bustin, Ronit and Shayevitz, Ofer},
  journal   = {IEEE Transactions on Information Theory},
  volume    = {68},
  number    = {4},
  pages     = {2101--2122},
  year      = {2021},
  publisher = {IEEE}
}

@inproceedings{van2017neural,
  title     = {\href{https://proceedings.neurips.cc/paper/2017/file/7a98af17e63a0ac09ce2e96d03992fbc-Paper.pdf}{Neural discrete representation learning}},
  author    = {Van Den Oord, Aaron and Vinyals, Oriol and others},
  booktitle = NIPS,
  volume    = {30},
  year      = {2017}
}

@inproceedings{liu2023openshape,
  title     = {\href{https://proceedings.neurips.cc/paper_files/paper/2023/file/8c7304e77c832ddc70075dfee081ca6c-Paper-Conference.pdf}{Openshape: Scaling up 3D shape representation towards open-world understanding}},
  author    = {Liu, Minghua and Shi, Ruoxi and Kuang, Kaiming and Zhu, Yinhao and Li, Xuanlin and Han, Shizhong and Cai, Hong and Porikli, Fatih and Su, Hao},
  booktitle = NIPS,
  volume    = {36},
  pages     = {44860--44879},
  year      = {2023}
}

@inproceedings{jang2016categorical,
  title     = {\href{https://arxiv.org/pdf/1611.01144}{Categorical reparameterization with Gumbel-Softmax}},
  author    = {Jang, Eric and Gu, Shixiang and Poole, Ben},
  booktitle = ICLR,
  year      = {2016}
}

@inproceedings{ramesh2021zero,
  title        = {\href{http://proceedings.mlr.press/v139/ramesh21a/ramesh21a.pdf}{Zero-shot text-to-image generation}},
  author       = {Ramesh, Aditya and Pavlov, Mikhail and Goh, Gabriel and Gray, Scott and Voss, Chelsea and Radford, Alec and Chen, Mark and Sutskever, Ilya},
  booktitle    = ICML,
  pages        = {8821--8831},
  year         = {2021},
  organization = {Pmlr}
}

@inproceedings{dai2017scannet,
  title     = {\href{https://openaccess.thecvf.com/content_cvpr_2017/html/Dai_ScanNet_Richly-Annotated_3D_CVPR_2017_paper.html}{Scannet: Richly-annotated 3D reconstructions of indoor scenes}},
  author    = {Dai, Angela and Chang, Angel X and Savva, Manolis and Halber, Maciej and Funkhouser, Thomas and Nie{\ss}ner, Matthias},
  booktitle = CVPR,
  pages     = {5828--5839},
  year      = {2017}
}

@article{littlefair2025flairgpt,
  title   = {\href{https://onlinelibrary.wiley.com/doi/abs/10.1111/cgf.70036}{FlairGPT: Repurposing LLMs for interior designs}},
  author  = {Littlefair, Gabrielle and Dutt, Niladri Shekhar and Mitra, Niloy J},
  journal = CGF,
  volume  = {44},
  number  = {2},
  pages   = {e70036},
  year    = {2025}
}

@article{wu2025sceneflow,
  title     = {\href{https://wutomwu.github.io/publications/2025-SceneFlow/paper.pdf}{SceneFlow: Synthesizing indoor scenes via geometry-enhanced flow matching}},
  author    = {Wu, Wenming and Shen, Akang and Yin, Yanzhe and Chen, Zixiang and Zhang, Gaofeng and Zheng, Liping},
  journal   = CAGD,
  volume    = {123},
  pages     = {102489},
  year      = {2025},
  publisher = {Elsevier}
}

@inproceedings{leimer2022layoutenhancer,
  title     = {\href{https://dl.acm.org/doi/pdf/10.1145/3550469.3555425}{LayoutEnhancer: Generating good indoor layouts from imperfect data}},
  author    = {Leimer, Kurt and Guerrero, Paul and Weiss, Tomer and Musialski, Przemyslaw},
  booktitle = {SIGGRAPH Asia Conference Papers},
  pages     = {27:1--27:8},
  year      = {2022}
}

@inproceedings{fang2025text,
  title        = {\href{https://onlinelibrary.wiley.com/doi/abs/10.1111/cgf.70039}{Text-guided interactive scene synthesis with scene prior guidance}},
  author       = {Fang, Shaoheng and Yang, Haitao and Mooney, Raymond and Huang, Qixing},
  booktitle    = CGF,
  volume       = {44},
  number       = {2},
  pages        = {e70039},
  year         = {2025},
  organization = {Wiley Online Library}
}

@article{wang2018deep,
  title   = {\href{https://dl.acm.org/doi/pdf/10.1145/3197517.3201362}{Deep convolutional priors for indoor scene synthesis}},
  author  = {Wang, Kai and Savva, Manolis and Chang, Angel X and Ritchie, Daniel},
  journal = TOG,
  volume  = {37},
  number  = {4},
  pages   = {70:1--70:14},
  year    = {2018}
}
\appendix

\arxiv{}{
This supplemental material provides:
(1) additional implementation details of \name, including the sparse relationship definition (\zcref{subsec:rd}), feature codes used for retrieval (\zcref{subsec:fi}), the definitions of node vectors at different stages (\zcref{subsec:vec}), the data augmentation methods employed during training (\zcref{subsec:da}), and further implementation details for different applications (\zcref{subsec:dapp});
(2) the prompts for LLM-guided scene generation (\zcref{subsec:p_llm}), text-driven scene generation (\zcref{subsec:p_text}), and image-driven scene generation (\zcref{subsec:p_image}); and
(3) additional experiments and discussions on physical constraint satisfaction (\zcref{subsec:pc}), comparisons with LayoutVLM~\cite{sun2025layoutvlm} (\zcref{subsec:layoutvlm}), detailed inference time analysis (\zcref{subsec:tapp}),
diversity (\zcref{subsec:div}), error propagation discussion (\zcref{subsec:ep}), and further generation results under the control of building elements (\zcref{subsec:mr}).
}

\section{Implementation Details}
\label{sec:relation_modeling}

\subsection{Sparse Relationship Definition}
\label{subsec:rd}
We extract low-entropy object combinations and group them into five semantic zones: lounging, dining, bedding, lighting (center-alignment only), and others (no relations). For the three primary zones (lounging/dining/bedding), we select the top-$k$ frequent large objects as anchors. Based on this partitioning, we construct the relational structure by (1) preserving intra-zone relations, (2) modeling inter-zone relations only through connections between zone anchors, and (3) clustering around anchors to distinguish multiple zones of the same type within a scene (\eg two dining areas).

These anchors appear in 82.80\%, 85.09\%, and 97.81\% of the corresponding zones. In addition, intra-zone co-occurrence rates (48.33\%, 26.93\%, 18.92\%) consistently exceed inter-zone co-occurrence rates (13.51\%, 14.43\%, 8.25\%).

This entropy-based structural extraction generalizes to novel scenes and can be fully automated, with manual labeling used only to assign semantic interpretations to the discovered clusters.

\subsection{Feature Indices}
\label{subsec:fi}

In our pipeline, \name retrieves objects from the database by matching generated shape features to their nearest neighbors. A standard approach is to encode shapes into a continuous latent space using a Variational Autoencoder (VAE)~\cite{tang2024diffuscene}. However, point cloud VAEs typically produce high-dimensional latent representations (\eg 64 or 128 dimensions). Directly generating such high-dimensional continuous vectors with diffusion models is challenging, as the optimization becomes unstable and the resulting latent space often lacks sufficient discriminability between distinct object styles. Consequently, continuous latent representations tend to introduce large retrieval errors, leading to stylistic inconsistency among furniture instances (\eg chairs and tables) within the same scene.

To address these limitations, following InstructScene~\cite{lin2024instructscene}, we adopt a Vector-Quantized Variational Autoencoder (VQ-VAE)~\cite{van2017neural} to obtain discrete feature indices for retrieval. This formulation compresses complex shape attributes into a low-dimensional discrete sequence (\eg length 4), which is significantly easier to generate and provides improved discriminability across objects.

\begin{figure}[t]
    \centering
    \includegraphics[width=0.95\linewidth]{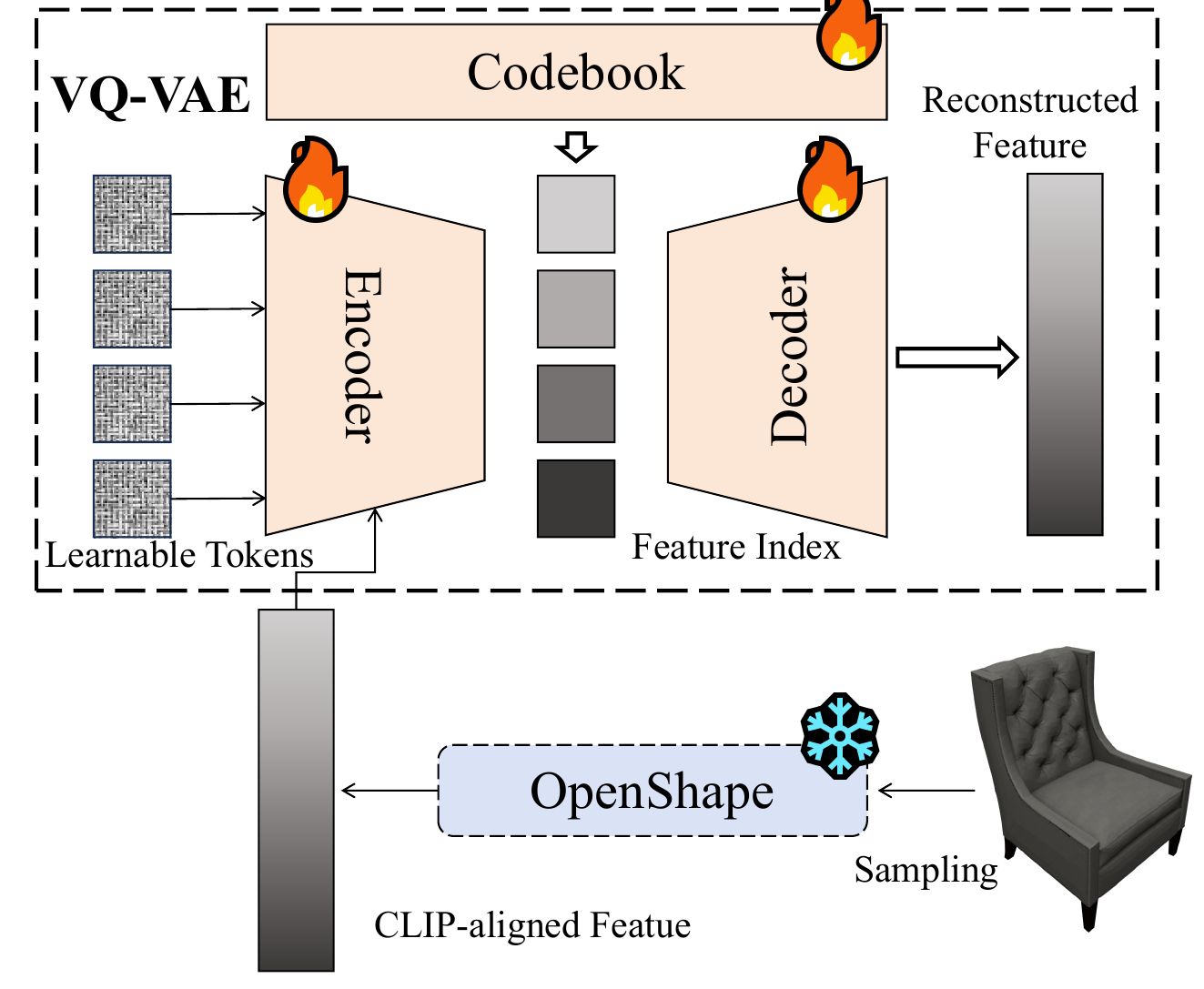}

    \caption{Schematic illustration of obtaining feature indices via VQ-VAE.}
    \label{fig:sp_feat}
\end{figure}

As illustrated in \zcref{fig:sp_feat}, we first employ OpenShape~\cite{liu2023openshape} to encode sampled colored point clouds into feature embeddings, which are subsequently quantized into discrete codes. The VQ-VAE consists of an encoder $E$, a decoder $D$, and a learnable codebook
$\mathcal{Z} \in \mathbb{R}^{K_f \times d_z}$. The model is trained by maximizing the Evidence Lower Bound (ELBO) of the feature log-likelihood:
\begin{equation}
    \mathbb{E}_{\mathbf{z} \sim p_E(\mathbf{z}|\mathbf{f})}
    \left[
        \log p_D(\mathbf{f}|\mathbf{z})
        - \beta D_{\text{KL}}\big(p_E(\mathbf{z}|\mathbf{f}) \,\|\, p(\mathbf{z})\big)
        \right],
\end{equation}
where $\mathbf{z}$ denotes latent vectors indexed by a sequence of discrete scalars $f_m \in \{1, \dots, K_f\}$. We adopt the Gumbel-Softmax relaxation~\cite{jang2016categorical,ramesh2021zero} to enable differentiable optimization of the quantization process.

During generation, we further observe that jointly modeling discrete feature indices and continuous object sizes introduces interference, where the discrete variables adversely affect the prediction of continuous attributes, resulting in suboptimal performance. As shown in \zcref{tab:feat}, although discrete indices improve stylistic consistency, they lead to inferior FID and KID scores when directly incorporated into the generation process. To mitigate this issue, we treat the one-hot encodings of feature indices as continuous variables during generation.

\begin{table}[t]
    \centering
    \caption{Comparison of generation quality across different feature codes and generation strategies.}
    \label{tab:feat}
    \scalebox{0.85}{
        \begin{tabular}{c|cc}
            \hline
            Method                                    & FID            & KID           \\ \hline
            Continuous Feature Code                   & 18.52          & 2.85          \\
            Discrete Feature Code                     & 18.81          & 3.47          \\
            Discrete Feature Code + DDPM (Continuous) & \textbf{18.51} & \textbf{2.38} \\ \hline
        \end{tabular}
    }
\end{table}
\begin{figure}[t]
    \centering
    \includegraphics[width=\linewidth]{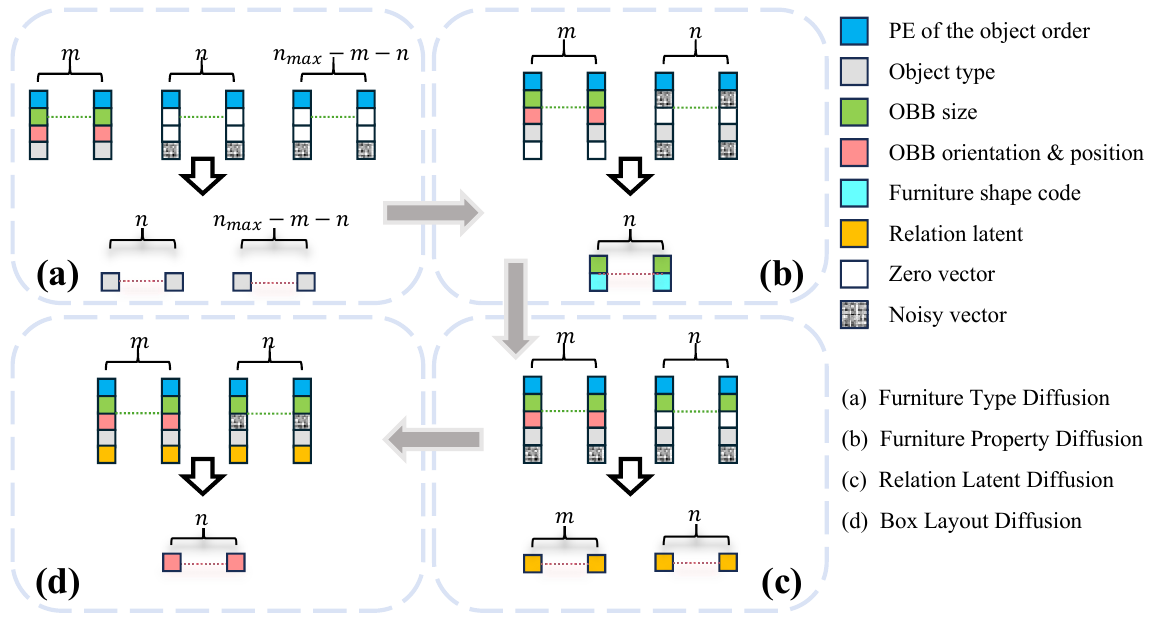}

    \caption{Evolution of node vectors during the cascaded generation process.}
    \label{fig:sp_vec}
\end{figure}

\subsection{Node Vector Generation}
\label{subsec:vec}

\zcref{fig:sp_vec} illustrates the node attribute generation process in our cascaded layout framework. The input initially consists of $N_{max}-m$ empty nodes and $m$ building element nodes. Building elements, similar to furniture instances, are represented by position embeddings, semantic types, and Oriented Bounding Boxes (OBBs) (with zero thickness). The generation proceeds in four stages:
\begin{enumerate}[leftmargin=*]\setlength\itemsep{1mm}
    \item[-] \textbf{Type Generation:} In the first stage, all unknown furniture attributes are initialized as zero vectors, while object types are initialized as noise and recovered through progressive denoising. After type prediction, $N_{max}-m-n$ redundant empty nodes are discarded, retaining only the $n$ valid furniture nodes together with the $m$ building element nodes for subsequent processing.

    \item[-] \textbf{Size and Feature Index Generation:} Conditioned on the first stage, object sizes and feature indices are generated via denoising. The predicted feature indices are used exclusively for object retrieval and are not employed as conditioning variables in subsequent stages.

    \item[-] \textbf{Relation Modeling:} In the third stage, relation latents are generated for all valid nodes.

    \item[-] \textbf{Layout Generation:} Finally, Oriented Bounding Boxes (OBBs) are predicted conditioned on the attributes synthesized in the preceding stages.
\end{enumerate}
Unknown attributes at each stage are represented by zero vectors as placeholders. Conditional generation is enabled by replacing these placeholders with the attributes predicted in earlier stages. Further training details are provided in \zcref{subsec:da}.

\subsection{Data Augmentation}
\label{subsec:da}
During the training of \name-G, the following data augmentation strategies are applied:
\begin{enumerate}[leftmargin=*]\setlength\itemsep{1mm}
    \item[-] Random masking of building elements, including walls, doors, and windows.
    \item[-] Random permutation of the node sequence.
    \item[-] Joint training with and without floor plan conditions.
    \item[-] Random rotations ($90^\circ$, $180^\circ$, $270^\circ$).
    \item[-] Completion-based training, where zero-vector placeholders for unknown attributes in preceding stages are replaced with ground-truth values, allowing the network to learn stage-wise conditional generation under varying levels of prior information.
\end{enumerate}

\subsection{Implementation Details of Applications}
\label{subsec:dapp}

\begin{figure}[t]
    \centering
    \includegraphics[width=\linewidth]{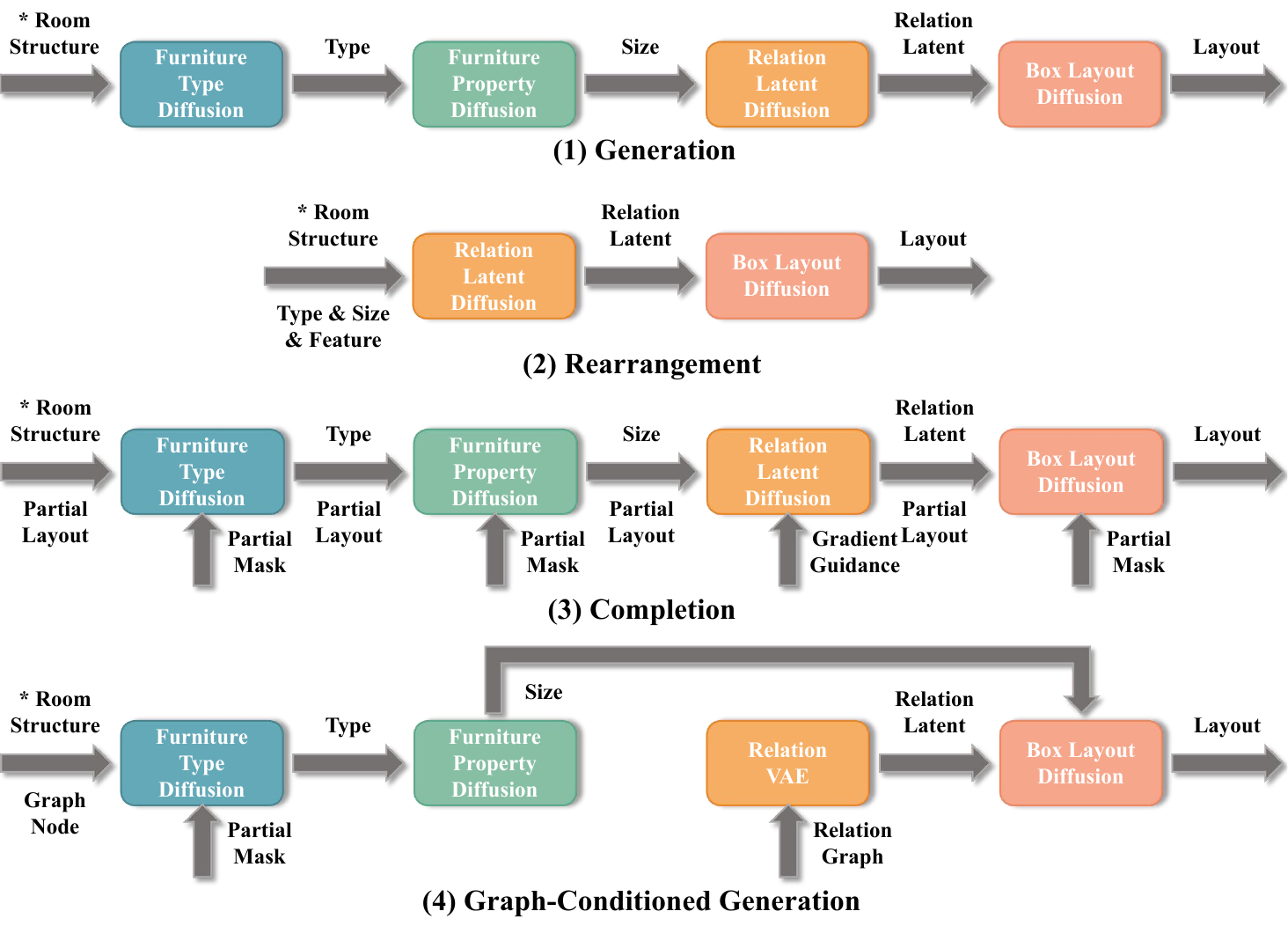}

    \caption{Pipelines for different applications. (1) Generation, (2) Rearrangement, (3) Completion, (4) Graph-conditioned generation.}
    \label{fig:edit_pipe}
\end{figure}

As illustrated in \zcref{fig:edit_pipe}, \name supports four primary applications: generation, rearrangement, completion, and graph-conditioned generation.

\paragraph{Generation}
In the generation pipeline, the room structure---including floor plan image features and building element vectors---can optionally be provided as input. The furniture type diffusion first produces the object type list. Subsequently, the furniture property diffusion, relation latent diffusion, and box layout diffusion stages sequentially generate object sizes, relation latents, and OBBs conditioned on the outputs of the preceding stages, yielding the final layout.

\paragraph{Rearrangement}
In the rearrangement task, the object type list and corresponding object sizes are provided as inputs. Consequently, only the relation latent diffusion and box layout diffusion stages are executed to obtain the rearranged layout.

\paragraph{Completion}
In the completion task, \name takes the complete OBB vectors as input at each stage, with ungenerated attributes padded by zero vectors. Conditioning on a partial layout is achieved by replacing the corresponding zero placeholders with the provided OBB attributes. The attributes of the partial layout remain fixed throughout all stages, thereby preserving the given layout while synthesizing a plausible complete scene.

For furniture type diffusion, furniture property diffusion, and box layout diffusion, this preservation is achieved by excluding the vectors corresponding to the partial layout objects during the denoising process. During the relation latent diffusion stage, relations in the partial layout are decoded using the Relation VAE and compared against the ground-truth relations in the original partial layout. The gradient of resulting loss is used to guide the generation of the relation latent, encouraging consistency with the relations present in the partial layout.

\paragraph{Graph-conditioned generation}
In this task, the input relation graph consists of object nodes and relation edges. The object type list is completed during the furniture type diffusion stage using the same strategy as in the completion task. After generating object sizes via the furniture property diffusion, the input relation graph is directly encoded using the Relation VAE to obtain the corresponding relation latent, which conditions the box layout diffusion to produce the final layout. Owing to our sparse relation formulation, the model supports generation controlled by relation graphs that are consistent with human scene descriptions, even when the provided graph is sparser than the predefined relation structure.

Other applications can be regarded as variants of these four primary tasks as follows.

\paragraph{LLM-guided scene generation}
This variant replaces the furniture type diffusion and relation latent diffusion stages with a Large Language Model (LLM). The relation latent is obtained by encoding the LLM output using the Relation VAE (see \zcref{subsec:p_llm} for prompt details).

\paragraph{Text/image-driven scene generation}
In this setting, an LLM or VLM translates the input text or image into a relation graph, after which the final scene is synthesized following the same pipeline as graph-conditioned generation (see \zcref{subsec:p_text} and \zcref{subsec:p_image} for prompt details).

\paragraph{Editing}
Editing is achieved by replacing user-specified attributes in the corresponding vectors at any stage, while preserving unchanged attributes during the denoising process. This mechanism is analogous to the partial layout conditioning used in the completion task.

\section{Prompts for LLMs \& VLMs}
\label{sec:prompts}
\subsection{Prompts for LLM-Guided Scene Generation}
\label{subsec:p_llm}

In the LLM-guided scene generation task, the LLM produces structured outputs in two stages: the first stage generates an object list, and the second stage predicts relations conditioned on this list. Both the object list and the relations are constrained to follow the same definitions and sparsity criteria as \name. The specific prompts are provided below:
\label{lst:llm}

\arxiv{
\begin{lstlisting}[language=HTML]
# Role Definition
You are an expert synthetic data generator and spatial logic engine. Your task is to simulate a realistic indoor scene (specifically a **Living Room** context) and generate structured data describing the objects and their spatial relationships.

# Master Task
Perform the following two steps sequentially and output the final result in a single JSON structure.

## Step 1: Object List Generation
1.  **Context:** Simulate a coherent, functional **Living Room** layout.
2.  **Select:** Choose 6 to 12 objects strictly from the **"Allowed Category List"** below.
3.  **Logic:** Ensure the combination makes sense (e.g., a `coffee_table` near a `multi_seat_sofa`; a `tv_stand` visible from the seating area).
4.  **Naming:** Assign unique IDs using the format `category_index` (1-based, resetting for each category). Example: `armchair_1`, `armchair_2`.

## Step 2: Relationship Extraction
Based on the objects generated in Step 1, simulate their 3D positions and determine spatial relationships.

**Sparsity Constraints (Group Logic):**
Iterate through every pair of objects (Object A, Object B). ONLY generate a relationship if the pair meets one of the following criteria:
1.  **Group Living Pair:** BOTH objects are in `living_type`.
2.  **Group Dining Pair:** BOTH objects are in `dining_type`.
3.  **Group Transitional Pair:** BOTH objects are in `between_living_dining`.
4.  **Lamp Exception:** At least ONE object is in `lamp_type` (Lamps relate to everything).

*If a pair does not meet any of these criteria, SKIP it.*

**Geometric Definitions (The Rules of Physics):**
Apply the following rules strictly to determine the relationship values.

**GLOBAL COORDINATE RULE:**
All definitions are based on **Object B's LOCAL Coordinate System**.
* **Origin:** Center of Object B.
* **Local +Y Axis:** The direction Object B is **FACING** (e.g., front of TV, seat of chair, drawers of cabinet).
* **Local +X Axis:** The "Right" side of Object B (Perpendicular to +Y).

1.  **Direction** (Values: `left`, `right`, `front`, `behind`, `under`, `above`)
    * *Mental Check:* Imagine standing at B, facing forward (+Y). Where is A?
    * `front`: A is in the +Y zone (B faces A).
    * `behind`: A is in the -Y zone (B backs A).
    * `right`/`left`: A is in the +X / -X zone.
    * `under`/`above`: Defined by Z-axis overlap (e.g., rug under table, lamp above table).

2.  **Distance** (Values: `attach to`, `adjacent`, `distant`)
    * Estimate physical distance ($d$) using standard furniture sizes (e.g., sofa width ~2m, chair ~0.5m).
    * `attach to`: $d < 0.2m$ (Touching).
    * `adjacent`: $0.2m \le d < 1.5m$ (Interaction range).
    * `distant`: $d \ge 1.5m$.

3.  **2D Alignment** (Values: `edge-align`, `x-align`, `y-align`)
    * `edge-align`: An edge of A is **collinear** with an edge of B.
    * `x-align`: Centers share similar **Local X** ($X_A \approx X_B \approx 0$).
        * *Meaning:* A lies on B's "Front-to-Back" center line.
    * `y-align`: Centers share similar **Local Y** ($Y_A \approx Y_B \approx 0$).
        * *Meaning:* A lies on B's "Left-to-Right" center line (Side-by-side).

4.  **Symmetry** (Values: `yes`)
    * Only if A and B are same type, size, mirror-symmetric, and on same horizontal plane.

---

# Definitions & Lists

### 1. Allowed Category List (Strict Vocabulary)
["armchair", "bookshelf", "cabinet", "ceiling_lamp", "chaise_longue_sofa", "chinese_chair", "coffee_table", "console_table", "corner_side_table", "desk", "dining_chair", "dining_table", "l_shaped_sofa", "lazy_sofa", "lounge_chair", "loveseat_sofa", "multi_seat_sofa", "pendant_lamp", "round_end_table", "shelf", "stool", "tv_stand", "wardrobe", "wine_cabinet"]

### 2. Group Definitions for Sparsity
* `living_type` = ["armchair", "chaise_longue_sofa", "chinese_chair", "coffee_table", "console_table", "corner_side_table", "desk", "l_shaped_sofa", "lazy_sofa", "lounge_chair", "loveseat_sofa", "multi_seat_sofa", "round_end_table", "stool", "tv_stand"]
* `dining_type` = ["chinese_chair", "console_table", "dining_chair", "dining_table"]
* `between_living_dining` = ["coffee_table", "console_table", "corner_side_table", "desk", "round_end_table", "dining_table"]
* `lamp_type` = ["ceiling_lamp", "pendant_lamp"]

---

# Output Format
Return **ONLY** a valid JSON object. Do not include markdown formatting (like ```json), explanations, or chatter.

**JSON Structure:**
{
  "object_list": [
    "string", ...
  ],
  "relationships": {
    "direction": [ ["obj_A", "obj_B", "value"], ... ],
    "distance": [ ["obj_A", "obj_B", "value"], ... ],
    "alignment": [ ["obj_A", "obj_B", "value"], ... ],
    "symmetry": [ ["obj_A", "obj_B", "value"], ... ]
  }
}
\end{lstlisting}
}
{
\begin{minted}[
    breaklines,       % 自动换行
    breakanywhere,    % 允许在任意字符处换行(防止长字符串不换行)
    frame=lines,      % 给顶部和底部加框，视觉上区分
    bgcolor=bg,   
    framesep=5mm,     % 框和内容的间距
    fontsize=\footnotesize
]{markdown}
# Role Definition
You are an expert synthetic data generator and spatial logic engine. Your task is to simulate a realistic indoor scene (specifically a **Living Room** context) and generate structured data describing the objects and their spatial relationships.

# Master Task
Perform the following two steps sequentially and output the final result in a single JSON structure.

## Step 1: Object List Generation
1.  **Context:** Simulate a coherent, functional **Living Room** layout.
2.  **Select:** Choose 6 to 12 objects strictly from the **"Allowed Category List"** below.
3.  **Logic:** Ensure the combination makes sense (e.g., a `coffee_table` near a `multi_seat_sofa`; a `tv_stand` visible from the seating area).
4.  **Naming:** Assign unique IDs using the format `category_index` (1-based, resetting for each category). Example: `armchair_1`, `armchair_2`.

## Step 2: Relationship Extraction
Based on the objects generated in Step 1, simulate their 3D positions and determine spatial relationships.

**Sparsity Constraints (Group Logic):**
Iterate through every pair of objects (Object A, Object B). ONLY generate a relationship if the pair meets one of the following criteria:
1.  **Group Living Pair:** BOTH objects are in `living_type`.
2.  **Group Dining Pair:** BOTH objects are in `dining_type`.
3.  **Group Transitional Pair:** BOTH objects are in `between_living_dining`.
4.  **Lamp Exception:** At least ONE object is in `lamp_type` (Lamps relate to everything).

*If a pair does not meet any of these criteria, SKIP it.*

**Geometric Definitions (The Rules of Physics):**
Apply the following rules strictly to determine the relationship values.

**GLOBAL COORDINATE RULE:**
All definitions are based on **Object B's LOCAL Coordinate System**.
* **Origin:** Center of Object B.
* **Local +Y Axis:** The direction Object B is **FACING** (e.g., front of TV, seat of chair, drawers of cabinet).
* **Local +X Axis:** The "Right" side of Object B (Perpendicular to +Y).

1.  **Direction** (Values: `left`, `right`, `front`, `behind`, `under`, `above`)
    * *Mental Check:* Imagine standing at B, facing forward (+Y). Where is A?
    * `front`: A is in the +Y zone (B faces A).
    * `behind`: A is in the -Y zone (B backs A).
    * `right`/`left`: A is in the +X / -X zone.
    * `under`/`above`: Defined by Z-axis overlap (e.g., rug under table, lamp above table).

2.  **Distance** (Values: `attach to`, `adjacent`, `distant`)
    * Estimate physical distance ($d$) using standard furniture sizes (e.g., sofa width ~2m, chair ~0.5m).
    * `attach to`: $d < 0.2m$ (Touching).
    * `adjacent`: $0.2m \le d < 1.5m$ (Interaction range).
    * `distant`: $d \ge 1.5m$.

3.  **2D Alignment** (Values: `edge-align`, `x-align`, `y-align`)
    * `edge-align`: An edge of A is **collinear** with an edge of B.
    * `x-align`: Centers share similar **Local X** ($X_A \approx X_B \approx 0$).
        * *Meaning:* A lies on B's "Front-to-Back" center line.
    * `y-align`: Centers share similar **Local Y** ($Y_A \approx Y_B \approx 0$).
        * *Meaning:* A lies on B's "Left-to-Right" center line (Side-by-side).

4.  **Symmetry** (Values: `yes`)
    * Only if A and B are same type, size, mirror-symmetric, and on same horizontal plane.

---

# Definitions & Lists

### 1. Allowed Category List (Strict Vocabulary)
["armchair", "bookshelf", "cabinet", "ceiling_lamp", "chaise_longue_sofa", "chinese_chair", "coffee_table", "console_table", "corner_side_table", "desk", "dining_chair", "dining_table", "l_shaped_sofa", "lazy_sofa", "lounge_chair", "loveseat_sofa", "multi_seat_sofa", "pendant_lamp", "round_end_table", "shelf", "stool", "tv_stand", "wardrobe", "wine_cabinet"]

### 2. Group Definitions for Sparsity
* `living_type` = ["armchair", "chaise_longue_sofa", "chinese_chair", "coffee_table", "console_table", "corner_side_table", "desk", "l_shaped_sofa", "lazy_sofa", "lounge_chair", "loveseat_sofa", "multi_seat_sofa", "round_end_table", "stool", "tv_stand"]
* `dining_type` = ["chinese_chair", "console_table", "dining_chair", "dining_table"]
* `between_living_dining` = ["coffee_table", "console_table", "corner_side_table", "desk", "round_end_table", "dining_table"]
* `lamp_type` = ["ceiling_lamp", "pendant_lamp"]

---

# Output Format
Return **ONLY** a valid JSON object. Do not include markdown formatting (like ```json), explanations, or chatter.

**JSON Structure:**
{
  "object_list": [
    "string", ...
  ],
  "relationships": {
    "direction": [ ["obj_A", "obj_B", "value"], ... ],
    "distance": [ ["obj_A", "obj_B", "value"], ... ],
    "alignment": [ ["obj_A", "obj_B", "value"], ... ],
    "symmetry": [ ["obj_A", "obj_B", "value"], ... ]
  }
}
\end{minted}
}

\subsection{Prompts for Text-Driven Scene Generation}
\label{subsec:p_text}

Similarly, in the text-driven scene generation task, the LLM produces outputs in two stages. In the first stage, objects are extracted from the input text and mapped to the closest valid categories in the predefined object vocabulary when necessary. In the second stage, relations are identified based on the extracted object list, while relations that violate the sparsity constraints are discarded. The corresponding prompts are provided below:
\label{lst:text}

\arxiv{
\begin{lstlisting}[language=HTML]
# Role Definition
You are an expert NLP information extraction engine. Your task is to parse a text description of an indoor scene and extract structured data (objects and relationships) strictly according to the defined schema.

# Input
A text paragraph describing an indoor scene.

# Task Instructions
Perform the extraction in two phases.

## Phase 1: Object Extraction & Synonym Mapping
1.  **Identify Objects:** Extract all furniture and objects mentioned in the text.
2.  **Map to Category:** Map each extracted object to the **closest match** in the "Allowed Category List" below.
    * *Example:* If text says "couch", output `multi_seat_sofa`. If text says "bedside table", output `round_end_table` or `corner_side_table`.
    * *Unknowns:* If an object cannot be mapped to ANY category in the list (e.g., "dog", "plant", "window"), **IGNORE** it.
3.  **ID Assignment:** Assign IDs (e.g., `_1`) to distinguish multiple instances of the same category.

## Phase 2: Relationship Extraction & Filtering
1.  **Extract Descriptions:** Identify spatial descriptions connecting two mapped objects (Object A -> Object B).
2.  **Map to Attributes:** Convert natural language descriptions into the standardized values:
    * *Direction:* left, right, front, behind, under, above.
    * *Distance:* attach to (touching), adjacent (close), distant (far).
    * *Alignment:* edge-align, x-align, y-align.
    * *Symmetry:* yes.
3.  **Sparsity Filter (The Gatekeeper):**
    * Check if the pair `(Object_A, Object_B)` belongs to a valid interaction group (Living, Dining, Transitional, or Lamp).
    * **Rule:** If the pair is **NOT** in a valid group, **DISCARD** the relationship immediately, even if the text mentions it.
    * **Rule:** If the text describes a relationship type not defined (e.g., "Object A is darker than Object B"), **DISCARD** it.

---

# Definitions & Schema

## 1. Allowed Category List (Target Vocabulary)
["armchair", "bookshelf", "cabinet", "ceiling_lamp", "chaise_longue_sofa", "chinese_chair", "coffee_table", "console_table", "corner_side_table", "desk", "dining_chair", "dining_table", "l_shaped_sofa", "lazy_sofa", "lounge_chair", "loveseat_sofa", "multi_seat_sofa", "pendant_lamp", "round_end_table", "shelf", "stool", "tv_stand", "wardrobe", "wine_cabinet"]

## 2. Group Logic (Sparsity Rules)
Only extract relationships if the pair exists in:
* **Group Living:** Both objects in ["armchair", "chaise_longue_sofa", "chinese_chair", "coffee_table", "console_table", "corner_side_table", "desk", "l_shaped_sofa", "lazy_sofa", "lounge_chair", "loveseat_sofa", "multi_seat_sofa", "round_end_table", "stool", "tv_stand"]
* **Group Dining:** Both objects in ["chinese_chair", "console_table", "dining_chair", "dining_table"]
* **Group Transitional:** Both objects in ["coffee_table", "console_table", "corner_side_table", "desk", "round_end_table", "dining_table"]
* **Lamp Exception:** At least one object is ["ceiling_lamp", "pendant_lamp"] (Connects to anything).

## 3. Relationship Definitions (Text Mapping)
* **Direction:** Based on B's local facing. "In front of", "behind", "to the left/right of", "under/above".
* **Distance:**
    * "Touching", "Attached", "Against" -> `attach to`
    * "Near", "Close to", "Next to" -> `adjacent`
    * "Far from", "Across the room" -> `distant`
* **Alignment/Symmetry:** Only output if explicitly stated (e.g., "aligned with", "centered on", "symmetrically placed").

---

# Output Format
Output **ONLY** a valid JSON object.

```json
{
  "object_list": [
    "multi_seat_sofa_1", "coffee_table_1", ...
  ],
  "relationships": {
    "direction": [ ["obj_A", "obj_B", "value"], ... ],
    "distance": [ ["obj_A", "obj_B", "value"], ... ],
    "alignment": [ ["obj_A", "obj_B", "value"], ... ],
    "symmetry": [ ["obj_A", "obj_B", "value"], ... ]
  }
}
\end{lstlisting}
}
{
\begin{minted}[
    breaklines,       % 自动换行
    breakanywhere,    % 允许在任意字符处换行(防止长字符串不换行)
    frame=lines,      % 给顶部和底部加框，视觉上区分
    bgcolor=bg,   
    framesep=5mm,     % 框和内容的间距
    fontsize=\footnotesize
]{markdown}
# Role Definition
You are an expert NLP information extraction engine. Your task is to parse a text description of an indoor scene and extract structured data (objects and relationships) strictly according to the defined schema.

# Input
A text paragraph describing an indoor scene.

# Task Instructions
Perform the extraction in two phases.

## Phase 1: Object Extraction & Synonym Mapping
1.  **Identify Objects:** Extract all furniture and objects mentioned in the text.
2.  **Map to Category:** Map each extracted object to the **closest match** in the "Allowed Category List" below.
    * *Example:* If text says "couch", output `multi_seat_sofa`. If text says "bedside table", output `round_end_table` or `corner_side_table`.
    * *Unknowns:* If an object cannot be mapped to ANY category in the list (e.g., "dog", "plant", "window"), **IGNORE** it.
3.  **ID Assignment:** Assign IDs (e.g., `_1`) to distinguish multiple instances of the same category.

## Phase 2: Relationship Extraction & Filtering
1.  **Extract Descriptions:** Identify spatial descriptions connecting two mapped objects (Object A -> Object B).
2.  **Map to Attributes:** Convert natural language descriptions into the standardized values:
    * *Direction:* left, right, front, behind, under, above.
    * *Distance:* attach to (touching), adjacent (close), distant (far).
    * *Alignment:* edge-align, x-align, y-align.
    * *Symmetry:* yes.
3.  **Sparsity Filter (The Gatekeeper):**
    * Check if the pair `(Object_A, Object_B)` belongs to a valid interaction group (Living, Dining, Transitional, or Lamp).
    * **Rule:** If the pair is **NOT** in a valid group, **DISCARD** the relationship immediately, even if the text mentions it.
    * **Rule:** If the text describes a relationship type not defined (e.g., "Object A is darker than Object B"), **DISCARD** it.

---

# Definitions & Schema

## 1. Allowed Category List (Target Vocabulary)
["armchair", "bookshelf", "cabinet", "ceiling_lamp", "chaise_longue_sofa", "chinese_chair", "coffee_table", "console_table", "corner_side_table", "desk", "dining_chair", "dining_table", "l_shaped_sofa", "lazy_sofa", "lounge_chair", "loveseat_sofa", "multi_seat_sofa", "pendant_lamp", "round_end_table", "shelf", "stool", "tv_stand", "wardrobe", "wine_cabinet"]

## 2. Group Logic (Sparsity Rules)
Only extract relationships if the pair exists in:
* **Group Living:** Both objects in ["armchair", "chaise_longue_sofa", "chinese_chair", "coffee_table", "console_table", "corner_side_table", "desk", "l_shaped_sofa", "lazy_sofa", "lounge_chair", "loveseat_sofa", "multi_seat_sofa", "round_end_table", "stool", "tv_stand"]
* **Group Dining:** Both objects in ["chinese_chair", "console_table", "dining_chair", "dining_table"]
* **Group Transitional:** Both objects in ["coffee_table", "console_table", "corner_side_table", "desk", "round_end_table", "dining_table"]
* **Lamp Exception:** At least one object is ["ceiling_lamp", "pendant_lamp"] (Connects to anything).

## 3. Relationship Definitions (Text Mapping)
* **Direction:** Based on B's local facing. "In front of", "behind", "to the left/right of", "under/above".
* **Distance:**
    * "Touching", "Attached", "Against" -> `attach to`
    * "Near", "Close to", "Next to" -> `adjacent`
    * "Far from", "Across the room" -> `distant`
* **Alignment/Symmetry:** Only output if explicitly stated (e.g., "aligned with", "centered on", "symmetrically placed").

---

# Output Format
Output **ONLY** a valid JSON object.

```json
{
  "object_list": [
    "multi_seat_sofa_1", "coffee_table_1", ...
  ],
  "relationships": {
    "direction": [ ["obj_A", "obj_B", "value"], ... ],
    "distance": [ ["obj_A", "obj_B", "value"], ... ],
    "alignment": [ ["obj_A", "obj_B", "value"], ... ],
    "symmetry": [ ["obj_A", "obj_B", "value"], ... ]
  }
}
\end{minted}
}

\subsection{Prompts for Image-Driven Scene Generation}
\label{subsec:p_image}

For image-driven scene generation, the VLM extracts object instances from the input image and predicts relations among them following the same relation definition criteria as \name. The corresponding prompts are provided below:
\label{lst:image}

\arxiv{
\begin{lstlisting}[language=HTML]
# Role Definition
You are an expert spatial perception assistant specializing in indoor scene understanding. Your task is to extract objects and their spatial relationships from the provided indoor image based on strict definition rules.

# Task Instructions
Complete the task in two sequential steps. Do not skip steps.

## Step 1: Object Extraction
Detect objects in the image belonging ONLY to the following allowed category list. Assign a unique ID to each detected object (e.g., `_1`, `_2`) based on its category.

**Allowed Categories:**
["armchair", "bookshelf", "cabinet", "ceiling_lamp", "chaise_longue_sofa", "chinese_chair", "coffee_table", "console_table", "corner_side_table", "desk", "dining_chair", "dining_table", "l_shaped_sofa", "lazy_sofa", "lounge_chair", "loveseat_sofa", "multi_seat_sofa", "pendant_lamp", "round_end_table", "shelf", "stool", "tv_stand", "wardrobe", "wine_cabinet"]

**Output Format for Step 1:**
List detected objects: [category_id, category_id, ...]

## Step 2: Relationship Extraction
Based on the objects extracted in Step 1, determine the spatial relationships.
Organize the output by **Relationship Category**. Within each category, list the relationships as **Triplets**: `[Object_A, Object_B, Value]`.

**Sparsity Constraints (Group Logic):**
Only calculate relationships if the pair of objects (Object A, Object B) both belong to at least one of the following specific groups:
1.  **Group Living:** ["armchair", "chaise_longue_sofa", "chinese_chair", "coffee_table", "console_table", "corner_side_table", "desk", "l_shaped_sofa", "lazy_sofa", "lounge_chair", "loveseat_sofa", "multi_seat_sofa", "round_end_table", "stool", "tv_stand"]
2.  **Group Dining:** ["chinese_chair", "console_table", "dining_chair", "dining_table"]
3.  **Group Transitional:** ["coffee_table", "console_table", "corner_side_table", "desk", "round_end_table", "dining_table"]

**Relationship Definitions & Triplet Values:**

**GLOBAL COORDINATE RULE:**
For ALL relationships (Direction, Distance, Alignment), strict adherence to **Object B's LOCAL Coordinate System** is required.
* **Origin:** Center of Object B.
* **Local +Y Axis:** The direction Object B is **FACING** (e.g., screen of TV, seat of chair, front drawers of cabinet).
* **Local +X Axis:** Perpendicular to +Y (The "Right" side of Object B).

1.  **Direction** (Values: `left`, `right`, `front`, `behind`, `under`, `above`)
    * **Calculation:** Project Object A onto Object B's Local X-Y plane.
    * `front`: Object A is in the +Y direction of B.
    * `behind`: Object A is in the -Y direction of B.
    * `right`: Object A is in the +X direction of B.
    * `left`: Object A is in the -X direction of B.
    * `under`/`above`: Based on Z-axis relative to B's bounding box.

2.  **Distance** (Values: `attach to`, `adjacent`, `distant`)
    * Estimate the physical distance ($d$) between the closest points of the two objects' OBBs.
    * **Scale Reference:** Use standard furniture sizes (e.g., chair width $\approx 0.5$m) to estimate.
    * **Thresholds:**
        * `attach to`: $d < 0.2\text{m}$ (Touching).
        * `adjacent`: $0.2\text{m} \le d < 1.5\text{m}$.
        * `distant`: $d \ge 1.5\text{m}$.

3.  **2D Alignment** (Values: `edge-align`, `x-align`, `y-align`)
    * **CRITICAL:** Calculate alignment purely based on **Object B's Local Frame** defined above.
    * **Definitions:**
        * `edge-align`: One edge of Object A is **collinear** with an edge of Object B.
        * `x-align`: The centers of A and B share the same **Local X coordinate** (Difference in X $\approx 0$).
            * *Visual Interpretation:* Object A lies on the **Center Line** (Front/Back axis) of Object B. (e.g., A chair tucked directly *under/front* of a desk).
        * `y-align`: The centers of A and B share the same **Local Y coordinate** (Difference in Y $\approx 0$).
            * *Visual Interpretation:* Object A lies directly to the **Side** (Left/Right) of Object B. (e.g., Two chairs placed perfectly side-by-side).
    * *Note: Only output if a clear alignment exists.*

4.  **Symmetry** (Values: `yes`)
    * Only output if A and B are of the same type, size, mirror-symmetric, and on the same horizontal plane.

## Output Format
Please output the result in a structured JSON format.

```json
{
  "step_1_objects": [
    "dining_table_1",
    "dining_chair_1",
    "dining_chair_2"
  ],
  "step_2_relationships": {
    "direction": [
      ["dining_chair_1", "dining_table_1", "front"]
    ],
    "distance": [
      ["dining_chair_1", "dining_table_1", "adjacent"]
    ],
    "alignment": [
      ["dining_chair_1", "dining_table_1", "edge-align"],
      ["dining_chair_2", "dining_table_1", "x-align"]
    ],
    "symmetry": [
      ["dining_chair_1", "dining_chair_2", "yes"]
    ]
  }
}
\end{lstlisting}
}
{
\begin{minted}[
    breaklines,       % 自动换行
    breakanywhere,    % 允许在任意字符处换行(防止长字符串不换行)
    frame=lines,      % 给顶部和底部加框，视觉上区分
    bgcolor=bg,   
    framesep=5mm,     % 框和内容的间距
    fontsize=\footnotesize
]{markdown}
# Role Definition
You are an expert spatial perception assistant specializing in indoor scene understanding. Your task is to extract objects and their spatial relationships from the provided indoor image based on strict definition rules.

# Task Instructions
Complete the task in two sequential steps. Do not skip steps.

## Step 1: Object Extraction
Detect objects in the image belonging ONLY to the following allowed category list. Assign a unique ID to each detected object (e.g., `_1`, `_2`) based on its category.

**Allowed Categories:**
["armchair", "bookshelf", "cabinet", "ceiling_lamp", "chaise_longue_sofa", "chinese_chair", "coffee_table", "console_table", "corner_side_table", "desk", "dining_chair", "dining_table", "l_shaped_sofa", "lazy_sofa", "lounge_chair", "loveseat_sofa", "multi_seat_sofa", "pendant_lamp", "round_end_table", "shelf", "stool", "tv_stand", "wardrobe", "wine_cabinet"]

**Output Format for Step 1:**
List detected objects: [category_id, category_id, ...]

## Step 2: Relationship Extraction
Based on the objects extracted in Step 1, determine the spatial relationships.
Organize the output by **Relationship Category**. Within each category, list the relationships as **Triplets**: `[Object_A, Object_B, Value]`.

**Sparsity Constraints (Group Logic):**
Only calculate relationships if the pair of objects (Object A, Object B) both belong to at least one of the following specific groups:
1.  **Group Living:** ["armchair", "chaise_longue_sofa", "chinese_chair", "coffee_table", "console_table", "corner_side_table", "desk", "l_shaped_sofa", "lazy_sofa", "lounge_chair", "loveseat_sofa", "multi_seat_sofa", "round_end_table", "stool", "tv_stand"]
2.  **Group Dining:** ["chinese_chair", "console_table", "dining_chair", "dining_table"]
3.  **Group Transitional:** ["coffee_table", "console_table", "corner_side_table", "desk", "round_end_table", "dining_table"]

**Relationship Definitions & Triplet Values:**

**GLOBAL COORDINATE RULE:**
For ALL relationships (Direction, Distance, Alignment), strict adherence to **Object B's LOCAL Coordinate System** is required.
* **Origin:** Center of Object B.
* **Local +Y Axis:** The direction Object B is **FACING** (e.g., screen of TV, seat of chair, front drawers of cabinet).
* **Local +X Axis:** Perpendicular to +Y (The "Right" side of Object B).

1.  **Direction** (Values: `left`, `right`, `front`, `behind`, `under`, `above`)
    * **Calculation:** Project Object A onto Object B's Local X-Y plane.
    * `front`: Object A is in the +Y direction of B.
    * `behind`: Object A is in the -Y direction of B.
    * `right`: Object A is in the +X direction of B.
    * `left`: Object A is in the -X direction of B.
    * `under`/`above`: Based on Z-axis relative to B's bounding box.

2.  **Distance** (Values: `attach to`, `adjacent`, `distant`)
    * Estimate the physical distance ($d$) between the closest points of the two objects' OBBs.
    * **Scale Reference:** Use standard furniture sizes (e.g., chair width $\approx 0.5$m) to estimate.
    * **Thresholds:**
        * `attach to`: $d < 0.2\text{m}$ (Touching).
        * `adjacent`: $0.2\text{m} \le d < 1.5\text{m}$.
        * `distant`: $d \ge 1.5\text{m}$.

3.  **2D Alignment** (Values: `edge-align`, `x-align`, `y-align`)
    * **CRITICAL:** Calculate alignment purely based on **Object B's Local Frame** defined above.
    * **Definitions:**
        * `edge-align`: One edge of Object A is **collinear** with an edge of Object B.
        * `x-align`: The centers of A and B share the same **Local X coordinate** (Difference in X $\approx 0$).
            * *Visual Interpretation:* Object A lies on the **Center Line** (Front/Back axis) of Object B. (e.g., A chair tucked directly *under/front* of a desk).
        * `y-align`: The centers of A and B share the same **Local Y coordinate** (Difference in Y $\approx 0$).
            * *Visual Interpretation:* Object A lies directly to the **Side** (Left/Right) of Object B. (e.g., Two chairs placed perfectly side-by-side).
    * *Note: Only output if a clear alignment exists.*

4.  **Symmetry** (Values: `yes`)
    * Only output if A and B are of the same type, size, mirror-symmetric, and on the same horizontal plane.

## Output Format
Please output the result in a structured JSON format.

```json
{
  "step_1_objects": [
    "dining_table_1",
    "dining_chair_1",
    "dining_chair_2"
  ],
  "step_2_relationships": {
    "direction": [
      ["dining_chair_1", "dining_table_1", "front"]
    ],
    "distance": [
      ["dining_chair_1", "dining_table_1", "adjacent"]
    ],
    "alignment": [
      ["dining_chair_1", "dining_table_1", "edge-align"],
      ["dining_chair_2", "dining_table_1", "x-align"]
    ],
    "symmetry": [
      ["dining_chair_1", "dining_chair_2", "yes"]
    ]
  }
}
\end{minted}
}

\section{More Experiments and Discussion}

\subsection{Evaluation of Physical Plausibility}
\label{subsec:pc}

To evaluate the physical plausibility of \name in comparison with existing methods, we adopt the following metrics: the out-of-bound rate ($\mathrm{R}_{\mathrm{out}}$)~\cite{yang2024physcene} and the walkable-area ratio ($\mathrm{R}_{\mathrm{walk}}$)~\cite{yang2024physcene}.
\begin{enumerate}[leftmargin=*]\setlength\itemsep{1mm}
    \item[-] \textbf{Out-of-bound rate ($\mathrm{R}_{\mathrm{out}}$):} The proportion of objects placed outside the floor plan.
    \item[-] \textbf{Walkable-area ratio ($\mathrm{R}_{\mathrm{walk}}$):} The average ratio of the largest connected walkable area to the total walkable area in the room, where walkable areas are defined as regions with a minimum width $\geq$ \SI{0.5}{m}.
\end{enumerate}

As shown in \zcref{tab:phy}, \name achieves the lowest $\mathrm{R}_{\mathrm{out}}$ and the highest $\mathrm{R}_{\mathrm{walk}}$ among the evaluated methods, indicating improved physical plausibility.

\begin{table}[t]
    \centering
    \caption{Comparison with different methods on physical plausibility.}
    \label{tab:phy}

    \footnotesize
    \setlength{\tabcolsep}{12pt}
    \renewcommand{\arraystretch}{1.35}

    \begin{tabular}{l | cc}
        \toprule
        \textbf{Method}
                   & $\mathbf{R}_{\mathbf{out}} \downarrow$
                   & $\mathbf{R}_{\mathbf{walk}} \uparrow$                   \\
        \midrule

        LayoutGPT  & 29.76                                  & 84.00          \\
        ATISS      & 20.28                                  & 83.39          \\
        DiffuScene & 24.42                                  & 84.72          \\
        GLTScene   & 19.87                                  & 87.37          \\
        \midrule

        Ours-G     & \textbf{18.21}                         & \textbf{88.72} \\
        \bottomrule
    \end{tabular}
\end{table}

\subsection{Comparison with LayoutVLM}
\label{subsec:layoutvlm}

\begin{table}[t]
    \centering
    \caption{Comparison with LayoutVLM~\cite{sun2025layoutvlm} on physical constraint metrics for the rearrangement task.}
    \label{tab:layoutvlm}

    \footnotesize
    \setlength{\tabcolsep}{12pt}
    \renewcommand{\arraystretch}{1.35}

    \begin{tabular}{l | ccc}
        \toprule
        \textbf{Method}
                  & \textbf{IoU} $\downarrow$
                  & $\mathbf{R}_{\mathbf{out}} \downarrow$
                  & $\mathbf{R}_{\mathbf{walk}} \uparrow$                                    \\
        \midrule

        LayoutVLM & 1.05                                   & 21.44          & \textbf{89.23} \\

        Ours      & \textbf{0.80}                          & \textbf{19.02} & 88.95          \\
        \bottomrule
    \end{tabular}
\end{table}
\begin{figure}[t]
    \centering
    \includegraphics[width=0.9\linewidth]{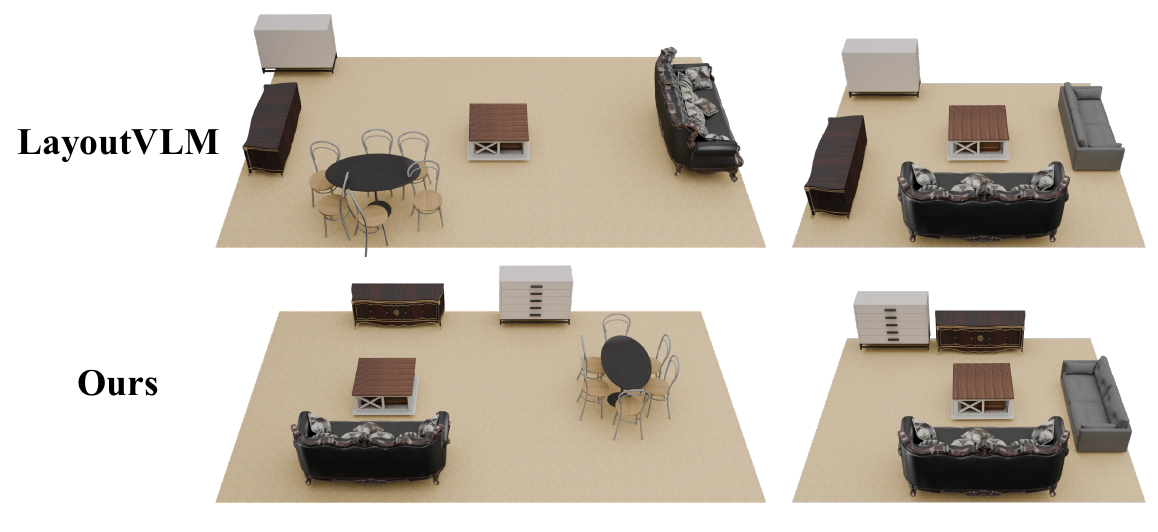}

    \caption{Qualitative comparison with LayoutVLM~\cite{sun2025layoutvlm}. LayoutVLM~\cite{sun2025layoutvlm} tends to place objects along the perimeter or at the center, resulting in implausible layouts.}
    \label{fig:layoutvlm}
\end{figure}

Since LayoutVLM~\cite{sun2025layoutvlm} requires object lists and text prompts as input, we compare our method with LayoutVLM on the rearrangement task under identical test set constraints. The quantitative results are presented in \zcref{tab:layoutvlm}.

\name demonstrates improved physical realism overall. Although LayoutVLM achieves a higher $\mathrm{R}_{\mathrm{walk}}$, qualitative inspection suggests that this may result from placing objects toward room centers or perimeters, which can lead to suboptimal alignments and less natural functional organization, as illustrated in \zcref{fig:layoutvlm}.

\subsection{Inference Time of Applications}
\label{subsec:tapp}

\begin{table}[t]
  \centering
  \caption{Average inference time per scene (in milliseconds) for different applications, using UniPC~\cite{zhao2023unipc} as the sampler with 15 inference steps.}
  \label{tab:time_app}

  \footnotesize
  \setlength{\tabcolsep}{2pt}
  \renewcommand{\arraystretch}{1.35}

  \begin{tabular}{ll | cc}
    \toprule
    \multicolumn{2}{c|}{\textbf{Application}}
     & \textbf{Inference Time}
     & \textbf{Stage Latency}                                                    \\
    \midrule

    \multirow{3}{*}{\textbf{With Floor Plan}}
     & Generation              & 25.57 & $7.82{\,+\,}5.48{\,+\,}5.95{\,+\,}6.32$ \\
     & Rearrange               & 11.53 & $5.62{\,+\,}5.91$                       \\
     & Completion              & 26.67 & $7.94{\,+\,}5.41{\,+\,}7.03{\,+\,}6.29$ \\
    \midrule 

    \multirow{3}{*}{\textbf{Without Floor Plan}}
     & Generation              & 24.27 & $7.63{\,+\,}5.21{\,+\,}5.47{\,+\,}5.96$ \\
     & Rearrange               & 10.75 & $5.33{\,+\,}5.42$                       \\
     & Completion              & 25.54 & $7.58{\,+\,}5.17{\,+\,}6.75{\,+\,}6.04$ \\
    \bottomrule
  \end{tabular}
\end{table}

To analyze the inference efficiency of \name, we measure the scene generation time across three applications (generation, rearrangement, and completion), both with and without room structure conditioning. UniPC~\cite{zhao2023unipc} is employed as the sampler with 15 inference steps.

As shown in \zcref{tab:time_app}, \name generates a scene within \SI{30}{ms} across all task settings. Scenarios conditioned on room structures generally require additional computation due to attention over floor plan image features and building element representations. Furthermore, the completion task incurs slightly higher inference time owing to the additional VAE decoding and gradient guidance involved in the relation latent diffusion stage.

\subsection{Diversity}
\label{subsec:div}

\begin{figure}[t]
    \centering
    \includegraphics[width=\linewidth]{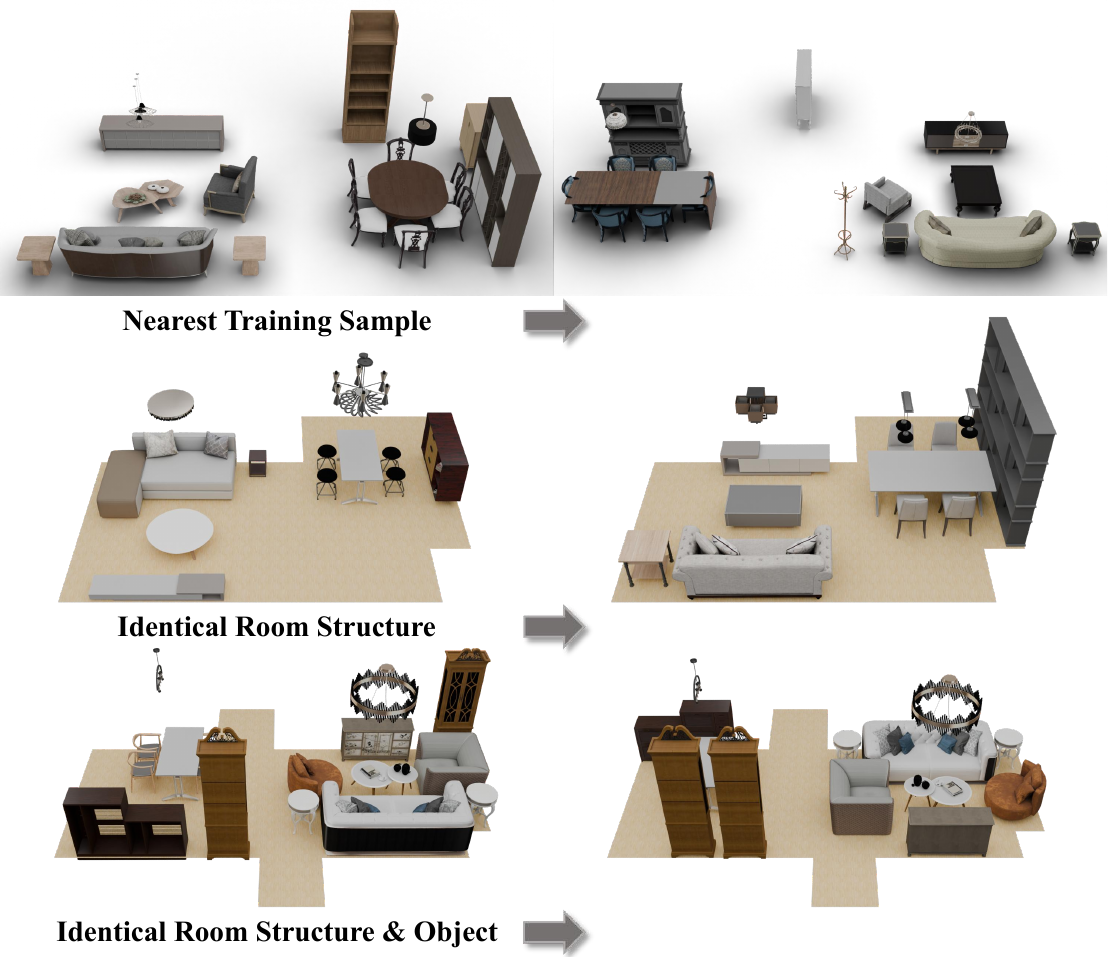}

    \caption{Qualitative diversity analysis: (Top) Nearest training samples (based on object type lists) for unconditionally generated scenes. (Middle) Scenes generated with different random seeds under identical room structures. (Bottom) Rearrangement results obtained with different random seeds given identical room structures and object lists.}
    \label{fig:diverse}
\end{figure}

To assess whether \name memorizes training data, we evaluate the diversity of generated scenes. As illustrated in \zcref{fig:diverse}, scenes generated under unconditional settings are consistent with the overall training distribution while exhibiting substantial variation at the individual layout level.

Quantitatively, we compute the IoU between \num{1000} generated scenes and their closest training layouts based on object lists, yielding an average IoU of 3.17\%. Under additional constraints (\eg fixed room structures), five random samples generated with identical room structures exhibit an average IoU of 25.52\%. Under the strictest condition of identical room structures and object lists, the IoU between generated layouts remains at 30.73\%.

\subsection{Error Propagation}
\label{subsec:ep}

While the four-stage cascading design may introduce potential error propagation, it mitigates the high-dimensional collapse commonly observed in single-stage models. Each stage contributes to reducing upstream noise:
\begin{enumerate}[leftmargin=*]\setlength\itemsep{1mm}
    \item[-] \textbf{S1-to-S2:} Object property diffusion leverages category-specific priors to maintain reasonable size predictions, even under out-of-distribution object lists.
    \item[-] \textbf{S2-to-S3:} Sparse relations generated by relation latent diffusion encode functional topology and are robust to moderate size deviations.
    \item[-] \textbf{S3-to-S4:} The jointly trained Relation VAE and box layout diffusion module accommodate mild latent perturbations.
\end{enumerate}

\begin{figure}[t]
    \centering
    \includegraphics[width=0.8\linewidth]{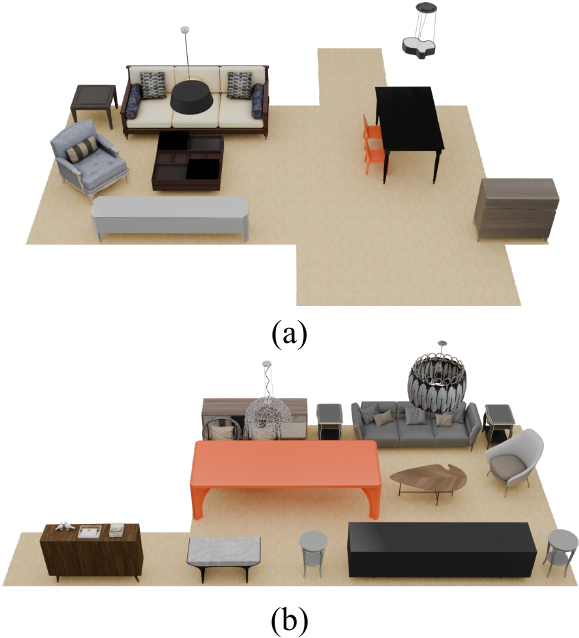}

    \caption{Example of error propagation. When the furniture property
        diffusion produces error sizes (highlighted in orange), the subsequent stages lack the capability to rectify the earlier mistakes, despite synthesizing relatively plausible layout.}
    \label{fig:failure_pipe}
\end{figure}

Large upstream errors (\eg excessively oversized objects) may still persist into the final stage. Nevertheless, \name typically produces globally plausible layouts under such conditions, as illustrated in \zcref{fig:failure_pipe}.

\subsection{More Qualitative Results}
\label{subsec:mr}

\begin{figure*}[t]
    \centering
    \includegraphics[width=\linewidth]{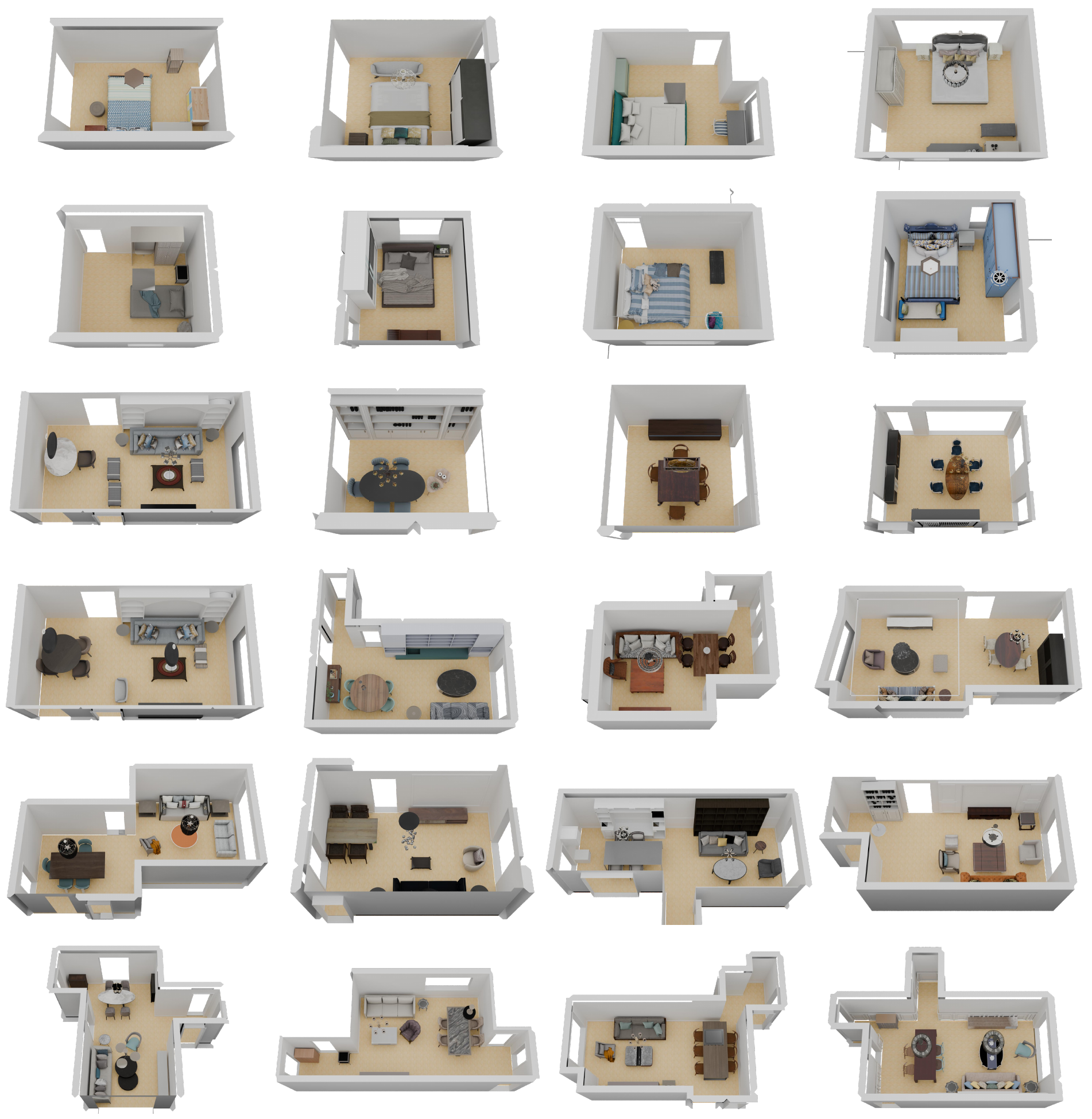}

    \caption{Diverse scenes generated by \name under architectural element constraints (\eg walls, doors, and windows). Explicit modeling of these elements allows the output layouts to reserve sufficient operational space, improving the physical plausibility of the scene.}
    \label{fig:sp_wall}
\end{figure*}

We provide additional examples of scenes generated under building element constraints. As shown in \zcref{fig:sp_wall}, \name adheres to these structural constraints and produces physically plausible layouts.
\end{document}